\documentclass{article} % DO NOT CHANGE THIS

\usepackage[main,final,nonatbib]{neurips_2025}

\usepackage{lipsum}
\usepackage{wrapfig}
\usepackage{thmtools,thm-restate}

\usepackage{dsfont}
\usepackage{xcolor}

\usepackage{xargs}                      % Use more than one optional parameter in a new commands
\usepackage[pdftex,dvipsnames]{xcolor}  % Coloured text etc.
\usepackage[colorinlistoftodos,prependcaption,textsize=tiny]{todonotes}
\newcommandx{\unsure}[2][1=]{\todo[linecolor=red,backgroundcolor=red!25,bordercolor=red,#1]{#2}}
\newcommandx{\change}[2][1=]{\todo[linecolor=blue,backgroundcolor=blue!25,bordercolor=blue,#1]{#2}}
\newcommandx{\info}[2][1=]{\todo[linecolor=OliveGreen,backgroundcolor=OliveGreen!25,bordercolor=OliveGreen,#1]{#2}}
\newcommandx{\improvement}[2][1=]{\todo[linecolor=Plum,backgroundcolor=Plum!25,bordercolor=Plum,#1]{#2}}
\newcommandx{\thiswillnotshow}[2][1=]{\todo[disable,#1]{#2}}

\usepackage{xspace}
\usepackage{etoolbox}
\robustify\parbox

\usepackage[utf8]{inputenc} % allow utf-8 input
\usepackage[T1]{fontenc}    % use 8-bit T1 fonts
\usepackage[colorlinks=true,
            linkcolor=blue,
            citecolor=blue,
            urlcolor=blue]{hyperref}
\usepackage{url}            % simple URL typesetting
\usepackage{booktabs}       % professional-quality tables
\usepackage{amsfonts}       % blackboard math symbols
\usepackage{nicefrac}       % compact symbols for 1/2, etc.
\usepackage{microtype}      % microtypography
\usepackage{xcolor}         % colors
\usepackage[toc,page]{appendix}

\usepackage{transparent}

\usepackage{pgfplots}
\usepgfplotslibrary{colorbrewer}  % <-- needed for Set1-7
\pgfplotsset{compat=1.17}
\usepackage{pgffor}

\usepackage{tocloft} % For customizing ToC
\usepackage{bm}
\usepackage{makecell} % Add this to your preamble

\usepackage{booktabs}
\usepackage{graphicx}

\usepackage{tikz}
\newcommand{\blackcircle}[1]{%
  \tikz[baseline=(char.base)]{
    \node[shape=circle,fill=black,inner sep=1.0pt] (char) {\textcolor{white}{\textbf{#1}}};
  }
}

\usepackage{mathrsfs}
\usepackage[most]{tcolorbox} 
\usepackage{booktabs, tabularx} % Add in your preamble
\usepackage{fontawesome5}

\usepackage{booktabs, bigdelim, rotating}

\usepackage{tcolorbox, xcolor}
\tcbuselibrary{listings, skins, breakable}

\usepackage{etoolbox} % Make macro writing/modifying possible
\usepackage{paracol}

\usepackage{pgfplots}
\usepackage{filecontents}
\usepackage{pgfplots}
\usepackage{pgfplotstable}
\usepackage{filecontents}
\usepgfplotslibrary{groupplots}
\usepackage{tikz}
\usepackage{pgfplots}
\usepackage{filecontents}

\usepackage{blindtext}
\usepackage{titlesec}

\usepackage{svg}
\usepackage{amsmath}
\usepackage{booktabs}
\usepackage{tikz} % For TikZ
\pgfplotsset{compat=1.18}
\usetikzlibrary{patterns}

\usepackage[T1]{fontenc} % use 8-bit T1 fonts
\usepackage{subcaption}
\usepackage{float}
\usepackage{stmaryrd}
\usepackage{dsfont}

\usepackage{bbm}
\usepackage[table]{xcolor}
\usepackage{amsfonts} % blackboard math symbols
\usepackage{nicefrac} % compact symbols for 1/2, etc.

\usepackage{color}
\usepackage{tabularray}
\definecolor{Ebb}{rgb}{0.917,0.901,0.901}

\definecolor{LightBlue}{rgb}{0.88,1,1}
\definecolor{LightGreen}{rgb}{0.88,1,0.88}
\definecolor{LightGray}{rgb}{0.95,0.95,0.95}
\definecolor{LightOrange}{rgb}{1,0.95,0.88}

\usepackage{times} % DO NOT CHANGE THIS
\usepackage{helvet} % DO NOT CHANGE THIS
\usepackage{courier} % DO NOT CHANGE THIS
\usepackage{colortbl}
\usepackage{booktabs}
\usepackage{multirow}
\usepackage{array}
\usepackage{colortbl}

\usepackage{graphicx} % DO NOT CHANGE THIS

\usepackage{natbib} % DO NOT CHANGE THIS AND DO NOT ADD ANY OPTIONS TO IT
\usepackage{caption} % DO NOT CHANGE THIS AND DO NOT ADD ANY OPTIONS TO IT
\usepackage{algorithm}
\usepackage{algorithmic}

\usepackage{graphicx} % for minipage+figure workaround
\usepackage{amsmath,amssymb}

\usepackage{newfloat}
\usepackage{listings}

\DeclareCaptionStyle{ruled}{labelfont=normalfont,labelsep=colon,strut=off} % DO NOT CHANGE THIS
\floatstyle{ruled}
\newfloat{listing}{tb}{lst}{}
\floatname{listing}{Listing}
\usepackage{hyperref}
\usepackage{amssymb}
\usepackage{amsthm}
\usepackage{amsmath}
\usepackage{enumitem}
\usepackage{graphicx}
\usepackage{subcaption}
\usepackage{mathtools}
\usepackage{multirow}
\usepackage{adjustbox}
\usepackage{relsize}
\usepackage[dvipsnames]{xcolor} % For names like RoyalBlue, ForestGreen

\usepackage[most]{tcolorbox}
\usepackage{float} % For [H] placement
\usepackage{mathtools}

\title{\papertitle }

\author{%
\textbf{Prateek Chanda}\thanks{These authors contributed equally.} \quad
\textbf{Prayas Agrawal}\footnotemark[1] \quad
\textbf{Saral Sureka} \\
\textbf{Lokesh Reddy Polu} \quad
\textbf{Atharv Kshirsagar} \quad
\textbf{Ganesh Ramakrishnan} \\
[1ex]
Department of Computer Science and Engineering, \\
Indian Institute of Technology Bombay \\
\texttt{\{prateekch, prayas, ssaral}\\
\texttt{\{lokeshreddypolu, atharvksagar, ganesh\}@cse.iitb.ac.in}
}

\usepackage{bibentry}
\usepackage{siunitx}
\usepackage{appendix}
\usepackage{minitoc}
\def \papertitle{Bandit Guided Submodular Curriculum for Adaptive Subset Selection}

\def\maxop{\mathop{\rm max}\limits} %max operator

\tcolorboxenvironment{theorem}{
  colback=!5!white,
 boxrule=0pt,
  boxsep=2pt,
left=2pt,right=2pt,top=2pt,bottom=2pt,
  oversize=2pt,
  after skip=\topsep,
    colframe=black!75!white,
  enhanced,
}

\tcolorboxenvironment{cor}{
  colback=!5!white,
  boxrule=1pt,
  boxsep=2pt,
left=2pt,right=2pt,top=2pt,bottom=2pt,
  oversize=2pt,
  before skip=\topsep,
  after skip=\topsep,
}

\newtheorem{definition}{Definition}

\newtheorem*{colorox*}{}
\newtheorem*{coloroxyellow*}{}

\tcolorboxenvironment{colorox}{
  colback=!5!white,
  boxrule=0pt,
left=-20pt,
  oversize=2pt,  
  before skip=\topsep,
  after skip=\topsep,
}

\tcolorboxenvironment{coloroxyellow}{
  colback=!1!white,
  boxrule=0pt,
  boxsep=2pt,
left=-20pt,
  oversize=2pt,  
  before skip=\topsep,
  after skip=\topsep,
}

\newcommand{\model}[1]{Online-SubMod}

\tcolorboxenvironment{lemm}{
  colback=!5!white,
  boxrule=0pt,
  boxsep=2pt,
left=2pt,right=2pt,top=2pt,bottom=2pt,
  oversize=2pt,  
  before skip=\topsep,
  after skip=\topsep,
}

\tcolorboxenvironment{lemma}{
  colback=!5!white,
  boxrule=0pt,
  boxsep=2pt,
left=2pt,right=2pt,top=2pt,bottom=2pt,
  oversize=2pt,  
  before skip=\topsep,
  after skip=\topsep,
}

\tcolorboxenvironment{restatable}{
 colback=gray!15, colframe=gray!15,
  left=2mm, right=2mm, top=1mm, bottom=1mm,
  boxsep=0pt, width=1\linewidth,  before skip=2pt, after skip=2pt, 
    enlarge left by=0mm, enlarge right by=0mm,breakable
}

\tcolorboxenvironment{proof}{
 colback=gray!5, colframe=gray!15,
  left=2mm, right=2mm, top=1mm, bottom=1mm,
  boxsep=0pt, width=1\linewidth,  before skip=2pt, after skip=2pt, 
    enlarge left by=0mm, enlarge right by=0mm,breakable
}

\newcommand{\methodprop}{\textsc{OnlineSubmod}}
\newcommand{\baselinecraig}{\textsc{Craig}}
\newcommand{\baselineglister}{\textsc{Glister}}
\newcommand{\baselinegradnorm}{\textsc{GradNorm}}
\newcommand{\baselinesbert}{\textsc{SBert}}
\newcommand{\baselinemaxloss}{\textsc{Max-Loss}}
\newcommand{\baselinegreats}{\textsc{Greats}}
\newcommand{\baselinerho}{\textsc{Rho-Loss}}
\newcommand{\baselineboss}{\textsc{BOSS}}
\newcommand{\baselinemilo}{\textsc{MILO}}
\newcommand{\baselineGradmatch}{\textsc{GradMatch}}

\newcommand{\DPruning}{\textsc{D2Pruning}}
\newcommand{\InfoMax}{\textsc{InfoMax}}
\newcommand{\CCS}{\textsc{CCS}}

\newtheorem{thm}{theorem}

\newtheorem{defn}[thm]{Definition}

\DeclareMathAlphabet{\pazocal}{OMS}{zplm}{m}{n}

\def\BibTeX{{\rm B\kern-.05em{\sc i\kern-.025em b}\kern-.08em
    T\kern-.1667em\lower.7ex\hbox{E}\kern-.125emX}}

\definecolor{xblue}{HTML}{4169E1}
\definecolor{xbluelight}{HTML}{6287F5}
\definecolor{xbluedark}{HTML}{3C56A6}
\definecolor{xgreen}{HTML}{036C3A}
\definecolor{xpurple}{HTML}{9838B1}
\definecolor{xgray}{HTML}{808080}
\definecolor{xsienna}{HTML}{8B4512}
\definecolor{xslategray}{HTML}{70818F}
\definecolor{xred}{HTML}{FF2121}
\definecolor{xorange}{HTML}{FF8C00}
\definecolor{xcyan}{HTML}{06AEEF}
\definecolor{xlightcyan}{HTML}{CEEFFC}
\definecolor{xolive}{HTML}{556B2F}
\definecolor{xxorange}{HTML}{FF8200}
\definecolor{xxgreen}{HTML}{009F86}
\definecolor{xxpurple}{HTML}{623E99}
\newcommand{\xblue}[1]{\textcolor{xblue}{#1}}
\newcommand{\xbluelight}[1]{\textcolor{xbluelight}{#1}}
\newcommand{\xbluedark}[1]{\textcolor{xbluedark}{#1}}

\newcommand{\xsienna}[1]{\textcolor{xsienna}{#1}}

\newcommand{\xxgreen}[1]{\textcolor{xxgreen}{#1}}
\newcommand{\xxpurple}[1]{\textcolor{xxpurple}{#1}}

\newcommand{\bD}[1]{\boldsymbol{\mathcal{D}}}

\newcommand{\method}[1]{CoreSetPFedBayes}

\newcommand{\thicktilde}[1]{\mathbf{\tilde{\text{$#1$}}}}

\DeclareMathAlphabet{\pazocal}{OMS}{zplm}{m}{n}

\def\BibTeX{{\rm B\kern-.05em{\sc i\kern-.025em b}\kern-.08em
    T\kern-.1667em\lower.7ex\hbox{E}\kern-.125emX}}

\def\A{\mathcal{A}}

\def\K{\mathcal{K}}

\def\B{\mathcal{B}}
\def\U{\mathcal{U}}

\def\det{\mathop{\rm det}\nolimits}

\def\bof{{\boldsymbol f}}

\def\bx{{\boldsymbol x}}

\def\bz{{\boldsymbol z}}

\def\bD{\mathbf D}

\def\bTheta{\boldsymbol{\Theta}}

\def\btheta{\boldsymbol{\theta}}
\def\bDelta{\boldsymbol{\Delta}}

\def\bXi{\boldsymbol{\Xi}}

\def\maxop{\mathop{\rm max}\limits} %max operator

\newcommand{\eat}[1]{}

\def\S{\mathcal{S}}

\begin{document}
    \maketitle

\begin{abstract}
Traditional curriculum learning proceeds from easy to hard samples, yet defining a reliable notion of difficulty remains elusive. Prior work has used submodular functions to induce difficulty scores in curriculum learning. We reinterpret adaptive subset selection and formulate it as a multi-armed bandit problem, where each arm corresponds to a submodular function guiding sample selection. We introduce $\methodprop$, a novel online greedy policy that optimizes a utility-driven reward and provably achieves no-regret performance under various sampling regimes. Empirically, $\methodprop$ outperforms both traditional curriculum learning and bi-level optimization approaches across \textbf{vision} and \textbf{language datasets}, showing superior accuracy-efficiency tradeoffs. More broadly, we show that validation-driven reward metrics offer a principled way to guide the curriculum schedule. Our code is publicly available at GitHub \footnote{\url{https://github.com/efficiency-learning/banditsubmod/}}.
\end{abstract}

\iffalse 
\begin{abstract}
        Traditional curriculum learning progresses from easy to hard samples,
        but defining a reliable criterion for sample difficulty remains
        challenging. Previous methods have shown promise using submodular functions as a measure of difficulty for curriculum learning settings. We reformulate adaptive subset selection as an online bandit
        problem, where $\K$ submodular functions act as bandit arms to optimize
        a utility-driven reward function, enhancing validation performance. We
        introduce a novel greedy utility metric as a policy and provide no-regret
        guarantees across various importance sampling configurations. Experiments
        demonstrate that our method, $\methodprop$, effectively utilizes bandit feedback
        to optimize submodular functions, surpassing traditional curriculum
        learning and bi-level optimization techniques in accuracy-efficiency
        trade-offs across diverse datasets. Our experiments on both \textbf{vision datasets} and \textbf{language datasets} demonstrate superior performance w.r.t traditinal baselines for a variety of subset selection fractions. More broadly our work shows that a more principled validation performance driven reward utility can guide the curriculum learning sequence. 
    \end{abstract}
\fi
    \addtocontents{toc}{\protect\setcounter{tocdepth}{-1}}
    \section{Introduction}
  % Temporarily suppress ToC entries

    Curriculum Learning (CL), inspired by cognitive development, posits that
    training machine learning models by gradually exposing them to data of
    increasing complexity can significantly enhance both learning efficiency and
    generalization performance \citep{bengio2009curriculum, zhoucurbench}. The
    underlying principle is that mastering simpler concepts first provides a
    robust foundation for acquiring more complex ones, leading to improved convergence
    and a more effective exploration of the hypothesis space \citep{killamsetty2023milo}.
   Empirical evidence shows CL improves model training, particularly in areas like code understanding \citep{nair2024curriculum}, enhances graph embeddings through complexity-based ordering \citep{zhang2024curriculum}, mitigates catastrophic forgetting \citep{aljundi2019task, lopez2017gradient, van2019three, shi2024continual}, and boosts learning efficiency in reinforcement learning \citep{narvekar2020curriculum}.  We first provide a formal definition of Curriculum Learning.
  
\begin{defn}(\xsienna{\emph{\textbf{Curriculum Learning}}})
Given a dataset \( \mathcal{D} = \bigcup_{i=1}^k \mathcal{B}_i \)  partitioned into disjoint batches \( \mathcal{B}_i \),  and a batch difficulty score function \( \boldsymbol{d}: \{\mathcal{B}_i\}_{i=1}^k \to \mathbb{R}_{\geq 0} \) assigning non-negative difficulty scores, a \textbf{batch-wise curriculum} can be represented as a permutation $\pi: [k] \mapsto [k]$ over the ordered indices such that the ordered sequence 
\[
\mathcal{C} = (\mathcal{B}_{\pi(1)}, \mathcal{B}_{\pi(2)}, \dots, \mathcal{B}_{\pi(k)}),
\]
satisfies the \textbf{monotonic difficulty score condition}: $\boldsymbol{d}(\B_{\pi(t)}) \leq \boldsymbol{d}(\B_{\pi(t+1)}) \quad \forall t \in \{1,\dots,k-1\}$.
\iffalse 
Given a dataset \( \mathcal{D} = \bigcup_{i=1}^k \mathcal{B}_i \) partitioned into disjoint batches \( \mathcal{B}_i \), define a batch difficulty function \( \boldsymbol{d}: \{\mathcal{B}_i\}_{i=1}^k \to \mathbb{R}_{\geq 0} \) assigning each batch a non-negative score. Let \( \pi: \{1,\dots,k\} \to \{1,\dots,k\} \) be a permutation of indices specifying a processing order. A \textbf{batch-wise curriculum} is the ordered sequence:
\[
\mathcal{C} = (\mathcal{B}_{\pi(1)}, \mathcal{B}_{\pi(2)}, \dots, \mathcal{B}_{\pi(k)}),
\]
satisfying the \textbf{monotonic difficulty condition}:$\boldsymbol{d}(\mathcal{B}_{\pi(t)}) \leq \boldsymbol{d}(\mathcal{B}_{\pi(t+1)}) \quad \forall t \in \{1,\dots,k{-}1\}.$  
\fi
\end{defn}

\textbf{Determining Difficulty is challenging} A critical challenge in realizing the full potential of Curriculum Learning (CL) is determining the optimal sequence of batches. This is complicated by the fact that the difficulty score, denoted as $\boldsymbol{d}$, is typically unknown. Traditional approaches often rely on domain expertise or practitioner's knowledge to assess the hardness or difficulty of samples.

Recent works, such as \citep{killamsetty2023milo}, have proposed using submodular function maximization over data batches as an intrinsic measure of sample difficulty. In particular, \textbf{ \xsienna{representative submodular functions}} representative submodular functions are used to identify easy samples, while \textbf{\xxpurple{diversity focused submodular functions}} are used to capture difficult ones. As a result, the CL objective is typically constructed by prioritizing diversity functions later and  representative functions earlier in the training phase. However, this definition of hardness is still restrictive, as it relies on a fixed pretraining phase and does not account for evolving training dynamics.

\noindent
\begin{minipage}{0.47\linewidth}
    \centering
    \vspace{0pt} % force top alignment
    \resizebox{\linewidth}{!}{

\begin{tikzpicture}
\begin{groupplot}[
        group style={
            group size=2 by 1,
            horizontal sep=2cm,
        },
    tick label style={font=\large},    % numbers on axes
    label style={font=\large},         % axis labels
    title style={font=\large},         % plot titles
    legend style={font=\large}   ,
    axis lines=box, % Ensures the axes have a full box border
       % thick, % Makes the plot 
         axis background/.style={ % Define the gradient background
                shade, 
                left color=blue!4, 
                right color=white
            },
        %lines thicker,
        width=8cm,
        height=6cm,
        xlabel style={font=\Large}, % Specifically enlarge x-axis label font
        ylabel style={font=\Large},
        xlabel={Epoch},
        grid=major,
        legend style={at={(0.5, -0.1)}, anchor=north, legend columns=1}, % Position the legend inside
        ylabel={Test Accuracy (\%)},
         legend style={font=\small},
        cycle list name=color list
    ]
    \nextgroupplot[
        align=center,
        title style={font=\Large\bfseries},
        title={\textbf{Submodular CL per step}},
        legend pos= south east
    ]
\addplot+[
    mark=*,
    mark size=2.5pt,
    thick,
    color=red,
    mark options={fill=red!20, draw=red!80},
each nth point=7, filter discard warning=false, unbounded coords=discard 
]
table [x index=0, y index=1, col sep=space] {data1.dat};
    \addlegendentry{\textbf{50\% div} $\mapsto$ \textbf{50\% repr}}

    \addplot+[mark=triangle*,mark size=2.5pt, thick,  color=blue,     mark options={
        fill=blue!20, % fill color
        draw=blue!80  % marker boundary color (sharper blue)
    },
    each nth point=7, filter discard warning=false, unbounded coords=discard
    ] table [x index=0, y index=1, col sep=space] {data2.dat};
    \addlegendentry{\textbf{50\% repr} $\mapsto$ \textbf{50\% div}}

    \nextgroupplot[
    align=center,
        title style={font=\Large\bfseries},
        title={Submodular CL per Epoch},
        legend pos= south east
    ]
    \addplot+[mark=*,
    thick,
        mark size=2.5pt, % Increase marker size (you can adjust the value as needed)
    color=red,
    mark options={fill=red!20, draw=red!80},
each nth point=7, filter discard warning=false, unbounded coords=discard ] table [x index=0, y index=1, col sep=space] {data3.dat};

% Div interval: dashed red line with label
\draw[thick, red, dashed] (axis cs:0,0.2) -- (axis cs:60,0.2);        % Left segment
\draw[thick, red, dashed] (axis cs:90,0.2) -- (axis cs:150,0.2);      % Right segment
% Left bar with a vertical extension at x = 0
\draw[thick, red] (axis cs:0,0.2) -- ++(0,0.025);              % Original left bar
\draw[thick, red] (axis cs:0,0.2) -- ++(0,-0.015);             % New vertical extension down to 0.015
\draw[thick, red] (axis cs:0,0.2) -- ++(0,0.027);              % New vertical extension up to 0.025

              % Right bar
\node[red, font=\large\bfseries] at (axis cs:75,0.225) {\textbf{Div}};           % Label for Div

% Repr interval: dashed red line with label
\draw[thick, red, dashed] (axis cs:150,0.2) -- (axis cs:210,0.2);     % Left segment
\draw[thick, red, dashed] (axis cs:240,0.2) -- (axis cs:300,0.2);     % Right segment
% Left bar at x = 150 with vertical extension
\draw[thick, red] (axis cs:150,0.2) -- ++(0,0.025);              % Original left bar at x = 150
\draw[thick, red] (axis cs:150,0.2) -- ++(0,-0.015);             % Vertical extension down to 0.015
\draw[thick, red] (axis cs:150,0.2) -- ++(0,0.025);              % Vertical extension up to 0.025

% Left bar at x = 300 with vertical extension
\draw[thick, red] (axis cs:300,0.2) -- ++(0,0.025);              % Original left bar at x = 300
\draw[thick, red] (axis cs:300,0.2) -- ++(0,-0.015);             % Vertical extension down to 0.015
\draw[thick, red] (axis cs:300,0.2) -- ++(0,0.025);              % Vertical extension up to 0.025
               % Right bar
\node[red, font=\large\bfseries] at (axis cs:225,0.225) {\textbf{Repr}};         % Label for Repr

% Horizontal line with vertical bars and text "Repr"

% Repr interval: dashed blue line with label

% Repr interval: dashed blue line with label
\draw[thick, blue, dashed] (axis cs:0,0.07) -- (axis cs:60,0.07);        % Left segment
\draw[thick, blue, dashed] (axis cs:90,0.07) -- (axis cs:150,0.07);      % Right segment
\draw[thick, blue] (axis cs:0,0.07) -- ++(0,0.025);                      % Left bar
\draw[thick, blue] (axis cs:150,0.07) -- ++(0,0.025);                    % Right bar
\draw[thick, blue] (axis cs:0,0.07) -- ++(0,-0.015);                     % Vertical extension down at x = 0
\draw[thick, blue] (axis cs:150,0.07) -- ++(0,-0.015);                   % Vertical extension down at x = 150
\draw[thick, blue] (axis cs:150,0.07) -- ++(0,0.025);                    % Vertical extension up at x = 150
\draw[thick, blue] (axis cs:300,0.07) -- ++(0,0.025);                    % Vertical extension up at x = 300

\node[blue, font=\large\bfseries] at (axis cs:75,0.075) {\textbf{Repr}};          % Label for Repr

% Div interval: dashed blue line with label
\draw[thick, blue, dashed] (axis cs:150,0.07) -- (axis cs:210,0.07);     % Left segment
\draw[thick, blue, dashed] (axis cs:240,0.07) -- (axis cs:300,0.07);     % Right segment
\draw[thick, blue] (axis cs:150,0.07) -- ++(0,0.025);                    % Left bar
\draw[thick, blue] (axis cs:300,0.07) -- ++(0,0.025);                    % Right bar
\draw[thick, blue] (axis cs:150,0.07) -- ++(0,-0.015);                   % Vertical extension down at x = 150
\draw[thick, blue] (axis cs:300,0.07) -- ++(0,-0.015);                   % Vertical extension down at x = 300

\node[blue, font=\large\bfseries] at (axis cs:225,0.075) {\textbf{Div}};           % Label for Div

    \addplot+[mark=triangle*,
        mark size=2.5pt, % Increase marker size (you can adjust the value as needed)
        thick,  
        color=blue,     mark options={
        fill=blue!20, % fill color
        draw=blue!80  % marker boundary color (sharper blue)
    },
    each nth point=7, filter discard warning=false, unbounded coords=discard] table [x index=0, y index=1, col sep=space] {data4.dat};
\addlegendimage{empty legend} % Ensures no legend is added

    \end{groupplot}
\end{tikzpicture}
}

    
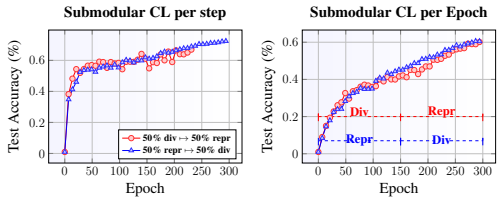
\captionof{figure}{\textbf{Sequential Ordering of Submodular Functions}: 
    \emph{Observations on CIFAR100}. Initial training with subsets sampled 
    using representation-based submodular functions followed by diversity 
    performs better than the opposite order.}
    \label{Fig:introplug}
\end{minipage}
\hfill
\begin{minipage}{0.50\linewidth}
    \vspace{0pt} % aligns text to top
    % Your text starts normally here:
    \textbf{Adaptive Subset Selection Induces CL} Many adaptive subset selection methods although can be viewed as forms of curriculum learning, incur substantial computational overhead. For instance, Glister \citep{killamsetty2021glister} solves costly bilevel optimization involving joint subset selection and model training with validation feedback. GradMatch \citep{killamsetty2021grad} minimizes gradient matching error by solving complex optimization problems to approximate full-dataset gradients. Importance sampling approaches \citep{calandriello2020sampling, sujit2023prioritizing} similarly require expensive importance score estimations. Such costs limit the scalability of advanced curriculum strategies, especially under resource constraints or large datasets.
\end{minipage}

\subsection{Our Contributions}

\textbf{Submodular curriculum learning via online bandits} 
We formulate the curriculum learning problem in conjunction with the adaptive subset selection as a multi-arm bandit problem, where each arm corresponds to a submodular function that captures its unique characteristics, thereby providing a good surrogate difficulty score required for curriculum learning design.

\textbf{A no-regret greedy policy for adaptive subset selection} 
We introduce $\methodprop$, a novel greedy utility-based policy that leverages feedback from  validation performance-driven reward signal to adaptively guide the subset selection process. We prove that $\methodprop$ achieves no-regret performance under general sampling regimes, providing theoretical grounding for its learning efficiency.

\textbf{Validation performance-aware reward design} 
Unlike prior work which uses static heuristics or model-dependent metrics, we define a utility function based on validation performance-driven reward improvements, thereby aligning curriculum progression with actual generalization objectives. 
\textbf{Empirical improvements across modalities} 
Through extensive experiments on large-scale language and vision benchmarks, we demonstrate that $\methodprop$ outperforms traditional curriculum strategies and state-of-the-art adaptive selection methods in terms of accuracy-efficiency trade-offs across diverse subset budgets and training stages.

% \vspace{-8pt}
\subsection{Brief Discussion on Related Work \& Limitations}

Here we detail some of the recent prior work in the space of adaptive subset selection and corresponding limitations.

\textbf{Leveraging Training Gradient information}: Efficiently training robust  machine learning models often involves selecting informative data subsets.  $\baselineglister$ \citep{killamsetty2021glister} directly addresses this through a mixed
    discrete-continuous bi-level optimization framework, leveraging validation
    likelihood for robustness. The concept of \emph{adaptive data subset
    selection}, where the subset evolves during training, is explored by
    methodologies like coreset selection \citep{mirzasoleiman2020coresets}.
    $\baselineGradmatch$ \citep{killamsetty2021grad} tackles the problem by focusing on minimizing
    \emph{gradient matching error}, as the quality of this matching significantly
    impacts convergence. By modeling this error as weakly submodular and using
    OMP \citep{elenberg2018restricted}, GradMatch achieves tighter convergence
    guarantees for various convex loss functions. Despite their advancements,
    many contemporary subset selection techniques, such as coreset selection and
    related methods~\citep{chandabayesian}, pose a considerable computational burden
    due to their complex optimization processes.

    \textbf{Reweighting Techniques}: In this context, \citep{jiang2017submodular}
    offered significant insights into strategies for selecting data subsets that
    focus on identifying high-quality subsets during the training of models. As we
    shift towards meta-learning and weighted loss techniques, traditional
    methods like importance sampling, first introduced by \citep{kahn1953estimation},
    and more contemporary approaches such as focal loss proposed by \citep{lin2017focal},
    provide essential perspectives on weighting samples to highlight more challenging
    examples during training. However, all these strategies entail additional costs.
    We further share a more detailed Related work section in Appendix \ref{related work}.

\section{Notation and Problem Setup}
\vspace{-3pt}
\textbf{Notation}: We consider a supervised learning setup where
we have a training dataset    $\boldsymbol{\mathcal{D}}_{\mathsf{tr}}= \{(\mathbf{x}_{i}, y_{i})\}_{i=1}^{n}$,
    with each instance independently and identically distributed (i.i.d.)
    according to a distribution $\mathbb{P}_{\mathcal{X} \times \mathcal{Y}}$
    over the feature space $\mathcal{X}$ and label space $\mathcal{Y}$. Similarly,
    we have a validation dataset
    $\boldsymbol{\mathcal{D}}_{\texttt{val}}= \{(\mathbf{x}_{j}^{\texttt{val}}, y
    _{j}^{\texttt{val}})\}_{j=1}^{m}$, also drawn i.i.d. from
    $\mathbb{P}_{\mathcal{X} \times \mathcal{Y}}$. Here,
    $\mathbf{x}\in \mathcal{X}$ represents the features and $y \in \mathcal{Y}$
    represents the labels. Let $\mathcal{M}_{\boldsymbol{\theta}}$ be a model parameterized
    by $\boldsymbol{\theta}\in \boldsymbol{\Theta}\subseteq \mathbb{R}^{\boldsymbol{d}}$,
    with $\boldsymbol{\Theta}$ being a compact and convex parameter space. The learning
    objective is to minimize the empirical risk $L(\mathcal{M}_{\boldsymbol{\theta}}
    ; \boldsymbol{\mathcal{D}}_{\mathsf{tr}})$ \citep{vapnik1991principles}.
    The training process unfolds over a discrete time horizon $T \in \mathbb{Z}^{+}$.  Let $\boldsymbol{\mathcal{F}}$ be the space of set functions, with
$\boldsymbol{\mathcal{F}}_{\text{sub}}\subset \boldsymbol{\mathcal{F}}$
    denoting the subspace of submodular functions.      
    
    \textit{Note:} Throughout this paper, we use \( \mathbf{z} \) to denote a training instance from \( \mathcal{B}_t \) , unless explicitly labeled as \( \mathbf{z}_{\text{val}} \), which refers to a validation instance. In Appendix Section \ref{Supplementary:Notation Summary} we provide an extensive notation summary. We provide here some important definitions which would be utilised in the later sections.
    \begin{defn}
        [\textbf{Submodularity}] Given a ground set $\mathcal{V}$, a set function
        $\bof: 2^{\mathcal{V}}\mapsto \mathbb{R}$ is \textit{submodular} if for all
        $\mathcal{S}\subseteq \mathcal{V}$ and $\mathcal{B}\subseteq \mathcal{A}\subseteq
        \mathcal{V}$, it holds that
        $\bof(\mathcal{S}\cup \mathcal{A}) - \bof(\mathcal{A}) \leq \bof(\mathcal{S}
        \cup \mathcal{B}) - \bof(\mathcal{B})$.
        \label{def:submodularity}
    \end{defn}

    \begin{defn}
        [\textbf{Monotonicity}] A set function $\bof: 2^{\mathcal{V}}\mapsto \mathbb{R}_{\geq 0}$ is
        \textit{monotone} if for all
        $\mathcal{B}\subseteq \mathcal{A}\subseteq \mathcal{V}$, it holds that $\bof
        (\mathcal{B}) \leq \bof(\mathcal{A})$.
    \end{defn}

    \begin{defn}
        [\textbf{Maximum High Value Subset}] \label{def:high-val-subset} Corresponding to
        a monotone submodular function $\bof$, the maximum high value subset of
        cardinality at most $\beta$, denoted by $\bof_{\arg}(\beta) = \mathcal{B}^{\texttt{opt}}
        \subseteq \mathcal{V}$, is defined as: $\mathcal{B}^{\texttt{opt}}= \underset
        {\mathcal{B} \subseteq \mathcal{V}; |\mathcal{B}| \leq \beta}{\operatorname{argmax}}
        \bof(\mathcal{B})$.
    \end{defn}

    \subsection{Problem Formulation : Adaptive Subset Selection posed as Curriculum Learning}

 At each discrete time step $t \in [T]$, we consider a  mini-batch
    $\mathcal{B}_{t}\subseteq \boldsymbol{\mathcal{D}}_{\mathsf{tr}}$ upon which
    the model $\mathcal{M}_{\boldsymbol{\theta}}$ is trained. Let $\ell : \mathcal{Z}
    \times \boldsymbol{\Theta}\mapsto \mathbb{R}$ denote the
    instance-wise loss function, where
    $\mathcal{Z}= \mathcal{X}\times \mathcal{Y}$ is the instance space, and the model
    parameter at time $t$ is denoted by $\boldsymbol{\theta}_{t}\in \boldsymbol{\Theta}$.
    The total loss over the mini-batch $\mathcal{B}_{t}$ is given by
    $\boldsymbol{\mathfrak{L}}_{t}(\boldsymbol{\theta}_{t}) = \sum_{\mathbf{z}
    \in \mathcal{B}_t}\ell(\mathbf{z}, \boldsymbol{\theta}_{t})$. Concurrently,
    we have access to a validation mini-batch
    $\mathcal{B}_{t}^{\texttt{val}}\subseteq \boldsymbol{\mathcal{D}}_{\texttt{val}}$
    at each time step $t$.

\begin{wrapfigure}{r}{0.48\textwidth}
  \begin{minipage}{0.47\textwidth}
    \centering
    \small
    \renewcommand{\arraystretch}{0.8}
    \begin{tabular}{@{} l c @{}}
      \toprule
      \textbf{Function} & $\bof(X)$ \\
      \midrule
      \multicolumn{2}{l}
    {\textbf{\xsienna{\emph{Representative}}}} \\
      \rowcolor{gray!15} Facility Location & $\sum_{i \in \mathcal{V}} \max_{j \in X} s_{ij}$ \\
      Graph Cut & $\sum_{i \in \mathcal{V}, j \in X} s_{ij} - \rho \sum_{i, j \in X} s_{ij}$\\
      \midrule
      \multicolumn{2}{l}{\textbf{\xxpurple{\emph{Diversity}}}} \\
      \rowcolor{gray!15} Log Determinant & $\log \det(\mathcal{S}_X)$ \\
      Disparity-Min & $\min_{i \ne j \in X} (1 - s_{ij})$ \\
      \rowcolor{gray!15} Disparity-Sum & $\sum_{i \ne j \in X} (1 - s_{ij})$ \\
      \bottomrule
    \end{tabular}
    \caption*{
      \textbf{Table 1:} \textit{Submodular functions used in arm definitions. $\mathcal{V}$ is the ground set, $X \subseteq \mathcal{V}$, $s_{ij}$ denotes pairwise similarity, and $\mathcal{S}_X$ is the similarity submatrix. $\rho$ indicates the balancing factor between representative and diversity nature. We also utilise mutual information variants (Details in Appendix)}
    }
  \end{minipage}
  \vspace{-6pt}
\end{wrapfigure}
\vspace{-2pt}

    \textbf{Gradient Matrix and Mean Gradient:} 
  Let $\mathbf{G}_{\btheta_t}=
    \begin{bmatrix}
        \boldsymbol{g}_{\btheta_{t}}(\mathbf{z}_{1}), &\dots, & \boldsymbol{g}_{\btheta_{t}}(\mathbf{z}_{|\mathcal{B}_t|}) \\
    \end{bmatrix}
    \in \mathbb{R}^{\boldsymbol{d} \times |\mathcal{B}_t|}$ be the batch gradient
    matrix at time step $t$, where each column
    $\boldsymbol{g}_{\btheta_{t}}(\mathbf{z}_{i}) = \nabla_{\boldsymbol{\theta}}\ell
    (\mathbf{z}_{i}, \boldsymbol{\theta}_{t}) \in \mathbb{R}^{\boldsymbol{d}}$
    is the \textbf{sample-wise gradient }of the loss function $\ell$ with respect to
    the model parameter $\boldsymbol{\theta}_{t}$, for all $\mathbf{z}_{i}\in \mathcal{B}
    _{t}$. Let
    $\mathbf{1}_{|\mathcal{B}_t|}\in \mathbb{R}^{|\mathcal{B}_t| \times 1}$
    denote the column vector of ones. We  define the \textbf{per-batch gradient} as $\bar
    {\mathbf{g}}^{(b)}_{\btheta_t}= \frac{1}{|\mathcal{B}_{t}|}\sum_{\mathbf{z}_i
    \in \mathcal{B}_t}\boldsymbol{g}_{\btheta_t}(\mathbf{z}_{i}) = \frac{1}{|\mathcal{B}_{t}|}
    \mathbf{G}_{\btheta_t}\mathbf{1}_{|\mathcal{B}_t|}$.

%\textbf{Formal Preliminaries: States, Actions, and Rewards}

%We consider a general decision-making framework over discrete time steps \( t \in [T] \), where the learner ($\M_{\btheta}$) interacts with an environment through the following components:

\textbf{Action Space and Submodular Selection Policy.}
At each time step \( t \in [T] \), the learner observes a mini-batch \( \mathcal{B}_t \subseteq \mathcal{D}_{\mathsf{tr}} \) and must select a subset of size \( \beta \) to compute a gradient update. The learner chooses an action \( a_t \in \mathscr{A} \) from a discrete action space: \(
\mathscr{A} := \left\{ \bof^{(1)}, \bof^{(2)}, \dots, \bof^{(\mathcal{K})} \right\}, \quad \bof^{(a)} \in \mathcal{F}_{\mathsf{sub}},
\)
where each \( \bof^{(a)} : 2^{\mathcal{B}_t} \rightarrow \mathbb{R} \) is a monotone submodular function used to score subsets of \( \mathcal{B}_t \). These functions encode different sample selection criteria such as diversity, coverage, and representativeness (see Table 1 for examples). The selected function \( \bof^{(a_t)} \) is then approximately maximized over \( \mathcal{B}_t \) under a fixed cardinality constraint to produce a training subset:
\(
\mathcal{S}_t := \arg\max_{S \subseteq \mathcal{B}_t,\ |S| \leq \beta} \bof^{(a_t)}(S),
\)
which is typically computed via a greedy algorithm. The model is updated using \( \mathcal{S}_t \), and the quality of the update is evaluated using a utility-based reward defined on a held-out validation mini-batch \( \mathcal{B}_t^{\mathsf{val}} \subseteq \mathcal{D}_{\mathsf{val}} \).

Specifically, let \(\boldsymbol{\vartheta}(a \mid \B_t) \) be the empirical estimate of the expected reward for arm \( a \in \mathscr{A} \).

\textbf{Policy: Greedy Deterministic Selection.}
We adopt a greedy deterministic policy \( \pi : 2^{\mathcal{D}_{\mathsf{tr}}} \rightarrow \mathscr{A} \) that selects the arm with the highest estimated reward at each time step i.e.
\(
a_t := \pi(\mathcal{B}_t) := \arg\max_{a \in \mathscr{A}} \boldsymbol{\vartheta}(a_t \mid \B_t).
\). where \( \mathcal{U}_t \) is the utility function defined in Section~\ref{ref:Section:UtilityMetric}.

\textbf{Regret as a Performance Measure} We denote by \((\bm{\ast})\) the index of an optimal action, so that \(\mu_{(\boldsymbol\ast)}(\mathcal{B}_t)\) represents the expected utility (e.g., value of the selected subset) of an optimal submodular function $\bof^{(a^\ast_t)}$ when applied to mini-batch \(\mathcal{B}_t\). For each action \(a_t \in \mathscr{A}\), we define the \textit{optimality gap} at time \(t\) as $\bDelta_{(a_t)}(\mathcal{B}_t) := \max\left\{0,\ \boldsymbol{\vartheta}(a^\ast_t \mid \B_t) - \boldsymbol{\vartheta}(a_t \mid \B_t)\right\}$. The \textit{cumulative regret} after \(T\) rounds is then defined as
\(
\mathsf{Regret}_{T} := \sum_{t=1}^{T} \bDelta_{a_t}(\mathcal{B}_t),
\).
Minimizing \( \mathsf{Regret}_T \) ensures that the learner approaches the performance of the best submodular selector in hindsight. We define $\boldsymbol{\vartheta}(\bullet \mid \B_t)$ in Sec \ref{section:samplewise_gain} 

\textbf{Reward Utility Metric for Performance Evaluation}\label{ref:Section:UtilityMetric}
Drawing upon the concept of training data influence \citep{pruthi2020estimating},
    we define a utility function $\U_{t}(\mathcal{B}_{t}, \mathbf{z}_{\mathsf{val}}
    ) : 2^{\mathcal{D}_{\mathsf{tr}}}\times \mathcal{D}_{\mathsf{val}}\mapsto \mathbb{R}$
    to quantify the impact of a training mini-batch $\mathcal{B}_{t}\subseteq \mathcal{D}
    _{\mathsf{tr}}$ at time step $t$ on a validation instance $\mathbf{z}_{\mathsf{val}}
    \in \mathcal{B}_{t}^{\mathsf{val}}$. Specifically, the utility is the reduction
    in the loss on the validation instance after one step of stochastic gradient
    descent:
    \vspace{-5pt}
    \begin{equation}
        \U_{t}(\mathcal{B}_{t}, \mathbf{z}_{\mathsf{val}}) = \ell(\mathbf{z}_{\mathsf{val}}
        , \boldsymbol{\theta}_{t}) - \ell(\mathbf{z}_{\mathsf{val}}, \Tilde{\boldsymbol{\theta}}
        _{t+1}(\mathcal{B}_{t})), \label{Eq:utility Metric}
    \end{equation}
    where the updated parameter vector
    $\Tilde{\boldsymbol{\theta}}_{t+1}(\mathcal{B}_{t}) = \boldsymbol{\theta}_{t}- \eta_{t}
    \boldsymbol\nabla_{\boldsymbol{\theta}}\left( \frac{1}{|\mathcal{B}_{t}|}\sum_{\mathbf{z}
    \in \mathcal{B}_t}\ell(\mathbf{z}, \boldsymbol{\theta}_{t}) \right
    ).$

   % To evaluate the overall utility of the training mini-batch $\mathcal{B}_{t}$
    %with respect to the entire validation mini-batch $\mathcal{B}_{t}^{\mathsf{val}}$,
    %we compute the average of the utility
    %$\U_{t}( \mathcal{B}_{t}, \mathbf{z}_{\mathsf{val}})$ over all validation
    %instances $\mathbf{z}_{\mathsf{val}}\in \mathcal{B}_{t}^{\mathsf{val}}$. This
   % average utility estimates the expected decrease in the loss on a randomly chosen
    %validation instance from $\mathcal{B}_{t}^{\mathsf{val}}$ after one training
%    step on $\mathcal{B}_{t}$.

\begin{comment}
    \\
        & = \ell( \mathbf{z}_{\mathsf{val}}, \boldsymbol{\theta}_{t+1}(\mathcal{B}_{t}^{(<i)})) - \ell(\mathbf{z}_{\mathsf{val}},\boldsymbol{\theta}_{t+1}(\mathcal{B}_{t}^{(<i)}\cup \{\mathbf{z}_{i}\})) \\
   & = \ell( \mathbf{z}_{\mathsf{val}}, \boldsymbol{\theta}_{t+1}(\mathcal{B}_{t}^{(<i)})) - \ell(\mathbf{z}_{\mathsf{val}},\boldsymbol{\theta}_{t+1}(\mathcal{B}_{t}^{(<i)}) - \eta_{t}\boldsymbol{g}_{\boldsymbol{\theta}_{t}}(\mathbf{z}_{i}))                            \\
\end{comment}

\textbf{First-Order Approximation of Marginal Utility Gain}: We define the instance-wise conditional marginal utility gain of including
    the $i$-th training instance $\mathbf{z}_{i}$ into a partially constructed
    mini-batch
    $\mathcal{B}_{t}^{(<i)}= \{\mathbf{z}_{1}, \mathbf{z}_{2}, \dots, \mathbf{z}_{i-1}
    \}$
    at time step $t$, with respect to a validation instance $\mathbf{z}_{\mathsf{val}}$,
    as the change in utility $\mathcal{U}_{t}$:
    \begin{align}
     \bDelta \mathcal{U}_{t}(\mathbf{z}_{i}\mid \mathcal{B}_{t}^{(<i)}, \mathbf{z}
        _{\mathsf{val}}) & = \mathcal{U}_{t}(\mathcal{B}_{t}^{(<i)}\cup \{\mathbf{z}
        _{i}\}; \mathbf{z}_{\mathsf{val}}) - \mathcal{U}_{t}(\mathcal{B}_{t}^{(<i)}
        ; \mathbf{z}_{\mathsf{val}})\\
     & \approx \eta_{t}\boldsymbol{\nabla}_{\boldsymbol{\theta}}\ell(\mathbf{z}_{i}, \boldsymbol{\theta}_{t}) \cdot\boldsymbol{\nabla}_{\boldsymbol{\theta}}\ell(\mathbf{z}_{\mathsf{val}}, \boldsymbol{\theta}_{t+1}(\mathcal{B}_{t}^{(<i)})) 
\label{eq:FirstApproximation_start}
    \end{align}

The approximation in the last step utilizes a first-order Taylor expansion, which
    is reasonable under the common assumption of a small learning rate $\eta_{t}$. We defer the derivation to Appendix

    \textbf{Second-Order Approximation and Gradient Influence}:  Further approximating the second term in Equation \eqref{eq:FirstApproximation_start}
    using another first-order Taylor expansion around $\boldsymbol{\theta}_{t}$,
    we obtain:
\setlength{\jot}{2pt} % adjust this value as needed; try 1pt or 0pt for tighter spacing
\begin{align}
& \eta_{t}\boldsymbol{g}_{\boldsymbol{\theta}_{t}}(\mathbf{z}_{i}) \cdot \boldsymbol{\nabla}_{\boldsymbol{\theta}}\ell(\mathbf{z}_{\mathsf{val}}, \boldsymbol{\theta}_{t+1}(\mathcal{B}_{t}^{(<i)})) \nonumber \approx \eta_{t}\boldsymbol{g}_{\boldsymbol{\theta}_{t}}(\mathbf{z}_{i}) \cdot \boldsymbol{\nabla}_{\boldsymbol{\theta}}\ell(\mathbf{z}_{\mathsf{val}}, \boldsymbol{\theta}_{t}- \eta_{t}\frac{1}{|\mathcal{B}_{t}^{(<i)}|}\sum_{\mathbf{z} \in \mathcal{B}_{t}^{(<i)}}\boldsymbol{g}_{\boldsymbol{\theta}_{t}}(\mathbf{z})) \nonumber \\
& \approx \eta_{t}\underbrace{\boldsymbol{g}_{\boldsymbol{\theta}_t}(\mathbf{z}_i) \cdot \boldsymbol{g}_{\boldsymbol{\theta}_{t}}(\mathbf{z}_{\mathsf{val}})}_{\text{Gradient Influence Function} (\textbf{Term I})}
- \eta_{t}^{2}\underbrace{\boldsymbol{g}_{\boldsymbol{\theta}_t}(\mathbf{z}_i)^\top \boldsymbol{\mathcal{H}}_{\mathbf{z}_{\mathsf{val}}}(\boldsymbol{\theta}_t)
( \frac{1}{|\mathcal{B}_{t}^{(<i)}|} \sum_{\mathbf{z} \in \mathcal{B}_{t}^{(<i)}} \boldsymbol{g}_{\boldsymbol{\theta}_{t}}(\mathbf{z}))}_{\text{Hessian Weighted Relative Similarity}(\textbf{Term II})}
\label{eq:SecondApprox}
\end{align}

    where $\boldsymbol{\mathcal{H}}_{\mathbf{z}_{\mathsf{val}}}(\boldsymbol{\theta}
    _{t}) = \nabla_{\boldsymbol{\theta}}^{2}\ell(\mathbf{z}_{\mathsf{val}}, \boldsymbol
    {\theta}_{t})$ denotes the Hessian of the loss function with respect to the model
    parameters $\boldsymbol{\theta}$ evaluated at $\boldsymbol{\theta}_{t}$ for the
    validation data point $\mathbf{z}_{\mathsf{val}}$.

    \textbf{Gradient Influence Function}: The first term indicates the importance
    score of $\mathbf{z}_{i}$ {\em w.r.t} validation data point $\mathbf{z}_{\mathsf{val}}$ which, in essence, captures the effectiveness of the gradient
    of the training instance $\mathbf{z}_{i}$ towards the reduction in the validation
    loss. This term closely resembles the influence function proposed in \citep{pruthi2020estimating}.

    \textbf{Relative Similarity Term} The second term indicates the Hessian
    weighted relative similarity of the current training instance with all other
    training instances in the batch $\mathcal{B}_{t}^{(<i)}$.

    \textbf{Hessian Approximation Strategies} The Hessian term \(\boldsymbol{\mathcal{H}}_{\mathbf{z}_{\mathsf{val}}}(\boldsymbol{\theta}_{t})\) in Equation~\eqref{eq:SecondApprox} presents a major computational bottleneck due to its high cost. To alleviate this, several approximation strategies are commonly employed: \emph{Kronecker-Factored Approximation} methods~\citep{wu2020dissecting} exploit layer-wise structure and approximate the Hessian using Kronecker products; \emph{Gauss-Newton Decomposition}~\citep{sagun2017empirical} replaces the Hessian with the covariance of output gradients, assuming a negligible residual; and the \emph{Identity Approximation}~\citep{martens2015optimizing, nichol2018first} simplifies the Hessian to \(\mathbf{I}_d\), yielding a low-cost diagonal preconditioner. In our current list of experiments, we consider Hessian to \(\mathbf{I}_d\) as it is has been shown to be usefull with low approximation error in large scale trainings e.g. LLM settings \citep{wang2024greats}. In Appendix Section \ref{appendix:fisher information matrix}, we include other Hessian Approximation strategies which we tried out along with corresponding ablation studies.

    \subsection{Sample-wise Expected Marginal Gain}\label{section:samplewise_gain}
    We define the sample-wise expected marginal gain as the expectation of the conditional marginal utility gain over a validation instance $\mathbf{z}^{\mathsf{val}}_{t}$
    and a training instance $\mathbf{z}_{i}$ from the partially constructed mini-batch
$\mathcal{B}_{t}^{(<i)}$ as \(\mathds{E}_{\mathbf{z}_t^{\mathsf{val}} \in \mathcal{B}_t^{\mathsf{val}},\, \mathbf{z}_i \in \mathcal{B}_t^{(<i)}}
\left[
\bDelta \mathcal{U}_t\left(\mathbf{z}_i \mid \mathcal{B}_t^{(<i)},\, \mathbf{z}_t^{\mathsf{val}}\right)
\right]\)
        Here, due to the property of permutation invariance over the samples in $\mathcal{B}_{t}^{(<i)}$ as shown in Lemma \ref{Lemma:permutation:invariance}, the inner expectation can be written as:
    \begin{equation}
        \mathds{E}_{\mathbf{z}_i \in \mathcal{B}_{t}^{(<i)}}\left[ \bDelta \mathcal{U}
        _{t}(\mathbf{z}_{i}\mid \mathcal{B}_{t}^{(<i)}, \mathbf{z}^{\mathsf{val}}
        _{t}) \right] \triangleq \eta_{t}\bar{\mathbf{g}}_{\btheta_t}^{(b)}\cdot
        \boldsymbol{g}_{\btheta_t}(\mathbf{z}^{\mathsf{val}}_{t}) - \eta_{t}^{2}\bar
        {\mathbf{g}}_{\btheta_t}^{(b)\top}\left( \mathbf{I}_{d}- \frac{1}{|\mathcal{B}_{t}|}
        \mathbf{1}_{d \times |\mathcal{B}_t|}\mathbf{G}_{\btheta_t}^{\top}\right)
        \boldsymbol{\mathcal{H}}_{\mathbf{z}^{\mathsf{val}}_{t}}(\boldsymbol{\theta}
        _{t}) \bar{\mathbf{g}}_{\btheta_t}^{(b)}. \label{eq:vectorexpansion}
    \end{equation}
   
A direct greedy
    approach to maximize the conditional marginal gain at each step $t$ by iteratively
    selecting the training instance $\mathbf{z}_{i}^{\ast}\notin \mathcal{B}_{t}^{(<i)}$
    that yields the maximal local reduction in validation loss, i.e.,
    $\mathbf{z}_{i}^{\ast}= \operatorname{argmax}_{\substack{\mathbf{z}_i \notin \mathcal{B}_{t}^{(<i)}}}
    \bDelta \mathcal{U}_{t}(\mathbf{z}_{i}\mid \mathcal{B}_{t}^{(<i)}, \mathbf{z}^{\mathsf{val}}
    _{t})$, is computationally prohibitive. Constructing the new subset batch
    $\S_t$ of size $\beta$ from the current mini-batch
    $\mathcal{B}_{t}$ via this exhaustive greedy maximization starting from an
    empty set ($\mathcal{B}_{t}^{(<0)}= \emptyset$) incurs a computational
    complexity of
    $\mathcal{O}\binom{|\mathcal{B}_t|}{\beta}$.

    \textbf{Submodular Relaxation for Efficient Selection:} To overcome the computational intractability of exact optimization, we introduce a relaxation that exploits the structure of submodular functions to enable efficient selection of high-value subsets. Specifically, for each submodular function arm \( a_t \in \mathscr{A} \), we compute an approximately optimal subset \( \mathcal{S}_{a_t}^{\texttt{opt}} \subseteq \mathcal{B}_t \) of size at most \( \beta \), chosen to maximize the submodular objective \( \bof^{(a_t)}(\mathcal{S}) \). Since exact maximization of submodular functions is NP-hard, we adopt a standard greedy algorithm that offers a provable \( (1 - 1/e) \)-approximation guarantee under cardinality constraints.

\textbf{Reward Formulation using Submodular Function Arms:} We define the overall expected marginal gain \( \boldsymbol\vartheta: \mathscr{A} \times T \mapsto \mathbb{R} \) for each submodular function arm \( a_t \in \mathscr{A}\) at time step \( t \) as the expectation of the instance-wise conditional marginal gain \( \bDelta \mathcal{U}_t \), conditioned on a validation instance and a training instance from the approximately optimal subset \( \mathcal{S}^{\texttt{opt}}_{a_t} \):
\vspace{-0.8em}

\begin{equation}
\boldsymbol\vartheta(a_t \mid \B_t) = \mathds{E}_{\substack{\mathbf{z}^{\mathsf{val}}_t \in \mathcal{B}_{t}^{\mathsf{val}} , \mathbf{z}_i \in \mathcal{S}^{\texttt{opt}}_{a_t}}}
\left[ \bDelta \mathcal{U}_{t}(\mathbf{z}_i \mid {\mathcal{S}^{\texttt{opt}}_{a_t}}^{(<i)}, \mathbf{z}^{\mathsf{val}}_{t}) \right]
\label{eq:overallmarginalgain}
\end{equation}
The best arm is then selected via \(\hat{a}_t = \arg\maxop_{a_t \in \mathscr{A}}(\boldsymbol{\vartheta}(a_t \mid \B_t))\). 

\subsection{Speedup for $\methodprop$}

\textbf{Gradient Computation} Full-model gradients in deep networks are expensive to compute due to high-dimensionality. For vision tasks, we adopt last-layer gradients following \citep{ash2020deep}, and for LLMs, we compute gradients over LoRA adapters (rank 128) as in \citep{wang2024greats}. Both reduce overhead while preserving informative signals for subset selection.

\textbf{$\methodprop$-Batch}\label{Sec:PerBatch-variant} To align with batch-level baselines \citep{killamsetty2021grad}, we extend our samplewise formulation to the batch setting, treating each batch as a unit. Let $\mathbf{M}_{t}^{(b)} = \begin{bmatrix}
\thicktilde{\boldsymbol{g}}^{(b)}_{1} & \cdots & \thicktilde{\boldsymbol{g}}^{(b)}_{|\mathbb{S}_t|}
\end{bmatrix}$ be the matrix of average gradients $\thicktilde
    {\boldsymbol{g}_i}$ for batches $\B_i \in \mathbb{S}_t$, where  $\mathbb{S}_{t}$ denotes the set of sampled batches at time $t$. The expected conditional marginal gain becomes:
    \vspace{-0.8em}
\begin{multline}
\mathds{E}_{\B_i \in \mathbb{S}_{t_{[: \prec i]}}} \bDelta \mathcal{U}_{t}(\bullet) \triangleq \left[ 
\eta_t \mathbf{M}_t^{(b)} \mathbf{1}_{|\mathbb{S}_t|} \, \boldsymbol{g}(\mathbf{z}^{\texttt{val}}_t) 
- \eta_t^2 \mathbf{M}_t^{(b)} \left( \mathbf{1}_{|\mathbb{S}_t|} \mathbf{1}_{|\mathbb{S}_t|}^T 
- \mathbf{I}_{|\mathbb{S}_t|} \right) \boldsymbol{\mathcal{H}}_{\mathbf{z}^{\texttt{val}}_t} 
(\mathbf{M}_t^{(b)})^T \mathbf{1}_{|\mathbb{S}_t|} \right]
\end{multline}
    \vspace{-0.8em}
Other methods can be analogously adapted by substituting samples $\bx_i$ with batches $\B_i$.

\section{Algorithm}

$\methodprop$ instantiates a contextual multi-armed bandit framework to adaptively select curriculum policies throughout training.
%This parameterized schedule governs the stochasticity of arm selection via a single draw $\zeta \sim \mathcal{U}(0,1)$ compared against $\bXi_t$. If $\zeta > \bXi_t$, the model selects $\hat{a}_t = \arg\max_{a} \boldsymbol{\vartheta}(a \mid \mathcal{B}_t)$, favoring exploitation; otherwise, it samples $\hat{a}_t \sim \text{Uniform}(\mathscr{A})$ to encourage exploration. This structured annealing ensures early-stage diversity and late-stage stability, dynamically adjusting the curriculum in accordance with the model’s training trajectory.
%\vspace{-10pt}

\begin{tcolorbox}[
  colback=gray!20, colframe=gray!20,
  left=2mm, right=2mm, top=1mm, bottom=1mm,
  boxsep=0pt, width=1\linewidth,  before skip=2pt, after skip=2pt, 
    enlarge left by=0mm, enlarge right by=0mm,
]
  % ---- two minipages side by side ----
  \begin{minipage}[t]{0.48\linewidth}
\begin{algorithm}[H]
\small
\caption{$\methodprop$}
\label{alg:greedy}

\begin{algorithmic}[1] % [1] = line numbers
  \STATE \textbf{Input:} $T \in \mathbb{N}$ (total training steps);
         $\{\boldsymbol{f}^{(a)}\}_{a=1}^{K}$ (candidate submodular arms);
         $\lambda(\cdot), \pi(\cdot)$ (time-varying exploration parameters)
  \STATE \textbf{Output:} Final model parameter $\boldsymbol{\theta}_{T+1}$

  \STATE Initialize $\boldsymbol{\theta}_0$
  \FOR{$t = 1$ to $T$}
    \STATE \textbf{Receive} batch $\mathcal{B}_t$
    \STATE \textbf{Sample} $\zeta \sim \mathcal{U}(0,1)$
    \STATE \textbf{Threshold:} $\boldsymbol{\Xi}_t \leftarrow \frac{t}{(t + \lambda(t))^{\pi(t)}}$
    \STATE 
    $\hat{a}_t \gets
  \begin{cases}
\xxgreen{\arg\max\limits_{a_t \in \mathscr{A}} \boldsymbol{\vartheta} (a_t \mid \mathcal{B}_t)} & \text{if } \zeta > \bXi_t \\
\xblue{\text{Uniform}(\mathscr{A})} & \text{otherwise}
\end{cases}$\;
    \STATE $\mathcal{S}_{(\hat{a}_t)} \leftarrow \arg\max\limits_{|\mathcal{S}| \le \beta,\, \mathcal{S} \subseteq \mathcal{B}_t} \boldsymbol{f}^{(\hat{a}_t)}(\mathcal{S})$
    \STATE $\boldsymbol{\theta}_{t+1} \leftarrow \boldsymbol{\theta}_t - \frac{\eta_t}{|\mathcal{S}_{(\hat{a}_t)}|} 
    \sum\limits_{\mathbf{z} \in \mathcal{S}_{(\hat{a}_t)}} 
    \boldsymbol{g}_{\boldsymbol{\theta}_t}(\mathbf{z})$
  \ENDFOR
  \STATE \textbf{return} $\boldsymbol{\theta}_{T+1}$
\end{algorithmic}
\end{algorithm}
  \end{minipage}%
  \hspace{-1em}%
  \begin{minipage}[t]{0.55\linewidth}
    \vspace{0.3em} % optional vertical alignment
    \small
    \begin{itemize}
        \item \xsienna{\emph{\textbf{Step: 1-2}}} The model receives a batch $\mathcal{B}_t$ and chooses an arm $\hat{a}_t \in \mathscr{A}$, each corresponding to a distinct submodular utility function $\bof^{(\hat{a}_t)} : 2^{\mathcal{B}_t} \to \mathbb{R}_{\geq 0}$.
        \item \xsienna{\emph{\textbf{Step: 5}}} The arm selection is governed by a exploration threshold $\xxpurple{\bXi_t} := \frac{t}{(t + \lambda(t))^{\pi(t)}}$, parameterized by time-dependent schedules $\lambda(t)$ and $\pi(t)$ that modulate the annealing from exploration to exploitation. Here, $\lambda(t)$ \textbf{(Exploration Dampening)} and $\pi(t)$ \textbf{(Exploration Sharpness)} act as curriculum schedulers. If a uniform sample satisfies $\zeta > \bXi_t$, the algorithm enters \textbf{\xxgreen{\emph{Exploitation Phase}}} and selects the arm maximizing $\boldsymbol{\vartheta}(a \mid \mathcal{B}_t)$; otherwise, an arm is sampled uniformly at random (\textbf{\xblue{\emph{Exploration Phase}}}).
        \item \xsienna{\emph{\textbf{Step: 6}}} Once an arm $\hat{a}_t$ is selected, the algorithm performs approximate maximization over $\mathcal{B}_t$ with respect to $\bof^{(\hat{a}_t)}$, selecting a subset $\mathcal{S}_{(\hat{a}_t)}$.
        \item \xsienna{\emph{\textbf{Step: 7}}} The model parameters $\boldsymbol{\theta}_t$ are then updated using a stochastic gradient step computed only on the selected subset.  
    \end{itemize}
\end{minipage}
\end{tcolorbox}

\section{Theoretical Results}

In this section, we present the main theoretical results of our work, focusing on regret guarantees for our best-arm selection policy. Specifically, we analyze the regret incurred by our method relative to the performance of the optimal arm in hindsight. This requires a set of structural assumptions (pertaining to describe properties of the exploration dynamics, utility approximation quality, and the existence of a reward gap between optimal and suboptimal arms).

\textbf{Assumption (a)}  (\textbf{Constant Fractional Exploration Dampening}):  
    The exploration dampening parameter \(\lambda(t)\) is time-invariant \(
    \lambda(t) = \epsilon\) where \(\quad \epsilon \in (0,1).\)

\textbf{Assumption (b)} (\textbf{Optimality Gap}):  
    There exists an optimality gap \(\boldsymbol{\varrho}\) such that for every suboptimal arm \(a_t \in \mathscr{A} \setminus \{a^*\}\) : \(
    0 \leq \boldsymbol{\varrho} \leq \bDelta_{(a_t)}(\mathcal{B}_t).
    \)

 \textbf{Assumption (c)}  (\textbf{Fractional Exploration Sharpness}):  The exploration sharpness parameter \(\pi(t)\) is a bounded quantity \(\pi(t) \in (0,1).\)

\textbf{Assumption (d)}  (\textbf{Utility Metric Approximation}):  
    The utility metric \(\mathcal{U}_t(\cdot, \cdot)\) satisfies the approximation bound as per Theorem 2 (Appendix) with constants \(\mathfrak{C}_{(a)}\) for each arm \(a \in \mathscr{A}\) and let \(n_a\) be a specific constant associated with arm $a$ such that Theorem 2 (Appendix) holds true.

\begin{restatable}[\textbf{Regret Guarantees}]{theorem}{RegretBound}\label{theorem:regretbound}
Under Assumptions \textbf{a - d}, for all \(t > t_0\), with probability at least
\[
1 - \mathcal{K} \exp\left(-\frac{3(t - 2)(1 + (1 - \pi)\epsilon)}{28 \mathcal{K} (2 - \pi)}\right),
\]
the expected instantaneous regret incurred by the arm selection policy satisfies
\begin{equation}
\begin{split}
\mathds{E}[\mathrm{Regret}_t] &:= \mathds{E}_{\mathcal{B}_t} \mathds{E}_{\hat{a}_t \in \mathscr{A}} \mathds{E}_{\boldsymbol{\vartheta}} \left[ \boldsymbol{\vartheta}(a_t^* \mid \mathcal{B}_t) - \boldsymbol{\vartheta}(\hat{a}_t \mid \mathcal{B}_t) \right] \\
&= O\left(\frac{1}{t}\right) + O\left( \frac{\mathcal{K}^{3/2} (\max_{a}\mathfrak{C}_{(a)} + \mathfrak{C}_*)}{\boldsymbol{\varrho}} \sqrt{\frac{\log t}{t}} \right),
\end{split}
\label{eq:regretbound}
\end{equation}
where \(\mathfrak{C}_*\) is the approximation constant corresponding to the optimal arm \(a^*\).
\end{restatable}

The theorem guarantees that, under the specified assumptions, the arm selection strategy based on maximizing the expected marginal utility gain converges to the optimal arm almost surely, with the regret decreasing at a rate combining a fast \(1/t\) decay and a slower \(\sqrt{\frac{\log t}{t}}\) decay modulated by constants related to the utility approximation and the number of arms. The presence of the optimality gap \(\boldsymbol{\varrho}\) in the denominator highlights the difficulty of distinguishing between arms when their utility values are close.  We also showcase proofs in Appendix when  Assumption (\textbf{a}) and Assumption (\textbf{c}) are relaxed with no constraints on the bounds of $\lambda(\cdot)$ and $\pi(\cdot)$.
\subsection{Supporting Lemmas}
Here we detail out Supporting Lemmas that are utilised in proofs and derivations above.

\begin{restatable}[\textbf{Permutation Invariance of Expected Marginal Gain}]{lemma}{PermInvariance}\label{Lemma:permutation:invariance}
Let $\Pi$ denote the set of all permutations over the elements of $\mathcal{B}_t^{(<i)}$. Then the expected marginal gain
\(
\mathds{E}_{\mathbf{z}_i \in \mathcal{B}_t^{(<i)}}\left[ \bDelta \mathcal{U}_t(\mathbf{z}_i \mid \mathcal{B}_t^{(<i)}, \mathbf{z}^{\mathsf{val}}_t) \right]
\)
is invariant under any permutation $\pi \in \Pi$, i.e.,
\[
\mathds{E}_{\mathbf{z}_i \in \mathcal{B}_t^{(<i)}}\left[ \bDelta \mathcal{U}_t(\mathbf{z}_i \mid \mathcal{B}_t^{(<i)}, \mathbf{z}^{\mathsf{val}}_t) \right]
=
\displaystyle \mathds{E}_{\mathbf{z}_i \in \pi(\mathcal{B}_t^{(<i)})}\left[ \bDelta \mathcal{U}_t(\mathbf{z}_i \mid \pi(\mathcal{B}_t^{(<i)}), \mathbf{z}^{\mathsf{val}}_t) \right].
\]
\end{restatable}

We provide the detailed derivations for all proofs in Appendix \ref{Appendix:TheoreticalSection}.

 \section{Experimental Setup}
 We evaluate $\methodprop$ across diverse datasets to highlight its advantages in terms of both accuracy and computational efficiency. 
 %  In order to showcase that $\methodprop$ performs better.   Our experiments aim to showcase the efficiency and accuracy tradeoffs of our    proposed approach $\methodprop$ across a range of configurations and on a variety of datasets. 
 %Experiments are conducted for vision and language tasks using NVIDIA A6000 and H100 GPUs respectively, ensuring consistent comparison with baselines.
 All vision-related experiments are conducted using NVIDIA 3 $\times$ A6000 GPUs, while large language model (LLM) experiments are performed on 8 $\times$ H100 GPUs to ensure fair comparisons with all baselines. We share more details in Appendix Section \ref{Appendix: Experiment Details}.

\subsection{Finetuning Large Language Models}

\textbf{Model-Training-Evaluation Pairs.} 
We evaluate $\methodprop$ using combinations of two LLMs: \texttt{LLAMA-2-7B} \citep{touvron2023llama2} and \texttt{MISTRAL-7B} \citep{Jiang2023Mistral7B} finetuned on LESS \citep{xia2024less}, with performance assessed on MMLU and TYDIQA (Table \ref{ref:table-main:llm}).
We use batch size of 16 and use 2 random validation points for computing the reward utility. We select 50\% of the batch data for gradient updates during each step. 

\begin{table}[h!]
\centering
\scriptsize
\caption{Performance comparison across tasks. Bold indicates best performance in each column.}
\rowcolors{2}{gray!20}{white}
\resizebox{0.85\textwidth}{!}{
\begin{tabular}{lccccccccccc}
\toprule
\rowcolor{gray!20}
\textbf{Method} & \textbf{Avg.} & \textbf{Soc.} & \textbf{Pol.} & \textbf{Hist.} & \textbf{Anat.} & \textbf{ML.} & \textbf{Eth.} & \textbf{Gen.} & \textbf{Bio.} & \textbf{Chem.} & \textbf{TydiQA} \\
\midrule
GradNorm      & 46.4\% & 61.0\% & 62.5\% & \textbf{52.1\%} & 40.5\% & 40.2\% & 43.0\% & 46.7\% & 42.9\% & 32.3\% & 54.6\% \\
MaxLoss       & 45.2\% & 60.2\% & 64.4\% & 48.0\%          & 39.5\% & 38.1\% & 44.4\% & 43.8\% & 42.6\% & 31.1\% & 55.4\% \\
RhoLoss       & 46.4\% & 60.6\% & 66.2\% & 49.4\%          & 41.5\% & 40.2\% & 42.8\% & 46.1\% & 41.1\% & 33.7\% & 55.2\% \\
SBERT         & 45.8\% & 62.3\% & 63.7\% & 47.0\%          & 43.1\% & 36.8\% & 43.4\% & 44.2\% & 42.0\% & 32.4\% & 54.2\% \\
GREATS        & 47.8\% & 63.2\% & 66.2\% & 48.3\%          & 42.6\% & 41.1\% & 46.2\% & 48.9\% & 43.1\% & 33.6\% & 55.7\% \\
\textbf{$\methodprop$} & \textbf{49.6\%} & \textbf{65.3\%} & \textbf{67.4\%} & \textbf{52.1\%} & \textbf{45.2\%} & \textbf{42.7\%} & \textbf{48.6\%} & \textbf{50.9\%} & \textbf{45.1\%} & \textbf{35.7\%} & \textbf{55.9\%} \\
\bottomrule
\end{tabular}
}
\label{ref:table-main:llm}
\end{table}

\begin{figure}[htbp]
 \centering

 \begin{minipage}[t]{\textwidth}
 \centering \scalebox{0.3}{\begin{tikzpicture}
    \begin{axis}[
        width=0.8\textwidth,
        height=0.6\textwidth,
        xlabel={Steps},
        ylabel={Test Perplexity},
         title={Test Perplexity (Policy)},
        grid=both,
        grid style={dashed, gray!30},
        legend pos=north east,
        font=\large, % <--- Increase font size here
        label style={font=\large}, % Optional: fine-grained control
        tick label style={font=\large}, % Optional: tick font
        title style={font=\large}, % Optional: title font
        legend style={font=\large} % Optional: legend font
    ]
        \addplot table [x index=0, y index=1] {AuthorKit25/AnonymousSubmission/LaTeX/LLMs_results/results_dat/2shot_poli_GREATS.dat};
        \addlegendentry{GREATS}
        \addplot table [x index=0, y index=1] {AuthorKit25/AnonymousSubmission/LaTeX/LLMs_results/results_dat/2shot_poli_MaxLoss.dat};
        \addlegendentry{MaxLoss}
        \addplot table [x index=0, y index=1] {AuthorKit25/AnonymousSubmission/LaTeX/LLMs_results/results_dat/2shot_poli_GradNorm.dat};
        \addlegendentry{GradNorm}
        \addplot table [x index=0, y index=1] {AuthorKit25/AnonymousSubmission/LaTeX/LLMs_results/results_dat/2shot_poli_onlineSubmod.dat};
        \addlegendentry{\textbf{$\methodprop$}}
    \end{axis}
\end{tikzpicture}}
 \scalebox{0.3}{%
 \begin{tikzpicture}
    \begin{axis}[
        width=0.8\textwidth,
        height=0.6\textwidth,
        xlabel={Steps},
        ylabel={Test Perplexity},
         title={Test Perplexity (Anatomy)},
        grid=both,
        grid style={dashed, gray!30},
        legend pos=north east,
        font=\large, % <--- Increase font size here
        label style={font=\large}, % Optional: fine-grained control
        tick label style={font=\large}, % Optional: tick font
        title style={font=\Large}, % Optional: title font
        legend style={font=\large} % Optional: legend font
    ]
        \addplot table [x index=0, y index=1] {AuthorKit25/AnonymousSubmission/LaTeX/LLMs_results/results_dat/2shot_anatomy_GREATS.dat};
        \addlegendentry{GREATS}
        \addplot table [x index=0, y index=1] {AuthorKit25/AnonymousSubmission/LaTeX/LLMs_results/results_dat/2shot_anatomy_MaxLoss.dat};
        \addlegendentry{MaxLoss}
        \addplot table [x index=0, y index=1] {AuthorKit25/AnonymousSubmission/LaTeX/LLMs_results/results_dat/2shot_anatomy_GradNorm.dat};
        \addlegendentry{GradNorm}
        \addplot table [x index=0, y index=1] {AuthorKit25/AnonymousSubmission/LaTeX/LLMs_results/results_dat/2shot_anatomy_onlineSubmod.dat};
        \addlegendentry{\textbf{$\methodprop$}}
    \end{axis}
\end{tikzpicture}}
 \scalebox{0.3}{%
 \begin{tikzpicture}
    \begin{axis}[
        width=0.8\textwidth,
        height=0.6\textwidth,
        xlabel={Steps},
        ylabel={Test Perplexity},
         title={Test Perplexity (Sociology)},
        grid=both,
        grid style={dashed, gray!30},
        legend pos=north east,
        font=\large, % <--- Increase font size here
        label style={font=\large}, % Optional: fine-grained control
        tick label style={font=\large}, % Optional: tick font
        title style={font=\large}, % Optional: title font
        legend style={font=\large} % Optional: legend font
    ]
        \addplot table [x index=0, y index=1] {AuthorKit25/AnonymousSubmission/LaTeX/LLMs_results/results_dat/2shot_redo2_soc_GREATS.dat};
        \addlegendentry{GREATS}
        \addplot table [x index=0, y index=1] {AuthorKit25/AnonymousSubmission/LaTeX/LLMs_results/results_dat/2shot_redo2_soc_MaxLoss.dat};
        \addlegendentry{MaxLoss}
        \addplot table [x index=0, y index=1] {AuthorKit25/AnonymousSubmission/LaTeX/LLMs_results/results_dat/2shot_redo2_soc_GradNorm.dat};
        \addlegendentry{GradNorm}
        \addplot table [x index=0, y index=1] {AuthorKit25/AnonymousSubmission/LaTeX/LLMs_results/results_dat/2shot_redo2_soc_onlineSubmod.dat};
        \addlegendentry{\textbf{$\methodprop$}}
    \end{axis}
\end{tikzpicture}}
 \scalebox{0.3}{%
 \begin{tikzpicture}
    \begin{axis}[
        width=0.8\textwidth,
        height=0.6\textwidth,
        xlabel={Steps},
        ylabel={Test Perplexity},
        title={Test Perplexity (Chemistry)},
        grid=both,
        grid style={dashed, gray!30},
        legend pos=north east,
        font=\large, % <--- Increase font size here
        label style={font=\large}, % Optional: fine-grained control
        tick label style={font=\large}, % Optional: tick font
        title style={font=\large}, % Optional: title font
        legend style={font=\large} % Optional: legend font
    ]
        \addplot table [x index=0, y index=1] {AuthorKit25/AnonymousSubmission/LaTeX/LLMs_results/results_dat/2shot_redo2_hchem_GREATS.dat};
        \addlegendentry{GREATS}
        \addplot table [x index=0, y index=1] {AuthorKit25/AnonymousSubmission/LaTeX/LLMs_results/results_dat/2shot_redo2_hchem_MaxLoss.dat};
        \addlegendentry{MaxLoss}
        \addplot table [x index=0, y index=1] {AuthorKit25/AnonymousSubmission/LaTeX/LLMs_results/results_dat/2shot_redo2_hchem_GradNorm.dat};
        \addlegendentry{GradNorm}
        \addplot table [x index=0, y index=1] {AuthorKit25/AnonymousSubmission/LaTeX/LLMs_results/results_dat/2shot_redo2_hchem_onlineSubmod.dat};
        \addlegendentry{\textbf{$\methodprop$}}
    \end{axis}
\end{tikzpicture}}
 \end{minipage}

 \caption{\textbf{Test perplexity dynamics} on \texttt{LLAMA-2-7B} during training with various \textbf{online batch selection strategies} on \texttt{MMLU}. We evaluate on \texttt{US Foreign Policy}, \texttt{Anatomy}, \texttt{Sociology}, and \texttt{Chemistry}. $\methodprop$ significantly outperforms baselines.}
 \label{fig:llm_analysis}
\end{figure}

\textbf{Baselines}: We compare our algorithm with a variety of online batch selection
algorithms: \blackcircle{1} $\baselinemaxloss$ \citep{maxloss}, which selects training data points with the highest loss values. \blackcircle{2}
$\baselinegradnorm$ \citep{katharopoulos2018not}, which prioritizes training data points with the highest gradient norms,\blackcircle{3} $\baselinerho$\cite{mindermann2022prioritized}, using LLaMA-3.1-8B-Instruct as the reference and LLaMA-2-7B as the target.
\blackcircle{4} $\baselinesbert$, which selects batches by semantic similarity to validation data using Sentence-BERT embeddings~\citep{reimers2019sentence}.
 \blackcircle{5} $\baselinegreats$ \citep{wang2024greats} which has a similar utility metric as ours, but the optimization objective instead relies on directly selecting samples greedily that maximizes the utility reward Eqn~\eqref{eq:SecondApprox} rather than utilizing any monotone submodular characteristics. 

\textbf{Observations:} As can be seen from the perplexity curves (Figure \ref{fig:llm_analysis}) and downstream performance (Table \ref{ref:table-main:llm}), $\methodprop$ significantly outperforms other existing baselines, thereby indicating how principled validation performance aware reward signal combined with induced submodular curriculum results in better generalization than static heuristic based approaches.

\vspace{-5pt}

\subsection{Image Classification}
    We showcase the utility of our method across 5 datasets primarily \textsc{CIFAR10}, \textsc{CIFAR100}
\citep{krizhevsky2009learning},\textsc{TinyImagenet}\citep{le2015tiny}, \textsc{MNIST} \citep{lecun2010mnist}  and \textsc{SVHN} \citep{svhn}. We compare $\methodprop$
    with:
   \blackcircle{1} $\baselineGradmatch$: \citep{killamsetty2021grad}, \blackcircle{2} $\baselinecraig$ \citep{mirzasoleiman2020coresets},
     \blackcircle{3} $\baselineglister$ \citep{killamsetty2021glister}, \blackcircle{4} $\baselinerho$ \citep{mindermann2022prioritized}
    and \blackcircle{5} $\baselineboss$ \citep{acharyabalancing}. All models are trained on a ResNet backbone, 300 epochs, with 20 epochs warm-start (we provide more details regarding cold start vs warm start in Appendix Sec: \ref{Warm Start}). We provide a detailed summary of individual baselines, comparision with more datasets and hyperparameters in Appendix \ref{Appendix: Experiment Details}. 

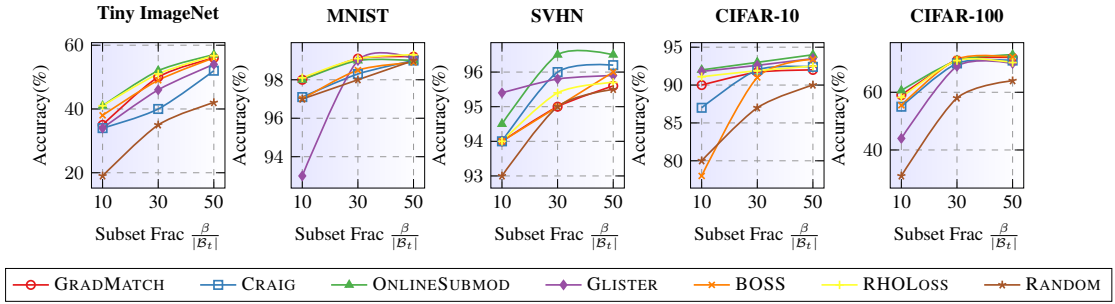
\begin{figure*}[h!]
        \centering
        \scalebox{0.8}{\pgfplotsset{%
    width=.8\textwidth,
    height=1.5\textwidth
}
% Data for Tiny ImageNet

% Data for MNIST

% Data for SVHN

% Data for CIFAR-10

% Data for CIFAR-100

\pgfplotsset{
    width=0.45\textwidth, % Reduce width of each plot for side-by-side
    height=0.45\textwidth, % Adjust height to maintain aspect ratio
}

\begin{tikzpicture}
    \begin{groupplot}[
        group style={
            group size=5 by 1,
            horizontal sep=1.1cm,
        },
                 axis background/.style={ % Define the gradient background
                shade, 
                left color=blue!10, 
                right color=white
            },
        width=3.8cm,
        height=4.0cm,
        symbolic x coords={10,30,50},
        xtick={10,30,50},
        xlabel={\small Subset Frac $\frac{\beta}{|\mathcal{B}_t|}$},
        ylabel={\small Accuracy(\%)},
        grid=major,
        grid style={dashed, gray!80},
        tick style={black},
        tick label style={font=\small},
        label style={font=\small},
        title style={font=\small\bfseries},
        legend style={
            font=\footnotesize,
            at={(3.5, -0.55)},
            anchor=north,
            legend columns=7,
            /tikz/every even column/.append style={column sep=0.5cm}
        },
        cycle list/Set1-7
    ]

    % Example plot group for Tiny ImageNet
    \nextgroupplot[title={Tiny ImageNet}]
    \addplot+[thick, mark=o, smooth] table[x index=0, y index=1, col sep=space] {tiny_imagenet_gradmatch.dat};
    \addlegendentry{\textsc{GradMatch}}
    \addplot+[thick, mark=square, smooth] table[x index=0, y index=1, col sep=space] {tiny_imagenet_craig.dat};
    \addlegendentry{\textsc{Craig}}
    \addplot+[thick, mark=triangle*, smooth] table[x index=0, y index=1, col sep=space] {tiny_imagenet_onlinesubmod.dat};
    \addlegendentry{\textsc{OnlineSubmod}}
    \addplot+[thick, mark=diamond*, smooth] table[x index=0, y index=1, col sep=space] {tiny_imagenet_glister.dat};
    \addlegendentry{\textsc{Glister}}
    \addplot+[thick, mark=x, smooth] table[x index=0, y index=1, col sep=space] {tiny_imagenet_boss.dat};
    \addlegendentry{\textsc{BOSS}}
    \addplot+[thick, mark=+, smooth] table[x index=0, y index=1, col sep=space] {tiny_imagenet_rho.dat};
    \addlegendentry{\textsc{RHOLoss}}
    \addplot+[thick, mark=star, smooth] table[x index=0, y index=1, col sep=space] {tiny_imagenet_random.dat};
    \addlegendentry{\textsc{Random}}

    % Repeat similarly for other datasets
    \nextgroupplot[title={MNIST}]
    \addplot+[thick, mark=o, smooth] table[x index=0, y index=1, col sep=space] {mnist_gradmatch.dat};
    \addplot+[thick, mark=square, smooth] table[x index=0, y index=1, col sep=space] {mnist_craig.dat};
    \addplot+[thick, mark=triangle*, smooth] table[x index=0, y index=1, col sep=space] {mnist_onlinesubmod.dat};
    \addplot+[thick, mark=diamond*, smooth] table[x index=0, y index=1, col sep=space] {mnist_glister.dat};
    \addplot+[thick, mark=x, smooth] table[x index=0, y index=1, col sep=space] {mnist_boss.dat};
    \addplot+[thick, mark=+, smooth] table[x index=0, y index=1, col sep=space] {mnist_rho.dat};
    \addplot+[thick, mark=star, smooth] table[x index=0, y index=1, col sep=space] {mnist_random.dat};

    \nextgroupplot[title={SVHN}]
    \addplot+[thick, mark=o, smooth] table[x index=0, y index=1, col sep=space] {svhn_gradmatch.dat};
    \addplot+[thick, mark=square, smooth] table[x index=0, y index=1, col sep=space] {svhn_craig.dat};
    \addplot+[thick, mark=triangle*, smooth] table[x index=0, y index=1, col sep=space] {svhn_onlinesubmod.dat};
    \addplot+[thick, mark=diamond*, smooth] table[x index=0, y index=1, col sep=space] {svhn_glister.dat};
    \addplot+[thick, mark=x, smooth] table[x index=0, y index=1, col sep=space] {svhn_boss.dat};
    \addplot+[thick, mark=+, smooth] table[x index=0, y index=1, col sep=space] {svhn_rho.dat};
    \addplot+[thick, mark=star, smooth] table[x index=0, y index=1, col sep=space] {svhn_random.dat};

    \nextgroupplot[title={CIFAR-10}]
    \addplot+[thick, mark=o, smooth] table[x index=0, y index=1, col sep=space] {cifar10_gradmatch.dat};
    \addplot+[thick, mark=square, smooth] table[x index=0, y index=1, col sep=space] {cifar10_craig.dat};
    \addplot+[thick, mark=triangle*, smooth] table[x index=0, y index=1, col sep=space] {cifar10_onlinesubmod.dat};
    \addplot+[thick, mark=diamond*, smooth] table[x index=0, y index=1, col sep=space] {cifar10_glister.dat};
    \addplot+[thick, mark=x, smooth] table[x index=0, y index=1, col sep=space] {cifar10_boss.dat};
    \addplot+[thick, mark=+, smooth] table[x index=0, y index=1, col sep=space] {cifar10_rho.dat};
    \addplot+[thick, mark=star, smooth] table[x index=0, y index=1, col sep=space] {cifar10_random.dat};

    \nextgroupplot[title={CIFAR-100}]
    \addplot+[thick, mark=o, smooth] table[x index=0, y index=1, col sep=space] {cifar100_gradmatch.dat};
    \addplot+[thick, mark=square, smooth] table[x index=0, y index=1, col sep=space] {cifar100_craig.dat};
    \addplot+[thick, mark=triangle*, smooth] table[x index=0, y index=1, col sep=space] {cifar100_onlinesubmod.dat};
    \addplot+[thick, mark=diamond*, smooth] table[x index=0, y index=1, col sep=space] {cifar100_glister.dat};
    \addplot+[thick, mark=x, smooth] table[x index=0, y index=1, col sep=space] {cifar100_boss.dat};
    \addplot+[thick, mark=+, smooth] table[x index=0, y index=1, col sep=space] {cifar100_rho.dat};
    \addplot+[thick, mark=star, smooth] table[x index=0, y index=1, col sep=space] {cifar100_random.dat};

    \end{groupplot}
\end{tikzpicture}}
        % \vspace{0.2em}
        \caption{\textbf{Samplewise Submodular Curriculum:} $\methodprop$ consistently achieves top-1 accuracy across all subset sizes on \textsc{TinyImageNet}, \textsc{SVHN}, \textsc{CIFAR-10}, and \textsc{CIFAR-100}, and remains competitive on \textsc{MNIST}. Notably, it matches or outperforms all baselines at early subset fractions (10\%, 30\%) on all datasets except MNIST.}
    \label{fig:samplewise_results}
\end{figure*}

\textbf{Performance Metrics across all baselines} We work with the batch-wise variant for $\methodprop$ (Section \ref{Sec:PerBatch-variant}) keeping in line with other methods. To compare various baselines,
    we utilize \textbf{Speedup} as a relative measure of the training times for each
    baseline in relation to full batch training. Our goal is to identify a baseline
    that achieves both high speedup and high test accuracy. We evaluate on budget fractions ($\frac{\beta}{\B_t} \times 100 \%$) of 10\%, 30\% and 50\%. From Table \ref{table:image batchwise:results}, we observe that our method significantly outperforms baselines in accuracy, with speedup extremely close to the optimal speedup, obtained by $\baselinemilo$ which relies on an expensive offline filtering step based on some assorted selection of submodular functions, contrast to our method which dynamically selects subsets in an online manner. However, this gap is minimal, while our method achieves
    higher accuracy. For completeness, we also showcase per samplewise selection in Figure \ref{fig:samplewise_results}.

\begin{table}[h!]
\centering
\vspace{0.5em}
\rowcolors{2}{gray!20}{white}
\resizebox{\textwidth}{!}{
\begin{tabular}{|l|c|c|c|c|c|c|c|c|c|c|}
\toprule
\rowcolor{gray!20}
\textbf{Method} & \multicolumn{3}{c|}{\textbf{TinyImageNet}} & \multicolumn{3}{c|}{\textbf{CIFAR-100}} & \multicolumn{3}{c|}{\textbf{CIFAR-10}} \\
\cline{2-10}
 & \textbf{10\%} & \textbf{30\%} & \textbf{50\%} & \textbf{10\%} & \textbf{30\%} & \textbf{50\%} & \textbf{10\%} & \textbf{30\%} & \textbf{50\%} \\
\midrule
$\baselinecraig$ \citep{mirzasoleiman2020coresets} & 0.524 / 4.82 & 0.555 / 2.41 & 0.615 / 1.7 & 0.672 / 5.1 & 0.723 / 2.5 & 0.751 / 1.5 & 0.900 / 6.7 & 0.924 / 1.9 & 0.931 / 1.15 \\
$\baselinemilo$ \citep{killamsetty2023milo} & \cellcolor{red!30}
 0.532 / 8.62 & 0.593 / 3.1 & \cellcolor{yellow!40} 0.623 / 2.6 & \cellcolor{yellow!40} 0.723 / 10.1 & \cellcolor{red!30}
0.746 / 3.5 & \cellcolor{red!30}
0.756 / 1.95 & \cellcolor{yellow!40} 0.922 / 5.8 & \cellcolor{yellow!40} 0.932 / 2.05 & \textbf{0.941} / \cellcolor{green!30} 2.15 \\
$\baselineGradmatch$ \citep{killamsetty2021grad} & 0.526 / 5.92 & 0.581 / 2.62 & 0.619 / 2.1 & 0.683 / 6.9 & 0.746 / 3.1 & 0.753 / 1.3 & \cellcolor{red!30}
0.922 / 4.3 & \cellcolor{red!30}
0.932 / 1.95 & \cellcolor{red!30}
0.941 / 1.48 \\
$\baselineglister$ \citep{killamsetty2021glister} & 0.515 / 5.5 & 0.563 / 2.65 & 0.621 / 1.7 & 0.642 / 7.7 & 0.723 / 2.6 & 0.746 / 1.2 & 0.911 / 4.5 & 0.921 / 1.7 & 0.926 / 1.3 \\
$\baselinerho$ \citep{mindermann2022prioritized} & \cellcolor{yellow!40}0.544 / 5 & \cellcolor{red!30}
 0.597 / 2.57 & 0.621 / 2 & 0.713 / 3.9 & \cellcolor{yellow!40} 0.748 / 1.9 & \cellcolor{yellow!40} 0.757 / 1.2 & 0.901 / 2.5 & 0.915 / 1.6 & 0.941 / 1.15 \\
$\baselineboss$ \citep{acharyabalancing} & 0.526 / 5.4 & \cellcolor{yellow!40}0.601 / 2.9 & \cellcolor{red!30}
 0.621 / 2.15 & \cellcolor{red!30}
 0.717 / 7.8 & 0.737 / 3 & 0.754 / 1.9 & 0.916 / 4.9 & 0.930 / 1.8 & 0.938 / 1.53 \\
$\methodprop$ & \cellcolor{green!30} \textbf{0.553} / 8.43 & \cellcolor{green!30} \textbf{0.607} / 3.08 & \cellcolor{green!30} \textbf{0.626} / 2.6 & \cellcolor{green!30} \textbf{0.736} / 9.2 & \cellcolor{green!30} \textbf{0.754} / 3.3 & \cellcolor{green!30} \textbf{0.758} / 1.92 & \cellcolor{green!30} \textbf{0.924} / 5.4 & \cellcolor{green!30} \textbf{0.937} / 2 & \cellcolor{yellow!40} 0.941 / 2.08 \\
\bottomrule
\end{tabular}
 }
\vspace{0.5em}
\caption{\textbf{Batchwise version performance:} Accuracy vs Speedup.
\colorbox{green!30}{\color{green!30}\rule{1.8em}{0.3em} } $\mapsto$ highest accuracy 
\colorbox{yellow!40}{\color{yellow!40}\rule{1.8em}{0.3em}} $\mapsto$ 2nd highest accuracy 
\colorbox{red!30}{\color{red!30}\rule{1.8em}{0.3em}} $\mapsto$ 3rd highest accuracy.: Performance comparison across different datasets and fractions. Bold indicates the best performance in each column.
}
\label{table:image batchwise:results}
\end{table}

\vspace{-2em}

    \subsection{Ablation Study results}
    To understand how the choice of submodular function at each step affects the
    model performance under \textbf{\xblue{\emph{Exploration}}/\xxgreen{\emph{Exploitation}}} scheme, it is
    important to understand how the underlying variables affect the overal
    submodular function selection and thereby model performance at each step.

    Here we study the effect of the $\lambda(t)$ and $\pi(t)$. Note for time independent constants we ignore the argument inside $\lambda(\cdot), \pi(\cdot)$.
    
\begin{figure}[t]
    \centering
\begin{tikzpicture}[font=\scriptsize]
    \begin{groupplot}[
        group style={
            group size=4 by 1,
            horizontal sep=0.55cm,
            vertical sep=2cm,
        },
        width=4.5cm,
        height=4cm,
        grid=both,
        tick label style={font=\scriptsize},
        cycle list name=color list,
        legend style={fill=none, draw=none, scale=0.1}
    ]

    \nextgroupplot[
        title={$\mathbf{\lambda(t)=0.5}$},
        title style={font=\scriptsize, align=center}
    ]
    \addplot+[mark=*,mark size=1pt,  line width=0.3mm] table [x index=0, y index=1, col sep=space] {data_one.dat};
    \addlegendentry{\textbf{Term I}}
    \addplot+[mark=square*,mark size=1pt,  line width=0.3mm] table [x index=0, y index=2, col sep=space] {data_one.dat};
    \addlegendentry{\textbf{Term II}}
    \addplot+[mark=triangle*,mark size=1pt,  line width=0.3mm] table [x index=0, y index=3, col sep=space] {data_one.dat};
    \addlegendentry{$\bDelta \mathcal{U}_{t}$}

    \nextgroupplot[
        title={$\mathbf{\lambda(t)=0.3}$},
        title style={font=\scriptsize, align=center}
    ]
    \addplot+[mark=*, mark size=1pt, line width=0.3mm] table [x index=0, y index=1, col sep=space] {data_two.dat};
     \addlegendentry{\textbf{Term I}}
    \addplot+[mark=square*,mark size=1pt,  line width=0.3mm] table [x index=0, y index=2, col sep=space] {data_two.dat};
        \addlegendentry{\textbf{Term II}}
    \addplot+[mark=triangle*,mark size=1pt,  line width=0.3mm] table [x index=0, y index=3, col sep=space] {data_two.dat};
        \addlegendentry{$\bDelta \mathcal{U}_{t}$}

    \nextgroupplot[
        title={$\mathbf{\lambda(t) = \exp^{-0.5t}}$},
        title style={font=\scriptsize, align=center}
    ]
    \addplot+[mark=*, mark size=1pt, line width=0.3mm] table [x index=0, y index=1, col sep=space] {data_three.dat};
    \addplot+[mark=square*, mark size=1pt, line width=0.3mm] table [x index=0, y index=2, col sep=space] {data_three.dat};
    \addplot+[mark=triangle*,mark size=1pt,  line width=0.3mm] table [x index=0, y index=3, col sep=space] {data_three.dat};

    \nextgroupplot[
        title={$\mathbf{\lambda(t)=0.7}$},
        title style={font=\scriptsize, align=center}
    ]
    \addplot+[mark=*, mark size=1pt, line width=0.3mm] table [x index=0, y index=1, col sep=space] {data_four.dat};
        \addlegendentry{\textbf{Term I}}
    \addplot+[mark=square*, mark size=1pt, line width=0.3mm] table [x index=0, y index=2, col sep=space] {data_four.dat};
            \addlegendentry{\textbf{Term II}}
    \addplot+[mark=triangle*,mark size=1pt, line width=0.3mm] table [x index=0, y index=3, col sep=space] {data_four.dat};
        \addlegendentry{$\bDelta \mathcal{U}_{t}$}

    \end{groupplot}

    % Place the legend in the middle of the 4 plots
% \path (group c1r2.south east) -- (group c2r2.south west) coordinate[right] (legendpos);
%     \node[below] at (legendpos) {\pgfplotslegendfromname{sharedlegend}};
\end{tikzpicture}
    \caption{ Evolution of \textbf{Term I} and \textbf{Term II} (Eq \ref{eq:SecondApprox}) across training epochs on \textsc{CIFAR-100}.}
    \label{fig:lambda-evolution}
\end{figure}
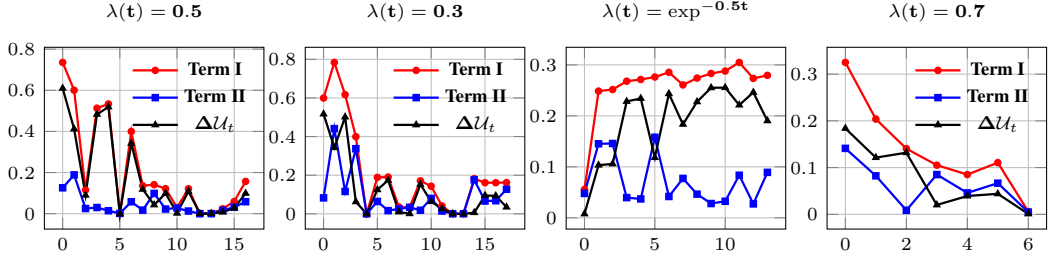

\begin{wrapfigure}{r}{0.5\linewidth}
    \centering

    % Top TikZ/PGFPlots Figure
    % \resizebox{0.9\linewidth}{!}{\input{AuthorKit25/AnonymousSubmission/LaTeX/plots_lambda.tex}}
    % \caption{Evolution of $\lambda(t)$ across training.}
    % \label{fig:lambda-evolution}

    % \vspace{1em}

    % Middle image
    \includegraphics[width=\linewidth]{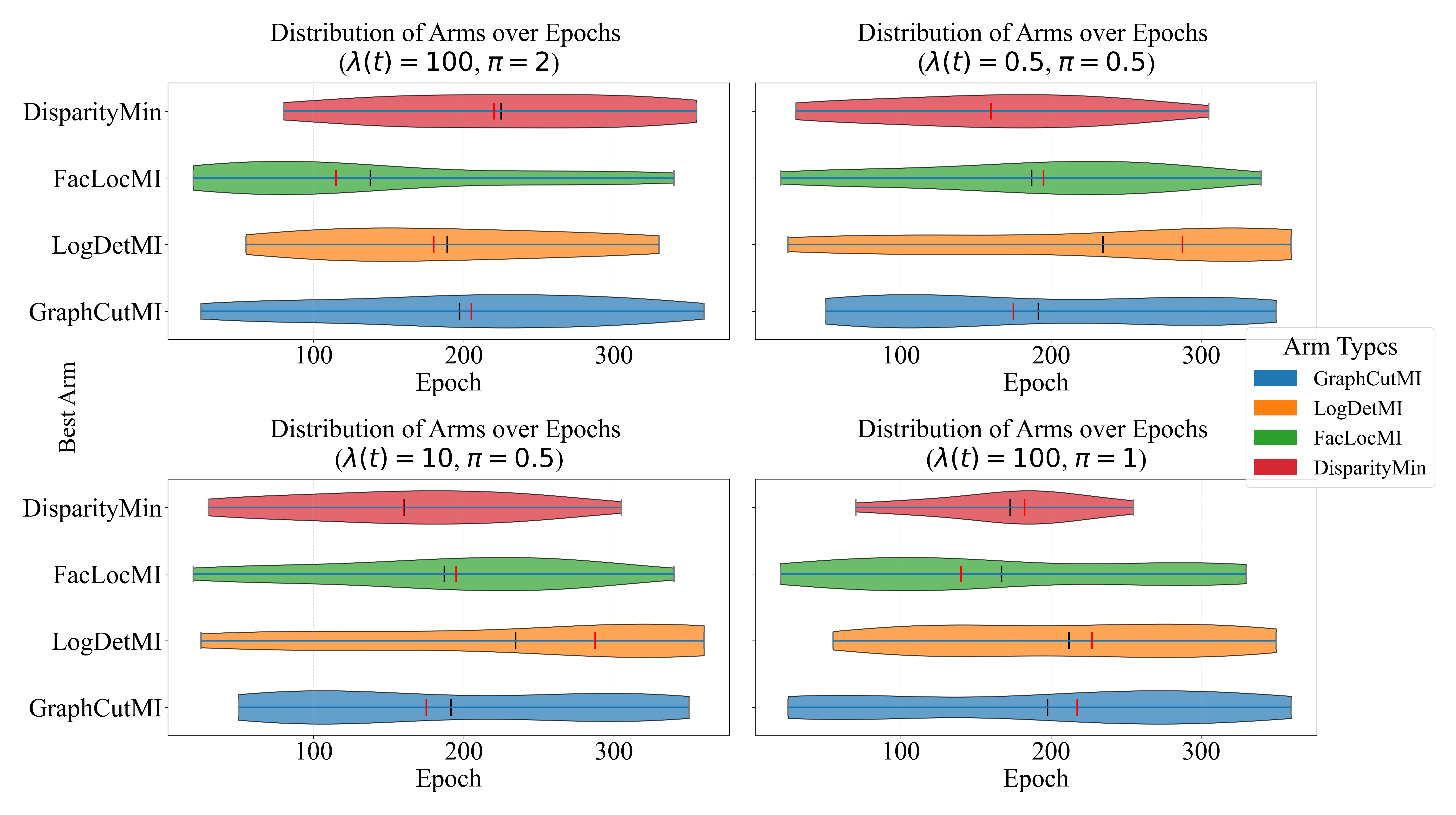}
    \caption{Arm selection distribution over epochs on CIFAR-100. \textbf{\xxpurple{\emph{Diversity based submodular functions}}} become increasingly active during training.}

    \label{fig:best-arm-violin-grid}

    \vspace{1em}

    % Bottom image
    \includegraphics[width=\linewidth]{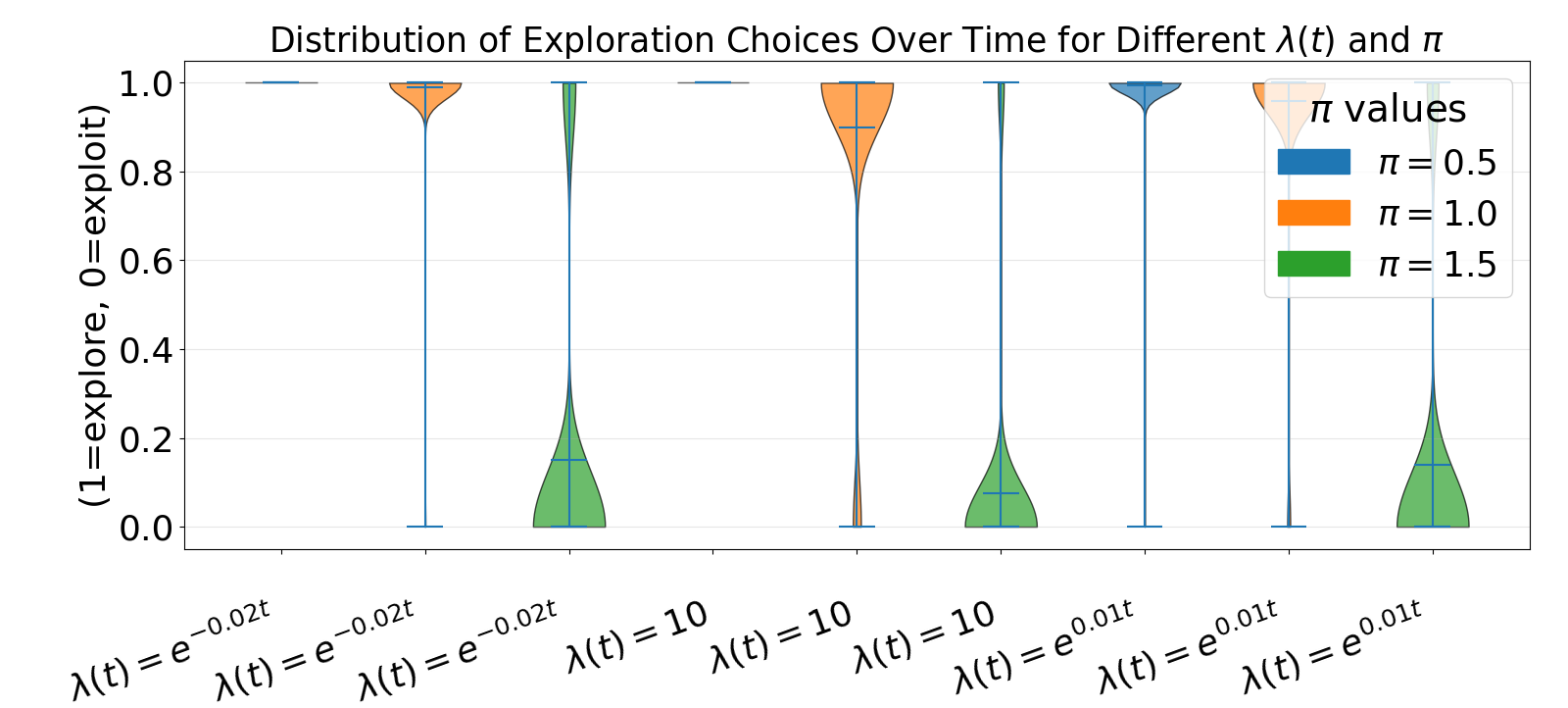}
    \caption{Cumulative \textbf{\xblue{\emph{exploration}}} vs. \textbf{\xxgreen{\emph{exploitation}}} choices over time on \textsc{CIFAR-100}.}
    \label{fig:explore-exploit}
    
    \begin{tabular}{l c}
        \toprule
    \textbf{Computation Breakdown} & \textbf{Average time} \\
        \midrule
        \textbf{Gradient Computation} & 630 ms \\
       \textbf{ Submodular Maximization} & 0.8 ms \\
        \textbf{Total time for Subset Selection} & 640 ms \\
        \bottomrule
    \end{tabular}
    \captionof{table}{Runtime breakdown showing submodular selection adds negligible overhead.}
    \label{tab:runtime-table}
    
\end{wrapfigure}

Formally, $\lambda(t)$: \textbf{ (\xblue{\emph{Exploration}} \xbluedark{\emph{Dampening}})} modulates the inertia of exploration. Larger values induce slower increases in $\bXi_t$, prolonging stochastic exploration across arms, while  smaller values accelerate convergence to greedy selection. On the other hand \textbf{$\pi(t)$ (\xblue{\emph{Exploration}} \xbluelight{\emph{Sharpness}})} controls the curvature of the annealing schedule. \xxgreen{\emph{High $\pi(t)$ enforces an abrupt shift to exploitation}}, \xblue{\emph{while low $\pi(t)$ yields smoother, prolonged exploration phases}}.
As shown in Figure \ref{fig:best-arm-violin-grid}, for any $\lambda(t)$, increasing $\pi$ leads to a higher degree of exploitation. For effective learning, the policy must exploit frequently while retaining sufficient exploration to ensure coverage of the state space. $\pi = 1.5$ offers a suitable trade-off—predominantly exploiting with occasional exploration—whereas $\pi = 1.0$ explores too uniformly and $\pi = 0.5$ almost always explores. Hence, $\pi = 1.5$ emerges as the most effective choice.

\textbf{Computational overhead of Submodular Optimization:} Submodular maximization is NP-hard, but most practical solvers use the greedy algorithm, which guarantees a $(1 - 1/e)$ approximation~\citep{nemhauser1978analysis}.  Table \ref{tab:runtime-table} discusses the tradeoff incurred for the submodular maximization problem w.r.t overall subset selection that involves gradient computation. In our LLM fine-tuning setup on \textsc{MMLU}, \textsc{LLaMa2-7b} using LoRA of rank 128, Table \ref{tab:runtime-table} shows that submodular selection takes 0.8~ms on average, while gradient computation takes 630~ms—a 800$\times$ gap. 

Thus, gradient computation remains the primary bottleneck, and submodular selection adds negligible overhead.

\section{Conclusion}

We introduce \(\methodprop\), a bandit-guided framework for online submodular subset selection that provides a principled alternative to traditional curriculum learning paradigms. By dynamically optimizing a utility-driven reward function, \(\methodprop\) effectively balances the trade-off between accuracy and efficiency across diverse training budgets. Our extensive empirical evaluation demonstrates consistent gains over strong state-of-the-art baselines on multiple benchmarks. Future work will focus on extending the proposed greedy utility metric to train neural scoring models, thereby enabling scalable and adaptive subset selection in large-scale pretraining regimes.

\section{Acknowledgements}
We thank the anonymous reviewers for their constructive feedback and insightful suggestions that helped improve the quality of this work. PC acknowledges the Microsoft Research India PhD Award and Prime Minister Research Fellowship to support this research work. GR thanks Bank of Baroda Chair Professorship. We also acknowledge the computing resources provided by the Department of Computer Science and Engineering at IIT Bombay. In addition, we are exceptionally grateful to the \emph{BharatGen Initiative}  \footnote{BharatGen: \url{http://bharatgen.tech/}} for providing compute resources for conducting large scale language model experiments. Finally, we thank our colleagues and collaborators for valuable discussions and feedback throughout the course of this research process.

\addtocontents{toc}{\protect\setcounter{tocdepth}{2}}  % Re-enable ToC entries
    \bibliographystyle{plain}
    \bibliography{ref}
    \newpage
    
\newpage
\section*{NeurIPS Paper Checklist}

\begin{enumerate}

\item {\bf Claims}
    \item[] Question: Do the main claims made in the abstract and introduction accurately reflect the paper's contributions and scope?
    \item[] Answer: \answerYes{} % Replace by \answerYes{}, \answerNo{}, or \answerNA{}.
    \item[] Justification: The main claims in the abstract and introduction are consistent with the technical contributions and empirical results presented in the paper. We clearly state our proposed method, theoretical foundations, and experimental validation, and these are substantiated in the body of the work without overstatement or omission.

\item {\bf Limitations}
    \item[] Question: Does the paper discuss the limitations of the work performed by the authors?
    \item[] Answer: \answerYes{} % Replace by \answerYes{}, \answerNo{}, or \answerNA{}.
    \item[] Justification: Yes we have discussed the limitations of our work at specific portions of the paper, and have also added in Appendix

\item {\bf Theory assumptions and proofs}
    \item[] Question: For each theoretical result, does the paper provide the full set of assumptions and a complete (and correct) proof?
    \item[] Answer: \answerYes{} % Replace by \answerYes{}, \answerNo{}, or \answerNA{}.
    \item[] Justification: The theoretical proofs are provided in the Appendix \ref{Appendix:TheoreticalSection}

    \item {\bf Experimental result reproducibility}
    \item[] Question: Does the paper fully disclose all the information needed to reproduce the main experimental results of the paper to the extent that it affects the main claims and/or conclusions of the paper (regardless of whether the code and data are provided or not)?
    \item[] Answer: \answerYes{} % Replace by \answerYes{}, \answerNo{}, or \answerNA{}.
    \item[] Justification: The full codebase is provided along with the supplementary material 

\item {\bf Open access to data and code}
    \item[] Question: Does the paper provide open access to the data and code, with sufficient instructions to faithfully reproduce the main experimental results, as described in supplemental material?
    \item[] Answer: \answerYes{} % Replace by \answerYes{}, \answerNo{}, or \answerNA{}.
    \item[] Justification:  The full codebase is provided along with the supplementary material along with running instructions commands in the Readme. Further, we test our algorithm on open source datasets only which we have cited sufficiently and have provided links in the codebase Readme file.

\item {\bf Experimental setting/details}
    \item[] Question: Does the paper specify all the training and test details (e.g., data splits, hyperparameters, how they were chosen, type of optimizer, etc.) necessary to understand the results?
    \item[] Answer: \answerYes{} % Replace by \answerYes{}, \answerNo{}, or \answerNA{}.
    \item[] Justification: All specific training details are specified in the Appendix \ref{Appendix: Experiment Details}

\item {\bf Experiment statistical significance}
    \item[] Question: Does the paper report error bars suitably and correctly defined or other appropriate information about the statistical significance of the experiments?
    \item[] Answer: \answerYes{} % Replace by \answerYes{}, \answerNo{}, or \answerNA{}.
    \item[] Justification: We report std error of most our results over 3 runs per baseline on average.

\item {\bf Experiments compute resources}
    \item[] Question: For each experiment, does the paper provide sufficient information on the computer resources (type of compute workers, memory, time of execution) needed to reproduce the experiments?
    \item[] Answer: \answerYes{} % Replace by \answerYes{}, \answerNo{}, or \answerNA{}.
    \item[] Justification: All specific training and compute details are specified in the Appendix \ref{Appendix: Experiment Details}
    
\item {\bf Code of ethics}
    \item[] Question: Does the research conducted in the paper conform, in every respect, with the NeurIPS Code of Ethics \url{https://neurips.cc/public/EthicsGuidelines}?
    \item[] Answer: \answerYes{} % Replace by \answerYes{}, \answerNo{}, or \answerNA{}.
    \item[] Justification: Yes we have reviewed the Code of Ethics Guidelines.

\item {\bf Broader impacts}
    \item[] Question: Does the paper discuss both potential positive societal impacts and negative societal impacts of the work performed?
    \item[] Answer: \answerNo{} % Replace by \answerYes{}, \answerNo{}, or \answerNA{}.
    \item[] Justification: We have not discussed any potential societal impacts (neither positive nor negative). We do hope that since we are able to show significant efficiency improvement both across different modalities (especially in LLM settings) this may be of siginificant potential impact for Large scale LLM training.
    
\item {\bf Safeguards}
    \item[] Question: Does the paper describe safeguards that have been put in place for responsible release of data or models that have a high risk for misuse (e.g., pretrained language models, image generators, or scraped datasets)?
    \item[] Answer: \answerNA{} % Replace by \answerYes{}, \answerNo{}, or \answerNA{}.
    \item[] Justification: In this work, we are not releasing any generative models or new datasets.

\item {\bf Licenses for existing assets}
    \item[] Question: Are the creators or original owners of assets (e.g., code, data, models), used in the paper, properly credited and are the license and terms of use explicitly mentioned and properly respected?
    \item[] Answer: \answerYes{} % Replace by \answerYes{}, \answerNo{}, or \answerNA{}.
    \item[] Justification: 

\item {\bf New assets}
    \item[] Question: Are new assets introduced in the paper well documented and is the documentation provided alongside the assets?
    \item[] Answer: \answerYes{} % Replace by \answerYes{}, \answerNo{}, or \answerNA{}.
    \item[] Justification: 

\item {\bf Crowdsourcing and research with human subjects}
    \item[] Question: For crowdsourcing experiments and research with human subjects, does the paper include the full text of instructions given to participants and screenshots, if applicable, as well as details about compensation (if any)? 
    \item[] Answer: \answerNA{} % Replace by \answerYes{}, \answerNo{}, or \answerNA{}.
    \item[] Justification: \answerNA{}

\item {\bf Institutional review board (IRB) approvals or equivalent for research with human subjects}
    \item[] Question: Does the paper describe potential risks incurred by study participants, whether such risks were disclosed to the subjects, and whether Institutional Review Board (IRB) approvals (or an equivalent approval/review based on the requirements of your country or institution) were obtained?
    \item[] Answer: \answerNA{} % Replace by \answerYes{}, \answerNo{}, or \answerNA{}.
    \item[] Justification: \answerNA{}

\item {\bf Declaration of LLM usage}
    \item[] Question: Does the paper describe the usage of LLMs if it is an important, original, or non-standard component of the core methods in this research? Note that if the LLM is used only for writing, editing, or formatting purposes and does not impact the core methodology, scientific rigorousness, or originality of the research, declaration is not required.
    %this research? 
    \item[] Answer: \answerNA{}{} % Replace by \answerYes{}, \answerNo{}, or \answerNA{}.
    \item[] Justification: Our paper's methodology is not involved in  LLM usage and nor is the  experimental pipeline.

\end{enumerate}

    \onecolumn \par
    \noindent
    \rule{\textwidth}{1pt}
    \begin{center}
        \large\textbf{Supplementary Material: \papertitle}
    \end{center}
    \par
    \noindent
    \rule{\textwidth}{0.4pt}
% In main.tex or at the end of your main file
\appendix
\addappheadtotoc
\doparttoc
% Make appendix table of contents smaller
{\smaller
  \tableofcontents
}
    \renewcommand{\thesection}{\Alph{section}} % Change section numbering to A, B, C...
    \renewcommand{\thesubsection}{\Alph{section}.\arabic{subsection}} % Change subsection numbering to A.1, A.2...

    \newpage
\onecolumn
\par\noindent\rule{\textwidth}{1pt}
\begin{center}
\large\textbf{Supplementary Material: 
\papertitle}
\end{center}
\par\noindent\rule{\textwidth}{0.4pt}

% \begin{appendices}
% \section{My First Appendix}
% \subsection{My First Subsection}
% \subsubsection{Theory} 
% \section{My Second Appendix}
% \tableofcontents 
% \end{appendices}

\section{Organization of the Appendix}

The appendix is organized as follows. 
Section~\ref{Supplementary:Impact} provides a summary of the impact of our work.Section~\ref{Supplementary:Notation Summary} provides a summary of the notation used throughout the paper. Section~\ref{Appendix:TheoreticalSection} presents our main theoretical results. Section~\ref{Appendix: Experiment Details} outlines the experimental setup and implementation details for both vision and language model tasks. Section \ref{related work} discusses additional related work. Section \ref{Appendix:SubmodDetails} describes the various submodular functions employed in our experiments.

\section{Notation Summary}\label{Supplementary:Notation Summary}

\begin{table}[h!]
\centering
\caption{Table of Notations}
\begin{tabular}{>{\hspace{0pt}}m{0.15\linewidth}|>{\hspace{0pt}}m{0.217\linewidth}|>{\hspace{0pt}}m{0.608\linewidth}}
% \begin{tabular}{>{\hspace{0pt}}m{0.15\linewidth}|>{\hspace{0pt}}m{0.22\linewidth}|>{\hspace{0pt}}m{0.58\linewidth}}
\hline
\textbf{Topic} & \textbf{Notation} & \textbf{Explanation} \\
\hline

% Dataset Indices
\multirow{5}{\linewidth}{\textbf{ Data (sub)Sets Indices}} & $\boldsymbol{\mathcal{D}}_\texttt{train}$ & Entire Training Set consisting of $n$ instances \\
& {\cellcolor[rgb]{0.902,0.902,0.902}}$\boldsymbol{\mathcal{D}}_\texttt{val}$ & {\cellcolor[rgb]{0.902,0.902,0.902}}Entire Validation Set consisting of $m$ instances \\
& {\cellcolor[rgb]{0.902,0.902,0.902}} $\mathbf{z}_i$ & {\cellcolor[rgb]{0.902,0.902,0.902}}$i$-th training instance in a batch\\
& $\B_t$ & Denotes the full sized $t$-th step train minibatch : $\{\boldsymbol{x}_p\}_{p=1}^{|\B_t|}$\\ & {\cellcolor[rgb]{0.902,0.902,0.902}}$\B_t^\texttt{val}$ & {\cellcolor[rgb]{0.902,0.902,0.902}}Denotes the full sized $t$-th step validation minibatch\\
& $\mathcal{B}_{t}^{(<i)}$ &  Denotes the $t$-th step minibatch being constructed uptil $\boldsymbol{x}_{i-1}$ i.e. $\{\boldsymbol{x}_p\}_{p=1}^{i-1}$ \\
& {\cellcolor[rgb]{0.902,0.902,0.902}} $\S^{\texttt{opt}}_{(a_t)}$ & {\cellcolor[rgb]{0.902,0.902,0.902}} Optimal subset obtained when  submodular function $f^{(a_t)}$ is applied \\ 
\hline

% Parameters  
\multirow{5}{\linewidth}{\textbf{Parameters}} 
& {\cellcolor[rgb]{0.902,0.902,0.902}} $\btheta^*$ & {\cellcolor[rgb]{0.902,0.902,0.902}} Optimal model parameter (vector) \\
& {\cellcolor{white}} $\btheta_t$ & {\cellcolor{white}} Model parameter at $t^\text{th}$ step \\
& {\cellcolor[rgb]{0.902,0.902,0.902}} $\btheta_{t+1}$ & {\cellcolor[rgb]{0.902,0.902,0.902}} Model parameter at  $(t+1)^\text{th}$ step \\
\hline

% Loss Function
\multirow{5}{\linewidth}{\textbf{Loss Function}} 
& $\ell$ & Strongly convex instance-wise loss function \\
& {\cellcolor[rgb]{0.902,0.902,0.902}} $\boldsymbol{\mathcal{L}}_t$ & {\cellcolor[rgb]{0.902,0.902,0.902}} Total loss over mini-batch \\
& $\U_t(\B_t; z^{\text{val}}_t)$ & Utility metric capturing validation loss drop for a particular validation data point \\
& $\U_t(\B_t; B^{\text{val}}_t)$ & Aggregated utility metric over validation set \\
\hline

% Hyperparameters
\multirow{4}{\linewidth}{\textbf{Hyperparams}} 
& {\cellcolor[rgb]{0.902,0.902,0.902}} $\lambda(t)$ & {\cellcolor[rgb]{0.902,0.902,0.902}} (\textit{Exploration Dampening}) modulates the inertia of exploration \\  
& {\cellcolor{white}} $\boldsymbol{\vartheta}$ & {\cellcolor{white}} Reward function \\  
& {\cellcolor[rgb]{0.902,0.902,0.902}} $\bXi_t$ & {\cellcolor[rgb]{0.902,0.902,0.902}} Exploration-exploitation threshold \\  
& $\boldsymbol{\mathcal{F}}_\texttt{sub}^{\text{div}}, \boldsymbol{\mathcal{F}}_\texttt{sub}^{\text{repr}}$ & Diversity/representative function subsets \\  
& {\cellcolor[rgb]{0.902,0.902,0.902}} $\zeta \sim \text{Uniform}(0,1)$ & {\cellcolor[rgb]{0.902,0.902,0.902}} Random sample for trade-off \\  
& $\pi(t)$ & (\textit{Exploration Sharpness}) controls the curvature of the annealing schedule rule \\  
\hline

\end{tabular}
\end{table}

\section{Implementation Details}
\subsection{Details about model architectures used}

\subsubsection{Vision Model Architecture Details:} The ResNet18 model architecture begins with a \textit{basic block}, which is composed of two main sections.
 The first section consists of a convolution layer followed by a batch normalization layer, and then a ReLU activation function.
 The second section similarly comprises a convolution layer followed by batch normalization.
This entire \textit{basic block} is repeated twice for each of the four layers in the network. These layers progress with input dimensions of $[64, 128, 256, 512]$ to form the complete ResNet18 architecture.

\subsubsection{Language Model Architecture Details} The LLaMA-2-7B model is a decoder-only transformer comprising approximately 7 billion parameters. It includes 32 transformer layers, each built with pre-normalization using RMSNorm and employing the SwiGLU activation function. The self-attention mechanism uses multi-head causal attention with 32 heads and a hidden dimensionality of 4096. Rotary positional embeddings (RoPE) are applied to the query and key vectors within each attention head. The model begins with a learned token embedding layer and concludes with a tied output projection layer to predict the next token.
Mistral-7B-v0.3 is architecturally similar to LLaMA-2-7B, also featuring 32 transformer layers and a 4096-dimensional hidden state, but introduces several efficiency-focused modifications. It uses grouped-query attention (GQA) with 8 query groups across 32 heads, improving inference throughput. Furthermore, Mistral replaces full causal attention with sliding-window attention to handle long contexts more efficiently. As with LLaMA, it utilizes RoPE for positional encoding and SwiGLU activations. These optimizations maintain strong modeling performance while enabling greater scalability in both training and inference settings.

\subsection{Details on submodular functions implementation}

We provide detailed formulations of the specific submodular functions employed as arms in our experiments in Section~\ref{Appendix:SubmodDetails}. From an implementation standpoint, each submodular arm operates on a similarity kernel computed over the set of instances within a given batch. This kernel, typically represented as a symmetric positive semi-definite matrix, encodes pairwise affinities between samples based on their embedding representations (e.g., cosine similarity or RBF kernel). Once the similarity structure is established, any submodular function can be instantiated over this ground set—such as facility location, log-determinant, or graph-cut functions—depending on the desired coverage, diversity, or representativeness property being optimized.

To operationalize this, we leverage the \texttt{Submodlib} library\footnote{\url{https://submodlib.readthedocs.io/en/latest/}}
, an open-source framework maintained by the Decile organization\footnote{\url{https://decile.org/}}
, which provides efficient and modular implementations of a wide family of submodular functions. The library supports both dense and sparse similarity representations and includes greedy as well as lazy-greedy optimization routines, enabling scalable computation even for large batch sizes.

\subsection{Gradient Computation}

Computing full-model gradients in modern deep networks is computationally prohibitive due to the extremely high dimensionality of parameter spaces—often exceeding billions of parameters for large vision or language models. Moreover, for the purpose of subset selection, what is typically required is not the full parameter gradient but an informative proxy that captures the \emph{relative contribution} of individual samples to the model’s training dynamics. 

Following this motivation, we adopt \emph{partial-gradient approximations} that preserve discriminative signal while substantially reducing computational cost. Specifically, for vision models, we compute gradients only with respect to the \emph{last linear classification layer}, as in \citep{ash2020deep}. This choice leverages the empirical observation that gradients in earlier layers are highly correlated and redundant, and that last-layer gradients retain sufficient information to distinguish hard, redundant, or noisy samples based on their contribution to the decision boundary. 

For large language models (LLMs), computing full backpropagation across all transformer layers is infeasible. We therefore restrict gradient computation to \emph{Low-Rank Adaptation (LoRA)} adapter parameters (rank 128), following the setup of \citep{wang2024greats}. This approach not only reduces memory and compute overhead by several orders of magnitude but also captures localized curvature information relevant to the fine-tuning or instruction-following objective. Since LoRA adapters are trained in the low-dimensional subspace most sensitive to task adaptation, their gradients provide a faithful and low-noise estimate of per-sample learning signals. 

Importantly, both approximations maintain \emph{gradient informativeness} under the assumption that the selected subspace (last layer or adapter) spans the most discriminative directions of parameter updates. Prior empirical evidence (see \citep{ash2020deep, wang2024greats}) shows that subset selection, influence estimation, and sample reweighting methods computed in these reduced spaces closely match those computed with full gradients. In our experiments, we verify that this approximation incurs negligible performance degradation while providing up to \(30\times\) faster per-batch computation. Thus, the proposed gradient computation scheme achieves a favorable balance between computational efficiency and fidelity of learning signal for submodular subset selection.

\paragraph{Warm-starting Data Selection.}\label{Warm Start} 
A common challenge in data subset selection methods lies in the instability of early-stage gradients. During the initial epochs of training, model parameters are far from any local minimum, and per-sample gradients tend to be highly noisy and uninformative. Consequently, performing subset selection too early can result in biased or suboptimal subsets that fail to represent the underlying data distribution or learning dynamics. To mitigate this issue, for the image experiments, we conduct a \emph{warm-start} strategy, wherein the model is first trained for a small number of epochs on the full dataset before invoking any subset selection procedure. 

Concretely, for each algorithm considered in this paper (i.e., $\methodprop$ \textsc{GradMatch}, \textsc{GradMatchPB}, \textsc{Craig}, \textsc{CraigPB}, and \textsc{Glister},), we include a warm-start variant. Let \( T \) denote the total number of training epochs and \( k \) the subset size. We define two quantities: \( T_f \) (the number of full-training epochs prior to subset selection) and \( T_s \) (the number of epochs during which subset selection is active). We set these in proportion as
\(
T_s = \kappa T, T_f = \frac{T_s k}{n},
\)
where \( n \) is the total number of training samples and \( \kappa \in (0,1] \) is the fraction of total training epochs used for subset selection. This parametrization ensures that the effective compute budget remains comparable across methods, while allowing early-stage training to stabilize the model representation before adaptive data selection begins.

Empirically, we observe that performing a few epochs of full-data warm-up (\(T_f\)) consistently improves convergence stability and downstream accuracy across all subset selection algorithms. The warm-start phase enables the gradient space to form a meaningful geometry, allowing the submodular or gradient-based selection objectives to more accurately identify informative and diverse samples. In contrast, starting selection from random initialization often leads to premature overfitting or unstable subset composition due to noisy or poorly conditioned gradients.

Setting \( T_f \) too large, however, diminishes the benefit of subset selection, as the model effectively performs full training with minimal adaptive sampling. In this limit, the behavior approaches that of the full-batch baseline with early stopping, which we include as a control setting in our experiments. Thus, the warm-start scheme provides a principled balance between computational efficiency and representational stability—retaining the benefits of subset-based training while ensuring robust and smooth convergence.

\subsection{Evaluation metrics}

\textbf{Image Classification}: For image classification experiments, we report the standard \emph{test accuracy} as the primary performance metric, measured as the proportion of correctly classified samples on the held-out validation or test split. This metric provides a direct and interpretable indicator of the model’s generalization performance under different subset selection strategies. 

In addition to accuracy, we evaluate the \emph{computational efficiency} of our method by comparing the total training time required to reach convergence across different selection policies. To ensure a fair comparison, all other hyperparameters—including optimizer configuration, learning rate schedule, batch size, and data augmentations—are held fixed across runs. The only varying factor is the subset selection mechanism applied at each training step. 

We define the \emph{speedup} metric with respect to the baseline model trained using full-batch selection (i.e., without any submodular or adaptive sampling). Formally, if \( T_{\text{full}} \) denotes the wall-clock training time for the full-batch model and \( T_{\text{sub}} \) denotes the time under our submodular selection strategy, the speedup is given by \(
\text{Speedup} = \frac{T_{\text{full}}}{T_{\text{sub}}}.
\)
A higher speedup thus indicates a more efficient training regime, achieved without sacrificing downstream accuracy. In practice, we observe consistent gains in training efficiency—typically in the range of \(3\times\)--\(8\times\)—depending on the dataset and the choice of submodular objective, confirming that adaptive selection substantially reduces redundant gradient computations while maintaining comparable predictive performance.

\section{Experimental Setup Details}\label{Appendix: Experiment Details}

\subsection{Software and Hardware}\label{Appendix:Software_and_Hardware}

\textbf{Vision Experiments}
All experiments were conducted using Python 3.10.13 and PyTorch 2.1.2. Our proposed methods, $\methodprop$ and $\methodprop$-Batch, along with their corresponding ablations, were trained on NVIDIA RTX A6000 GPUs (48 GB). Baseline methods, including $\baselinerho$ and $\baselineboss$, were also trained using the same GPU configuration to ensure comparability.

For reference, a typical training run of our ResNet18-based model on an RTX A6000 consists of 300 epochs, with each epoch averaging approximately one minute (excluding certain baselines). Model checkpointing is employed to retain only the best-performing model based on validation accuracy, as well as the final model. Running multiple training jobs concurrently on the same GPU incurs only a slight overhead in training time due to resource contention.

\subsection{Language Model Experiments} \label{Appendix:Hyp}
We experiment for the LLM finetuning setup using a \textsc{RANDOM} subset of 9 datasets from MMLU, and on TydiQA. 
We choose Sociology, Policy, History, Anatomy, ML, Ehics, Genetics, High School Biology, High School Chemistry.
All language model experiments, including both our proposed methods and the baselines, were conducted using 8 NVIDIA H100 GPUs. Additionally, Weights \& Biases (WandB) \footnote{https://wandb.ai/site/}{wandb} was used to manage and monitor all experiments. For all experiments we take batch size of 16, initial learning rate of 2e-5 using adam optimizer with default state, finetuned on 10\% of LESS\citep{xia2024less} version of OpenWebText.

\textbf{Additional Experiment Results}: For MMLU we also showcase additional experiments on LLaMa2-7b and Mistral-7b for TydiQA, later in the appendix. 

Mathematical definitions of the submodular objectives used as arms are provided in Appendix~\ref{Appendix:SubmodDetails}. For this experiment, each arm is a mutual information variant of a classical submodular function, designed to maximize $I_f(X; Q) = f(X) + f(Q) - f(X \cup Q)$, where $X$ is the candidate training set, $Q$ is the validation set, and $f$ is a base submodular function (Facility-Location, Graph-Cut, or Log-Determinant).

We use mutual information forms to ensure the selected subset is explicitly conditioned on the current validation set, making the acquisition process adaptive to the downstream task. Features for $X$ and $Q$ are derived either from Sentence-BERT embeddings or from gradient vectors, with the latter shown to yield better alignment with task-specific error signals and improved selection performance.

\subsection{Vision Model Experiments}\label{Appendix:vision}

The experimental setup was configured to evaluate the proposed method on several datasets, including \textsc{CIFAR-10}, \textsc{CIFAR-100}, Tiny-ImageNet-200, and \textsc{SVHN}. The data module used a batch size of 128, with four workers for data loading. The model architecture employed was ResNet18~\citep{he2016deep}, and the training followed a curriculum-based mode, progressively utilizing 10\%, 30\%, and 50\% of the training data. The optimizer used was SGD with a learning rate of 0.05, momentum of 0.9, weight decay of 0.0005, and Nesterov momentum enabled. 

For all our settings (across different baselines and dataset), we consider ResNet18 ~\citep{he2016deep} as our primary model with the following architecture and training details:

In our training setup, we employed batch-wise Nesterov accelerated gradient descent with a batch size of 128. The optimization configuration included a learning rate of 0.05 and a momentum of 0.9, alongside a cosine-annealing scheduler.

Across all dataset comparisons, we set the submodular function budget $\beta$ to 10\%, 30\%, and 50\% of the entire batch size.

\textbf{Dataset Specifics.} We conduct experiments across a range of standard vision benchmarks. For the \textbf{\textsc{MNIST}} dataset, we use 60{,}000 training instances, 10{,}000 test instances, and 10{,}000 validation instances, with training proceeding until full convergence, typically around \textbf{200 epochs}. On \textsc{CIFAR-10}, we use 50{,}000 training instances, 10{,}000 test instances, and 10{,}000 validation instances, with models trained for up to \textbf{300 epochs}. For \textbf{\textsc{CIFAR-100}}, we similarly use 50{,}000 training examples spread across 100 classes (500 per class), and a validation set of 10{,}000 examples (100 per class). The \textbf{\textsc{SVHN}} dataset comprises 73{,}257 training images across 10 classes with variable class frequencies, and a validation set of 26{,}032 images distributed proportionally. Finally, for \textbf{\textsc{TinyImageNet}}, we use 100{,}000 training images across 200 classes (500 per class), and a validation set of 10{,}000 images (50 per class), covering the same label space as the training data.

\subsection{Baseline Training Details}

We compare our method $\methodprop$ with several state of the art baselines for our experiments:

\textbf{$\baselinemaxloss$}:  Within each training batch, the loss is computed for every example. A fixed fraction (e.g., top-K\%) of samples with the highest per-example loss is selected for gradient computation and model update. This assumes that high-loss samples are currently mis-predicted and could contribute the most to updating the decision boundary.

\textbf{$\baselinegradnorm$ }: The $l_2$ norm of each example's per-sample gradient i.e. $\|\nabla_{\btheta} \mathcal{L}(\bz;\btheta)\|_2$ is computed, and a subset with the highest norms is selected for each batch. This prioritizes examples inducing the largest parameter updates under the current model, helping direct learning toward sensitive or uncertain regions.

\textbf{$\baselinerho$}: Each example's \textit{reducible loss} is estimated as the difference between the current model’s loss and its \textit{irreducible loss}, the latter approximated by a small auxiliary model trained on held-out clean data. Examples with high reducible loss are selected, as they are considered learnable but not yet learned, making them useful for continued training. For LLM experiments, we use \texttt{LLaMa3-7b-instruct} as our auxiliary model.

\textbf{$\baselinesbert$}: In this case, each training and validation example is encoded into a sentence embedding using a pre-trained SBERT model. Cosine similarity is computed between training examples and the validation set, and those with the highest average similarity are selected. This favors examples that align semantically with the validation distribution.

For \textbf{GREATS}, each example's impact on the loss is approximated using a first-order Taylor expansion of the objective. For model parameters $\theta$ and a batch $\{x_1, \dots, x_n\}$, gradients $\nabla \mathcal{L}(x_i;\theta)$ are used to estimate loss reduction. A greedy selection strategy then chooses the subset expected to most decrease validation loss under this approximation.

For fair comparison against our model  we considered the configuration where subset selection happens at every epoch for all the 3 baselines with a \textit{lazy} optimizer. Due to our multi-class image classification setup we utilise \textit{CrossEntropy} loss for our model training.

\textbf{$\baselinerho$}:  We begin by training an irreducible model on the specific task for 100 epochs. Subsequently, we precompute the irreducible losses for the training set, which are required for the target model training. During the target model training phase, we train the model for 300 epochs across the \textsc{CIFAR-10}, \textsc{CIFAR-100}, \textsc{SVHN}, and \textsc{TinyImageNet} datasets, using subset ratios of 0.1, 0.3, and 0.5. We employ the ResNet-18 architecture for both the irreducible model and the target model. For training, we use the SGD optimizer with Nesterov accelerated gradient descent, a batch size of 128, and the following configuration: a learning rate of 0.05, momentum of 0.9, and a cosine-annealing scheduler. One observation we made is that RHO Loss converges to reasonably good accuracy within a few epochs, but further training does not significantly improve performance, and it fails to reach the accuracy levels of other state-of-the-art (SOTA) baselines.

\textbf{$\baselineboss$}: For $\baselineboss$, to select the subset, we first initialized a model by training it using the full dataset. With the help of the training dynamics obtained from the initialized model, we calculated the difficulty score for each sample that is used to select the subset. We evaluated the selected subset keeping the subset fixed and using it to train a new \textsc{RANDOM} initialized model. For the difficulty score, we experimented using the EL2N score because it can be efficiently computed early on during training.We trained the model for 300 epochs across the \textsc{CIFAR-10}, \textsc{CIFAR-100}, \textsc{SVHN} and \textsc{TinyImageNet} datasets, using subset ratios of 0.1, 0.3, and 0.5. We employed ResNet18 model using SGD with a learning rate  of 0.1, and momentum of 0.9 with a batch size of 128. 

\subsection{Comparison between \textsc{DINO} and Gradient-Based Features for Submodular Selection}

To evaluate how closely our feature representations must align with the downstream objective, we compared two ways of representing each training item when optimising submodular acquisition functions (and their mutual–information variants):

\begin{figure}[h!]
    \centering
    \includegraphics[width=0.8\linewidth]{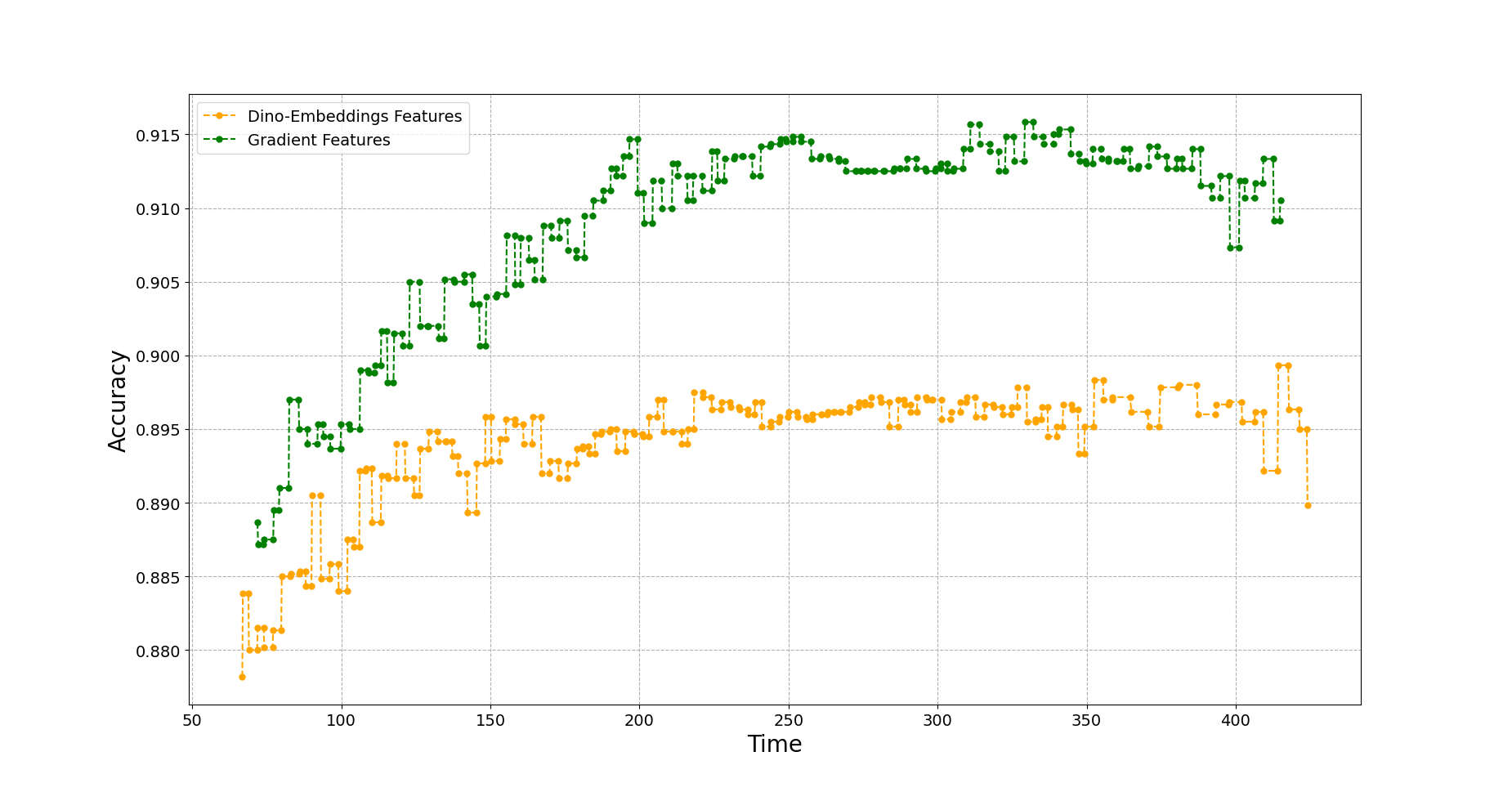}
        \caption{Comparison of Fashion-\textsc{MNIST} with DINO-embeddings, and with Gradient Features for submodular optimization}
    \label{fashion-cmp}
\end{figure}

\begin{enumerate}[leftmargin=1.5em]
    \item \textbf{\textsc{DINO} embeddings}.  
    We obtain a fixed $d$-dimensional feature vector for every image by running it through a frozen \textsc{DINO} vision transformer, exactly as one would use a CLIP encoder. These representations are task–agnostic and remain static throughout training.

    \item \textbf{Gradient-based features}.  
    At every training step we compute the gradient of the scalar loss with respect to the parameters of the final layer.  We average these per-example gradients within the mini-batch to form a single vector\footnote{Using the batch-wise average was consistently superior to concatenating per-example gradients, and last-layer gradients are sufficient while keeping the computation inexpensive.}.  For mutual–information objectives, validation gradients serve as the query features.
\end{enumerate}

Figure \ref{fashion-cmp} shows that gradient features yield substantially higher test accuracy on \textsc{Fashion-\textsc{MNIST}} across all subset sizes: they encode task-specific error signals that guide the submodular optimiser toward examples most useful for loss reduction, whereas \textsc{DINO} embeddings capture only generic visual similarity.  Hence, directly leveraging gradients as features is the more effective choice for data subset selection in this setting.

\subsection{Fisher Information Matrix}\label{appendix:fisher information matrix}

    \textbf{Fisher Information Matrix Approximation} An alternative and
    potentially more informative approach to approximating the Hessian is
    through the use of the Fisher Information Matrix (FIM) \citep{fujita2022fisher}.
    The FIM provides insights into the curvature of the loss landscape and can
    serve as a useful surrogate for the Hessian. While the exact computation of the
    FIM requires calculating an expectation, which can be computationally
    intensive, it can be efficiently approximated using an exponential moving average
    of the outer product of the gradients from the validation data points.

   \begin{wrapfigure}
        {r}{0.35\linewidth}
        \begin{equation}
            \widehat{\boldsymbol{\mathcal{H}}}_{\mathcal{B}_t}^{(t)}=
            \begin{cases}
                \frac{1}{|\mathcal{B}_0|}\sum_{\mathbf{z}_i \in \mathcal{B}_0}\mathbf{\Omega}_{i},        & \text{if }t = 0 \\
                (1 - \alpha) \widehat{\boldsymbol{\mathcal{H}}}_{\mathcal{B}_{t-1}}^{(t-1)}+               \\
                \alpha \frac{1}{|\mathcal{B}_t|}\sum_{\mathbf{z}_i \in \mathcal{B}_t}\mathbf{\Omega}_{i}, & \text{else}
            \end{cases}
        \end{equation}
    \end{wrapfigure}

    Let $\mathbf{\Omega}_{i}:=\boldsymbol{g}(\mathbf{z}_{i}, \boldsymbol{\theta}_{t}
    ) \boldsymbol{g}(\mathbf{z}_{i}, \boldsymbol{\theta}_{t})^{\top}$ denote the
    outer product of the gradient for the $i$th data point in the current batch
    $\B_{t}$. The approximate FIM $\widehat{\boldsymbol{\mathcal{H}}}_{\mathcal{B}_t}
    ^{(t)}$ at time step $t$ for the current mini-batch $\mathcal{B}_{t}$ can be
    computed recursively as

    where $\widehat{\boldsymbol{\mathcal{H}}}_{\mathcal{B}_{t-1}}^{(t-1)}$ is the
    approximate FIM at the previous time step $t-1$ for the mini-batch $\mathcal{B}
    _{t-1}$, and $\alpha \in (0, 1]$ is the smoothing parameter for the
    exponential moving average. This recursive formulation provides a computationally
    efficient approach to approximating the Hessian, particularly in high-dimensional
    settings where direct Hessian computation is prohibitively expensive.

\section{Additional Experiments}

\subsection{Sensitivity of Validation Dataset Configuration}

To better understand the influence of validation data composition on model performance, we investigate how sensitive the algorithm is to different validation set configurations. In particular, we examine what occurs during the early stages of training when the validation dataset includes harder samples that the model has not yet adequately learnt. This analysis provides a deeper perspective on how the distribution and difficulty of validation examples can affect optimization dynamics and generalization, thereby strengthening the empirical validity of our findings.

Specifically, we aim to understand the following questions:

\begin{itemize}[leftmargin=*,itemsep=0.1pt]
\item \textbf{Q1:} To what extent is the algorithm’s performance sensitive to the configuration and composition of the validation dataset?
\item \textbf{Q2:} How does the presence of harder, yet-unlearned samples in the validation set during early training stages affect convergence and generalization?
\end{itemize}

\begin{minipage}{0.52\linewidth}

To better understand this issue, we conducted a controlled experiment on \textsc{CIFAR-100} (300 epochs) where we varied the hardness of the validation dataset. Hardness was measured via gradient norm where the gradient is calculated w.r.t model parameter at that time step. In accordance with other literature, a crude way to approximate difficulty of a sample is to check if the gradient norm is high. (higher gradient norm $\sim$ harder example). We compared four validation subset configurations:
\end{minipage}
\hfill
\begin{minipage}{0.44\linewidth}
\centering
\setlength{\tabcolsep}{5pt}
\renewcommand{\arraystretch}{1.15}
\begin{tabular}{lccc}
\toprule
\textbf{Validation Subset} & \textbf{10\%} & \textbf{20\%} & \textbf{30\%} \\
\midrule
Easiest & 72.3 & 74.4 & 76.03 \\
EasyHard & 73.1 & 74.5 & 76.4 \\
HardEasy & 72.29 & 74.6 & 76.2 \\
Hardest & 71.31 & 74.3 & 75.9 \\
\bottomrule
\end{tabular}
\captionof{table}{Final test accuracies under different validation subset configurations.}
\end{minipage}

\begin{itemize}[leftmargin=*,itemsep=0.1pt]
    \item \textbf{Easiest:} Lowest gradient norms
    \item \textbf{EasyHard:} Easy samples early, hard samples later
    \item \textbf{HardEasy:} Hard samples early, easy samples later
    \item \textbf{Hardest:} Highest gradient norms
\end{itemize}

Each configuration was evaluated at validation subset sizes of 10\%, 20\%, and 30\%.

\vspace{4pt}

\begin{tcolorbox}[
colback=gray!20, colframe=gray!20,
  left=2mm, right=2mm, top=1mm, bottom=1mm,
  boxsep=0pt, width=1\linewidth, before skip=2pt, after skip=2pt,
  enlarge left by=0mm, enlarge right by=0mm]

\textbf{Observations:} \\
Validation sets composed of the most difficult examples tend to yield lower performance, particularly when smaller subsets are used. This decline likely stems from noisy or overly pessimistic reward signals during the early stages of training. In contrast, mixed validation configurations such as \textbf{EasyHard} and \textbf{HardEasy} generally perform best, indicating that a balanced distribution of sample difficulty across training can enhance robustness. These findings suggest that further exploring how validation sample difficulty and ordering interact especially through the lens of curriculum learning could be a promising direction for future work. Importantly, even validation sets containing difficult samples early in training do not lead to instability or model collapse.

\end{tcolorbox}

\subsection{Effect of Submodular Functions Individually and \textsc{RANDOM} Selection over Arms}

To assess the contribution of the multi-armed bandit formulation in our framework, we perform ablation experiments on \textsc{CIFAR-100} (10\% subset, 300 epochs) under two simplified settings: 
(a) using a single, fixed submodular arm throughout training (i.e., no bandit-driven adaptation), and 
(b)\textsc{RANDOM}ly selecting an arm at each round (i.e., no explore--exploit balancing). 
These ablations isolate the effect of static versus dynamic subset selection policies on training efficiency and generalization.

% ---- left text ----
\begin{minipage}{0.48\linewidth}
We compare various submodular selection strategies that define the reward structure of the curriculum. 
\textit{Representative} functions (e.g., \textbf{GraphCut}, \textbf{FacilityLocation}) promote coverage and ensure the selected subset reflects the global data distribution, while 
\textit{Diversity}-oriented functions (e.g., \textbf{DisparitySum}, \textbf{LogDeterminant}) encourage maximal dissimilarity among chosen samples. 
These functions capture different inductive biases: representation versus decorrelation. 
\end{minipage}
\hfill
% ---- right table ----
\begin{minipage}{0.48\linewidth}
\centering
\setlength{\tabcolsep}{5pt}
\renewcommand{\arraystretch}{1.15}
\begin{tabular}{l c}
\toprule
\textbf{Selection Strategy} & \textbf{Accuracy (\%)} \\
\midrule
DisparitySum (Div., Static) & 68.6 \\
FacilityLocation (Rep., Static) & 72.0 \\
LogDeterminant (Div., Static) & 71.1 \\
GraphCut (Rep., Static) & 72.6 \\
Random arm per round & 72.0 \\
\textbf{\textsc{OnlineSubmod} (ours)} & \textbf{73.6} \\
\bottomrule
\end{tabular}
\captionof{table}{Performance comparison of individual and \textsc{RANDOM} arm selection strategies on \textsc{CIFAR-100}. $\methodprop$ adaptively balances diversity and representativeness over training epochs.}
\end{minipage}

Our proposed \textbf{$\methodprop$} method dynamically alternates between these functions through an adaptive explore--exploit policy governed by the bandit controller. 
This dynamic weighting enables the model—whether a \textbf{ResNet-18} backbone or a small \textbf{LLM fine-tuning setup}—to exploit high-yield submodular arms while continually exploring others that may improve validation loss or perplexity.

\vspace{3pt}
\begin{tcolorbox}[
colback=gray!20, colframe=gray!20,
left=2mm, right=2mm, top=1mm, bottom=1mm,
boxsep=0pt, width=1\linewidth, before skip=4pt, after skip=4pt,
enlarge left by=0mm, enlarge right by=0mm,
]
\textbf{Observations:} \\ 
Different submodular functions show complementary but limited strengths. Coverage-based methods such as \textit{GraphCut} and \textit{FacilityLocation} converge quickly early in training, while diversity-based ones like \textit{DisparitySum} and \textit{LogDet} encourage better generalization but can become unstable when applied uniformly. Random arm selection gives reasonable results, suggesting that diversity matters, but it lacks feedback to adapt to validation performance. In contrast, the proposed \textbf{$\methodprop$} approach adjusts arm selection based on past rewards, maintaining a stable balance between exploration and exploitation. This supports our main hypothesis that adaptive, reward-driven selection leads to more robust and generalizable outcomes than static or random strategies.
\end{tcolorbox}

\newpage

\subsection{Additional Experiments on Vision datasets}

Table~\ref{tab:vision_results} summarizes batchwise data selection results across multiple vision datasets. Across all budgets and datasets, $\methodprop$ consistently achieves the highest test accuracy while maintaining competitive or lower training time compared to prior methods. Notably, it surpasses strong baselines such as \textsc{GRADMATCH}, \textsc{MILO}, and \textsc{RHO-LOSS}, particularly at low data budgets (10\%–30\%), indicating superior sample efficiency and adaptivity under constrained training regimes. The improvement is most pronounced on \textsc{CIFAR100} and \textsc{TinyImageNet}, where the model benefits from dynamic online selection over diverse feature manifolds. In contrast, static coreset-based methods (e.g., \textsc{CRAIG}, \textsc{GLISTER}) exhibit slower convergence and lower performance as the budget increases. 

Here, \textcolor{red}{red} highlights the best result and \textcolor{blue}{blue} denotes the second-best result for each setting. Overall, these results confirm that $\methodprop$ provides a strong trade-off between accuracy and computational efficiency across datasets of varying complexity.

\begin{table}[h!] \label{additional:results}
\centering
\caption{(Batchwise) Data Selection Results on Vision Datasets}
\label{tab:vision_results}
\small
\begin{tabular}{cccccccc} 
\hline
\textbf{Dataset}      & \textbf{Selection Strategy}        & \multicolumn{3}{c}{\textbf{Test accuracy (\%)}}                                & \multicolumn{3}{c}{\textbf{Training time (hrs)}}                    \\ 
\hline
                      & Budget(\%)                         & 10\%                      & 30\%                     & 50\%                    & 10\%                 & 30\%                 & 50\%                  \\ 
\hline
\textbf{CIFAR10}      & FULL (skyline for test accuracy)   & 95.09                     & 95.09                    & 95.09                   & 1.73                 & 1.73                 & 1.73                  \\
                      &\textsc{RANDOM} (skyline for training time) & 77.49                      & 89.62                     & 91.85                   & 0.29                 & 0.75                 & 0.85                   \\
                      &\textsc{CRAIG}                              & 90.07                     & 92.4                     & 93.12                   & 0.26                 & 0.62                  & 1.54                   \\
\multicolumn{1}{l}{}  & GLISTER                            & 91.15                     & 92.18                    & 92.65                    & 0.38                 & 1.05                 & 1.34                   \\
                      & GRADMATCH                          & \textcolor{blue}{92.27}   & \textcolor{blue}{93.28}   & 93.15  & 0.42                  & 0.95                  & 1.21                   \\
\multicolumn{1}{l}{}  & MILO                               & 92.25 & 93.21                    & 94.16                   & 0.34                  & 0.85                 & 0.89                   \\
                      & RHO-LOSS                           & 90.16                     & 91.54                    & \textcolor{blue}{94.03}                   & 0.76                  & 1.13                  & 1.54                   \\
                      &$\baselineboss$                               & 91.64                     & 93.04                    & 93.8  & 0.36                 & 0.94                 & 1.18                   \\
                      & \textbf{$\methodprop$ (ours)}      & \textcolor{red}{92.44}    & \textcolor{red}{93.75}  & \textcolor{red}{94.18}                   & 0.32                 & 0.87                 & 0.83                  \\ 
\hline
\textbf{CIFAR100}     & FULL (skyline for test accuracy)   & 76.8                      & 76.8                    & 76.8                   & 1.52                & 1.52                & 1.52                 \\
                      &\textsc{RANDOM} (skyline for training time) & 35.03                     & 61.93                    & 64.67                       & 0.15               & 0.42                & 0.78                  \\
                      &\textsc{CRAIG}                              & 67.25                      & 72.38                    & 73.12                   & 0.31             & 0.62               & 1.12                  \\
\multicolumn{1}{l}{}  & GLISTER                            & 64.27     & 72.36     & 74.62    & 0.26 & 0.57 & 1.3  \\
                      & GRADMATCH                          & 68.34   & 74.63  & 72.36 & 0.22               & 0.48               & 1.22                  \\
\multicolumn{1}{l}{}  & MILO                               & \textcolor{blue}{72.36}      & 74.66     & 75.60    & 0.15 & 0.44 & 0.82  \\
                      & RHO-LOSS                           & 71.37                        & \textcolor{blue}{74.82}                    & \textcolor{blue}{75.74}                      & 0.53                 & 0.86                 & 1.46                  \\
                      &$\baselineboss$                               & 71.73                     & 73.77                    & 75.41                   & 0.27                & 0.53                & 0.85                 \\
                      & \textbf{$\methodprop$ (ours)}      & \textcolor{red}{73.67}    & \textcolor{red}{75.46}   & \textcolor{red}{75.78}  & 0.165                & 0.47                & 0.82               \\ 
\hline
\textbf{\textsc{SVHN}}         & FULL (skyline for test accuracy)   & 96.49                     & 96.49                    & 96.49                   & 6.436                & 6.436                & 6.436                 \\
                      &\textsc{RANDOM} (skyline for training time) & 93.47                   & 95.31                    & 95.84                   & 0.6383               & 1.90                 & 3.19                  \\
                      &\textsc{CRAIG}                              & 95.27   & 96.15 & 96.40                   & 0.934              & 2.332                & 4.17                  \\
\multicolumn{1}{l}{}  & GLISTER                            & 95.52      & 95.69     & 96.42    & 0.83 & 2.42 & 4.26  \\
                      & GRADMATCH                          & \textcolor{blue}{95.64}   & \textcolor{red}{96.4}  & \textcolor{blue}{96.42}  & 0.789               & 2.398                & 4.19                  \\
\multicolumn{1}{l}{}  & MILO                               & 95.62      & 96.36     & 96.41    & 0.69 & 2.09 & 3.25  \\
                      & RHO-LOSS                           & 94.64                     & 94.27                    & 94.85                   & 1.08                 & 2.56                 & 3.94                  \\
                      &$\baselineboss$                               & 94.31                     & 95.75                    & 96.01                   & 0.76               & 2.39               & 3.56                \\
                      & \textbf{$\methodprop$ (ours)}      & \textcolor{red}{95.68}                    & \textcolor{blue}{96.38}                    & \textcolor{red}{96.46} & 0.68               & 2.12               & 3.28                \\ 
\hline
\textbf{\textsc{TinyImageNet}} & FULL (skyline for test accuracy)   & 64.36                     & 64.36                    & 64.36                   & 15.4                & 15.4                & 15.4                 \\
                      &\textsc{RANDOM} (skyline for training time) & 19.61                     & 35.68                    & 43.84                   & 1.82                 & 4.92                 &       6.12           \\
                      &\textsc{CRAIG}                              & 52.42                    & 55.56                    & 61.48                   &   3.27                   &      6.46                &       9.23                \\
\multicolumn{1}{l}{}  & GLISTER                            & 51.54      & 56.37     & 62.15    & 2.84 & 5.93 & 9.47  \\
                      & GRADMATCH                          & 52.63                    & 58.19                    & 61.93                   &     2.63                 &      5.94                &      7.24                 \\
\multicolumn{1}{l}{}  & MILO                               & 53.24      & 59.36     & \textcolor{blue}{62.28}    & 1.81 & 4.97 & 6.16  \\
                      & RHO-LOSS                           & \textcolor{blue}{54.46}                     & 59.78                    & 62.15                   &   3.16                  &  6.38                     &         7.94              \\
                      &$\baselineboss$                               & 52.63                     & \textcolor{blue}{60.17}                    & 62.13                   & 2.85               & 5.47               & 7.26                \\
                      & \textbf{$\methodprop$ (ours)}      & \textcolor{red}{55.3}    & \textcolor{red}{60.74}   & \textcolor{red}{62.58}  & 1.84                 & 5.16                 & 6.14                
\end{tabular}
\end{table}

\newpage

\subsection{Additional Experiments on Large Language Models}

We evaluate the evolution of test perplexity during pretraining on the \texttt{MMLU} benchmark using the \texttt{LLAMA-2-7B} model under different \textbf{online batch selection strategies}. 
Each method is trained under identical hyperparameter and compute budgets to ensure fair comparison. 

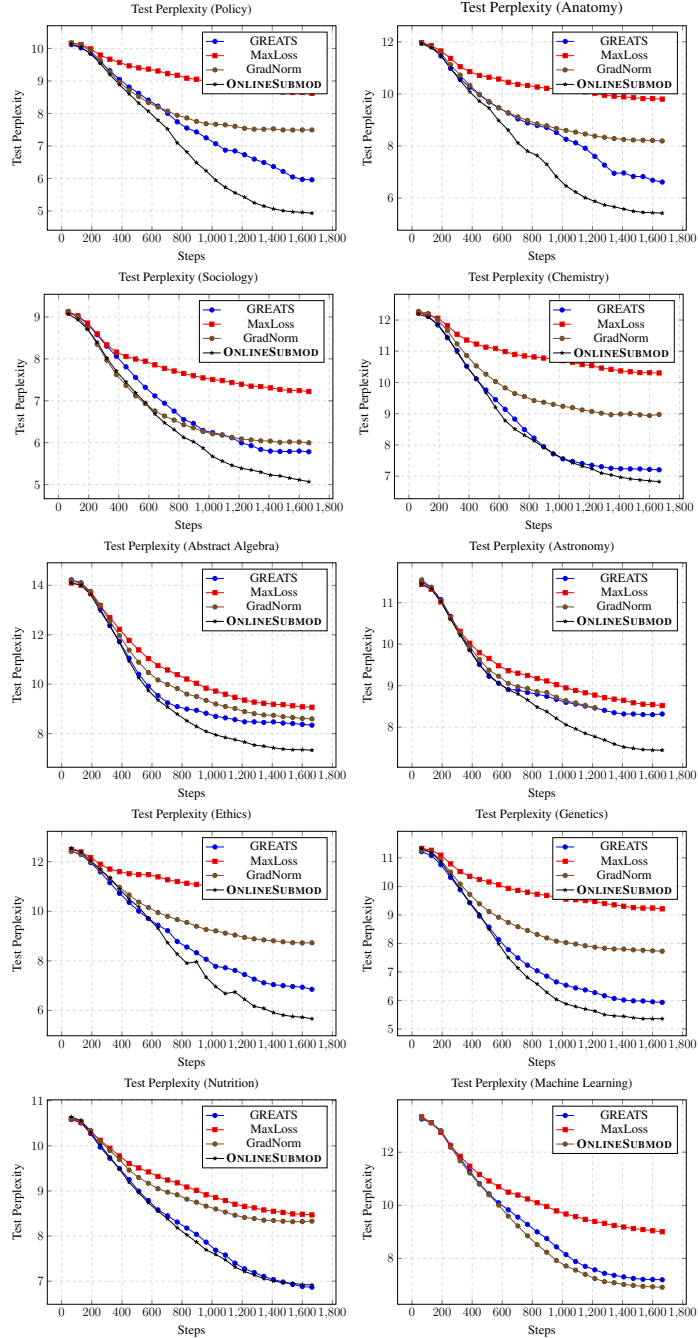
\begin{figure}[h!]
 \centering

 \begin{minipage}[t]{\textwidth}
 \centering
 
 \scalebox{0.4}{\begin{tikzpicture}
    \begin{axis}[
        width=0.8\textwidth,
        height=0.6\textwidth,
        xlabel={Steps},
        ylabel={Test Perplexity},
         title={Test Perplexity (Policy)},
        grid=both,
        grid style={dashed, gray!30},
        legend pos=north east,
        font=\large, % <--- Increase font size here
        label style={font=\large}, % Optional: fine-grained control
        tick label style={font=\large}, % Optional: tick font
        title style={font=\large}, % Optional: title font
        legend style={font=\large} % Optional: legend font
    ]
        \addplot table [x index=0, y index=1] {AuthorKit25/AnonymousSubmission/LaTeX/LLMs_results/results_dat/2shot_poli_GREATS.dat};
        \addlegendentry{GREATS}
        \addplot table [x index=0, y index=1] {AuthorKit25/AnonymousSubmission/LaTeX/LLMs_results/results_dat/2shot_poli_MaxLoss.dat};
        \addlegendentry{MaxLoss}
        \addplot table [x index=0, y index=1] {AuthorKit25/AnonymousSubmission/LaTeX/LLMs_results/results_dat/2shot_poli_GradNorm.dat};
        \addlegendentry{GradNorm}
        \addplot table [x index=0, y index=1] {AuthorKit25/AnonymousSubmission/LaTeX/LLMs_results/results_dat/2shot_poli_onlineSubmod.dat};
        \addlegendentry{\textbf{$\methodprop$}}
    \end{axis}
\end{tikzpicture}}
 \scalebox{0.4}{%
 \begin{tikzpicture}
    \begin{axis}[
        width=0.8\textwidth,
        height=0.6\textwidth,
        xlabel={Steps},
        ylabel={Test Perplexity},
         title={Test Perplexity (Anatomy)},
        grid=both,
        grid style={dashed, gray!30},
        legend pos=north east,
        font=\large, % <--- Increase font size here
        label style={font=\large}, % Optional: fine-grained control
        tick label style={font=\large}, % Optional: tick font
        title style={font=\Large}, % Optional: title font
        legend style={font=\large} % Optional: legend font
    ]
        \addplot table [x index=0, y index=1] {AuthorKit25/AnonymousSubmission/LaTeX/LLMs_results/results_dat/2shot_anatomy_GREATS.dat};
        \addlegendentry{GREATS}
        \addplot table [x index=0, y index=1] {AuthorKit25/AnonymousSubmission/LaTeX/LLMs_results/results_dat/2shot_anatomy_MaxLoss.dat};
        \addlegendentry{MaxLoss}
        \addplot table [x index=0, y index=1] {AuthorKit25/AnonymousSubmission/LaTeX/LLMs_results/results_dat/2shot_anatomy_GradNorm.dat};
        \addlegendentry{GradNorm}
        \addplot table [x index=0, y index=1] {AuthorKit25/AnonymousSubmission/LaTeX/LLMs_results/results_dat/2shot_anatomy_onlineSubmod.dat};
        \addlegendentry{\textbf{$\methodprop$}}
    \end{axis}
\end{tikzpicture}}
 \end{minipage}
 % LINE 2
 \begin{minipage}[t]{\textwidth}
 \centering
 \scalebox{0.4}{%
 \begin{tikzpicture}
    \begin{axis}[
        width=0.8\textwidth,
        height=0.6\textwidth,
        xlabel={Steps},
        ylabel={Test Perplexity},
         title={Test Perplexity (Sociology)},
        grid=both,
        grid style={dashed, gray!30},
        legend pos=north east,
        font=\large, % <--- Increase font size here
        label style={font=\large}, % Optional: fine-grained control
        tick label style={font=\large}, % Optional: tick font
        title style={font=\large}, % Optional: title font
        legend style={font=\large} % Optional: legend font
    ]
        \addplot table [x index=0, y index=1] {AuthorKit25/AnonymousSubmission/LaTeX/LLMs_results/results_dat/2shot_redo2_soc_GREATS.dat};
        \addlegendentry{GREATS}
        \addplot table [x index=0, y index=1] {AuthorKit25/AnonymousSubmission/LaTeX/LLMs_results/results_dat/2shot_redo2_soc_MaxLoss.dat};
        \addlegendentry{MaxLoss}
        \addplot table [x index=0, y index=1] {AuthorKit25/AnonymousSubmission/LaTeX/LLMs_results/results_dat/2shot_redo2_soc_GradNorm.dat};
        \addlegendentry{GradNorm}
        \addplot table [x index=0, y index=1] {AuthorKit25/AnonymousSubmission/LaTeX/LLMs_results/results_dat/2shot_redo2_soc_onlineSubmod.dat};
        \addlegendentry{\textbf{$\methodprop$}}
    \end{axis}
\end{tikzpicture}}
 \scalebox{0.4}{%
 \begin{tikzpicture}
    \begin{axis}[
        width=0.8\textwidth,
        height=0.6\textwidth,
        xlabel={Steps},
        ylabel={Test Perplexity},
        title={Test Perplexity (Chemistry)},
        grid=both,
        grid style={dashed, gray!30},
        legend pos=north east,
        font=\large, % <--- Increase font size here
        label style={font=\large}, % Optional: fine-grained control
        tick label style={font=\large}, % Optional: tick font
        title style={font=\large}, % Optional: title font
        legend style={font=\large} % Optional: legend font
    ]
        \addplot table [x index=0, y index=1] {AuthorKit25/AnonymousSubmission/LaTeX/LLMs_results/results_dat/2shot_redo2_hchem_GREATS.dat};
        \addlegendentry{GREATS}
        \addplot table [x index=0, y index=1] {AuthorKit25/AnonymousSubmission/LaTeX/LLMs_results/results_dat/2shot_redo2_hchem_MaxLoss.dat};
        \addlegendentry{MaxLoss}
        \addplot table [x index=0, y index=1] {AuthorKit25/AnonymousSubmission/LaTeX/LLMs_results/results_dat/2shot_redo2_hchem_GradNorm.dat};
        \addlegendentry{GradNorm}
        \addplot table [x index=0, y index=1] {AuthorKit25/AnonymousSubmission/LaTeX/LLMs_results/results_dat/2shot_redo2_hchem_onlineSubmod.dat};
        \addlegendentry{\textbf{$\methodprop$}}
    \end{axis}
\end{tikzpicture}}
 \end{minipage}
  % LINE 3
 \begin{minipage}[t]{\textwidth}
 \centering
 \scalebox{0.4}{%
 \begin{tikzpicture}
    \begin{axis}[
        width=0.8\textwidth,
        height=0.6\textwidth,
        xlabel={Steps},
        ylabel={Test Perplexity},
         title={Test Perplexity (Abstract Algebra)},
        grid=both,
        grid style={dashed, gray!30},
        legend pos=north east,
        font=\large, % <--- Increase font size here
        label style={font=\large}, % Optional: fine-grained control
        tick label style={font=\large}, % Optional: tick font
        title style={font=\large}, % Optional: title font
        legend style={font=\large} % Optional: legend font
    ]
        \addplot table [x index=0, y index=1] {AuthorKit25/AnonymousSubmission/LaTeX/LLMs_results/results_dat/alg_GREATS.dat};
        \addlegendentry{GREATS}
        \addplot table [x index=0, y index=1] {AuthorKit25/AnonymousSubmission/LaTeX/LLMs_results/results_dat/alg_MaxLoss.dat};
        \addlegendentry{MaxLoss}
        \addplot table [x index=0, y index=1] {AuthorKit25/AnonymousSubmission/LaTeX/LLMs_results/results_dat/alg_GradNorm.dat};
        \addlegendentry{GradNorm}
        \addplot table [x index=0, y index=1] {AuthorKit25/AnonymousSubmission/LaTeX/LLMs_results/results_dat/alg_onlineSubmod.dat};
        \addlegendentry{\textbf{$\methodprop$}}
    \end{axis}
\end{tikzpicture}}
 \scalebox{0.4}{%
 \begin{tikzpicture}
    \begin{axis}[
        width=0.8\textwidth,
        height=0.6\textwidth,
        xlabel={Steps},
        ylabel={Test Perplexity},
         title={Test Perplexity (Astronomy)},
        grid=both,
        grid style={dashed, gray!30},
        legend pos=north east,
        font=\large, % <--- Increase font size here
        label style={font=\large}, % Optional: fine-grained control
        tick label style={font=\large}, % Optional: tick font
        title style={font=\large}, % Optional: title font
        legend style={font=\large} % Optional: legend font
    ]
        \addplot table [x index=0, y index=1] {AuthorKit25/AnonymousSubmission/LaTeX/LLMs_results/results_dat/astro_GREATS.dat};
        \addlegendentry{GREATS}
        \addplot table [x index=0, y index=1] {AuthorKit25/AnonymousSubmission/LaTeX/LLMs_results/results_dat/astro_MaxLoss.dat};
        \addlegendentry{MaxLoss}
        \addplot table [x index=0, y index=1] {AuthorKit25/AnonymousSubmission/LaTeX/LLMs_results/results_dat/astro_GradNorm.dat};
        \addlegendentry{GradNorm}
        \addplot table [x index=0, y index=1] {AuthorKit25/AnonymousSubmission/LaTeX/LLMs_results/results_dat/astro_onlineSubmod.dat};
        \addlegendentry{\textbf{$\methodprop$}}
    \end{axis}
\end{tikzpicture}}
 \end{minipage}
   % LINE 4
 \begin{minipage}[t]{\textwidth}
 \centering
 \scalebox{0.4}{%
 \begin{tikzpicture}
    \begin{axis}[
        width=0.8\textwidth,
        height=0.6\textwidth,
        xlabel={Steps},
        ylabel={Test Perplexity},
         title={Test Perplexity (Ethics)},
        grid=both,
        grid style={dashed, gray!30},
        legend pos=north east,
        font=\large, % <--- Increase font size here
        label style={font=\large}, % Optional: fine-grained control
        tick label style={font=\large}, % Optional: tick font
        title style={font=\large}, % Optional: title font
        legend style={font=\large} % Optional: legend font
    ]
        \addplot table [x index=0, y index=1] {AuthorKit25/AnonymousSubmission/LaTeX/LLMs_results/results_dat/eth_onlineSubmod.dat};
        \addlegendentry{GREATS}
        \addplot table [x index=0, y index=1] {AuthorKit25/AnonymousSubmission/LaTeX/LLMs_results/results_dat/eth_MaxLoss.dat};
        \addlegendentry{MaxLoss}
        \addplot table [x index=0, y index=1] {AuthorKit25/AnonymousSubmission/LaTeX/LLMs_results/results_dat/eth_GradNorm.dat};
        \addlegendentry{GradNorm}
        \addplot table [x index=0, y index=1] {AuthorKit25/AnonymousSubmission/LaTeX/LLMs_results/results_dat/eth_GREATS.dat};
        \addlegendentry{\textbf{$\methodprop$}}
    \end{axis}
\end{tikzpicture}}
 \scalebox{0.4}{%
 \begin{tikzpicture}
    \begin{axis}[
        width=0.8\textwidth,
        height=0.6\textwidth,
        xlabel={Steps},
        ylabel={Test Perplexity},
         title={Test Perplexity (Genetics)},
        grid=both,
        grid style={dashed, gray!30},
        legend pos=north east,
        font=\large, % <--- Increase font size here
        label style={font=\large}, % Optional: fine-grained control
        tick label style={font=\large}, % Optional: tick font
        title style={font=\large}, % Optional: title font
        legend style={font=\large} % Optional: legend font
    ]
        \addplot table [x index=0, y index=1] {AuthorKit25/AnonymousSubmission/LaTeX/LLMs_results/results_dat/gen_onlineSubmod.dat};
        \addlegendentry{GREATS}
        \addplot table [x index=0, y index=1] {AuthorKit25/AnonymousSubmission/LaTeX/LLMs_results/results_dat/gen_MaxLoss.dat};
        \addlegendentry{MaxLoss}
        \addplot table [x index=0, y index=1] {AuthorKit25/AnonymousSubmission/LaTeX/LLMs_results/results_dat/gen_GradNorm.dat};
        \addlegendentry{GradNorm}
        \addplot table [x index=0, y index=1] {AuthorKit25/AnonymousSubmission/LaTeX/LLMs_results/results_dat/gen_GREATS.dat};
        \addlegendentry{\textbf{$\methodprop$}}
    \end{axis}
\end{tikzpicture}}
 \end{minipage}
   % LINE 5
 \begin{minipage}[t]{\textwidth}
 \centering
 \scalebox{0.4}{%
 \begin{tikzpicture}
    \begin{axis}[
        width=0.8\textwidth,
        height=0.6\textwidth,
        xlabel={Steps},
        ylabel={Test Perplexity},
         title={Test Perplexity (Nutrition)},
        grid=both,
        grid style={dashed, gray!30},
        legend pos=north east,
        font=\large, % <--- Increase font size here
        label style={font=\large}, % Optional: fine-grained control
        tick label style={font=\large}, % Optional: tick font
        title style={font=\large}, % Optional: title font
        legend style={font=\large} % Optional: legend font
    ]
        \addplot table [x index=0, y index=1] {AuthorKit25/AnonymousSubmission/LaTeX/LLMs_results/results_dat/nut_GREATS.dat};
        \addlegendentry{GREATS}
        \addplot table [x index=0, y index=1] {AuthorKit25/AnonymousSubmission/LaTeX/LLMs_results/results_dat/nut_MaxLoss.dat};
        \addlegendentry{MaxLoss}
        \addplot table [x index=0, y index=1] {AuthorKit25/AnonymousSubmission/LaTeX/LLMs_results/results_dat/nut_GradNorm.dat};
        \addlegendentry{GradNorm}
        \addplot table [x index=0, y index=1] {AuthorKit25/AnonymousSubmission/LaTeX/LLMs_results/results_dat/nut_onlineSubmod.dat};
        \addlegendentry{\textbf{$\methodprop$}}
    \end{axis}
\end{tikzpicture}}
 \scalebox{0.4}{%
 \begin{tikzpicture}
    \begin{axis}[
        width=0.8\textwidth,
        height=0.6\textwidth,
        xlabel={Steps},
        ylabel={Test Perplexity},
         title={Test Perplexity (Machine Learning)},
        grid=both,
        grid style={dashed, gray!30},
        legend pos=north east,
        font=\large, % <--- Increase font size here
        label style={font=\large}, % Optional: fine-grained control
        tick label style={font=\large}, % Optional: tick font
        title style={font=\large}, % Optional: title font
        legend style={font=\large} % Optional: legend font
    ]
        \addplot table [x index=0, y index=1] {AuthorKit25/AnonymousSubmission/LaTeX/LLMs_results/results_dat/ml_GREATS.dat};
        \addlegendentry{GREATS}
        \addplot table [x index=0, y index=1] {AuthorKit25/AnonymousSubmission/LaTeX/LLMs_results/results_dat/ml_MaxLoss.dat};
        \addlegendentry{MaxLoss}
     %   \addplot table [x index=0, y index=1] {AuthorKit25/AnonymousSubmission/LaTeX/LLMs_results/results_dat/ml_GradNorm.dat};
      %  \addlegendentry{GradNorm}
        \addplot table [x index=0, y index=1] {AuthorKit25/AnonymousSubmission/LaTeX/LLMs_results/results_dat/ml_onlineSubmod.dat};
        \addlegendentry{\textbf{$\methodprop$}}
    \end{axis}
\end{tikzpicture}}
 \end{minipage}

 \caption{\textbf{Test perplexity dynamics} on \texttt{LLAMA-2-7B} during training with various \textbf{online batch selection strategies} on \texttt{MMLU}. $\methodprop$ significantly outperforms baselines.}
 \label{fig:llm_analysis}
\end{figure}

\newpage

\section{Details of Submodular Function used in all our training settings}\label{Appendix:SubmodDetails}

We describe here the submodular functions we broadly utilised for all our experiments.

\subsection{Diversity based Submodular Function}

Here we share the details on the diversity based submodular functions we used for our training purposes.

\begin{definition}
\textbf{Log-determinant Function} is a diversity-based submodular function. It is non-monotone in nature. Let $\mathbf{L}$ denote a positive semidefinite kernel matrix and $\mathbf{L_S}$ denote the subset of rows and columns indexed by set $\mathbf{S}$. Log-determinant function $f$ is specified as:
\begin{equation}
    f(\mathbf{S}) = \text{logdet}(\mathbf{L_S})
\end{equation}
\end{definition}

The log-det function models diversity and is closely related to a determinantal point process.

\begin{definition}
\textbf{Disparity Sum Function} characterizes diversity by considering the sum of distances between every pair of points in a subset $\mathbf{S}$. For any two points $i,j \in \mathbf{S}$, let $d_{ij}$ denote the distance between them.
\begin{equation}
    f(\mathbf{S})=\sum_{i, j \in \mathbf{S}}d_{i j}
\end{equation}
The aim is to select a subset $\mathbf{S}$ such that $f(\mathbf{S})$ is maximized. 
\end{definition}

\begin{definition}
\textbf{Disparity Min Function} characterizes diversity by considering the minimum distance between any two non-similar points in a subset $\mathbf{S}$.
\begin{equation}
    f(\mathbf{S})=\min _{i, j \in \mathbf{S}, i \neq j}d_{i j}
\end{equation}
The aim is to select a subset $\mathbf{S}$ such that $f(\mathbf{S})$ is maximized. 
\end{definition}

\subsection{Representative based Submodular Function}

Here we share the details on the representative based submodular functions we used for our training purposes.

\begin{definition}
\textbf{Facility Location Function} characterizes the representativeness in the dataset by considering the minimum distance between any two non-similar points in a subset $\mathbf{S}$.
\begin{equation}
    f(\mathbf{S})=\sum_{i \in \mathcal{V}}\max_{j\in \mathbf{S}}d_{ij}
\end{equation}
The aim is to select a subset $\mathbf{S}$ such that $f(\mathbf{S})$ is maximized. 
\end{definition}

\begin{definition}
\textbf{Graph Cut Function} characterizes representativeness by using the parameter $\lambda$ which governs the tradeoff between representation and diversity. When $\lambda$ becomes large, graph cut function also tries to model diversity in the subset. $\mathbf{S}$.
\begin{equation}
    f(\mathbf{S})=\sum_{i \in \mathcal{V}, j \in \mathbf{S}}d_{ij} - \lambda\sum_{i,j \in \mathbf{S}}d_{ij}
\end{equation}
The aim is to select a subset $\mathbf{S}$ such that $f(\mathbf{S})$ is maximized. 
\end{definition}

\textbf{Submodular Mutual Information}
We first provide a definition of Submodular Mutual Information:

\[
\mathcal{I}_{f}(\A;\B) = f(\A) + f(\B) - f(\A \cup \B)
\]

\begin{definition}
\textbf{Log-Determinant Mutual Information Function} is an instantiation of a submodular mutual information function using a $\texttt{LogDeterminantFunction}$. Let $S_{A, B}$ be the cross-similarity matrix between the items in sets $A$ and $B$. Also, denote $S_{AB} = S_{A \cup B}$. We construct a similarity matrix $S^{\eta}$ (on a base matrix $S$) such that the cross-similarity between $A$ and $Q$ is multiplied by $\eta$ (i.e., $S^{\eta}_{A,Q} = \eta S_{A,Q}$) to control the trade-off between query relevance and diversity. Higher values of $\eta$ ensure greater query-relevance while lower values favor diversity. Using a similarity matrix defined above and with $f(A) = \log\det(S^{\eta}_{A})$, we have:
\begin{equation}
    I_f(A; Q) = \log\det(S_{A}) -\log\det(S_{A} - \eta^2 S_{A,Q}S_{Q}^{-1}S_{A,Q}^T)
\end{equation}
\end{definition}

\begin{definition}[\textbf{Generalized Submodular Mutual Information}]
Let $\Omega$ be a ground set and $\mathcal{V} \subseteq \Omega$ be a domain of interest. Let $f: 2^\Omega \to \mathbb{R}_{\geq 0}$ be a \emph{restricted submodular function}, i.e., submodular when restricted to subsets of $\mathcal{V}$. A \emph{Submodular Mutual Information (SMI)} function defined via such a function $f$ is called a \emph{Generalized Submodular Mutual Information (GMI)} function.

A notable instance of GMI is the \emph{Concave Over Modular (COM)} function~\cite{kothawade2022prism}, defined for subsets $\mathcal{A} \subseteq \mathcal{V}$ and $\mathcal{Q} \subseteq \mathcal{V}'$ as:
\[
I_{f_\eta}(\mathcal{A}; \mathcal{Q}) = \eta \sum_{i \in \mathcal{A}} \psi\left( \sum_{j \in \mathcal{Q}} s_{ij} \right) + \sum_{j \in \mathcal{Q}} \psi\left( \sum_{i \in \mathcal{A}} s_{ij} \right),
\]
where $\eta \in \mathbb{R}_{\geq 0}$ controls the trade-off between query-relevance and diversity, $\psi: \mathbb{R}_{\geq 0} \to \mathbb{R}_{\geq 0}$ is a concave function,$S = [s_{ij}]$ is a kernel similarity matrix such that $s_{ij} = \mathbbm{1}(i = j)$ for $i, j \in \mathcal{V}$ or $i, j \in \mathcal{V}'$.
\end{definition}

\begin{definition}[Facility Location Mutual Information (FL1MI)]
Let $I_f(\mathcal{A}; \mathcal{Q})$ denote a \emph{Submodular Mutual Information (SMI)} function. An instantiation of SMI using the \texttt{FacilityLocationFunction} is known as the \emph{Facility Location Mutual Information (FL1MI)} function.

Formally, given subsets $\mathcal{A} \subseteq \mathcal{V}$ and $\mathcal{Q} \subseteq \mathcal{V}'$, FL1MI is defined as:
\[
I_f(\mathcal{A}; \mathcal{Q}) = \sum_{i \in \mathcal{V}} \min\left( \max_{j \in \mathcal{A}} s_{ij},\; \eta \max_{j \in \mathcal{Q}} s_{ij} \right),
\]
where:
 $\eta \in \mathbb{R}_{\geq 0}$ is a relevance-diversity trade-off parameter,
 $s_{ij}$ denotes similarity between elements $i$ and $j$ in the kernel similarity matrix $S$,
 $\mathcal{V}$ is the candidate set and $\mathcal{V}'$ is the query set domain.
\end{definition}

\begin{definition}[Graph Cut Mutual Information (GCMI)]
Let $I_f(\mathcal{A}; \mathcal{Q})$ denote a \emph{Submodular Mutual Information (SMI)} function. An instantiation of SMI using the \texttt{GraphCutFunction} is called the \emph{Graph Cut Mutual Information (GCMI)} function.

Formally, for subsets $\mathcal{A} \subseteq \mathcal{V}$ and $\mathcal{Q} \subseteq \mathcal{V}'$, GCMI is defined as:
\[
I_f(\mathcal{A}; \mathcal{Q}) = 2\lambda \sum_{i \in \mathcal{A}} \sum_{j \in \mathcal{Q}} s_{ij},
\]
where $\lambda \in \mathbb{R}_{\geq 0}$ controls the scale of mutual information,$s_{ij}$ denotes the similarity between elements $i$ and $j$ in the kernel similarity matrix $S$,$\mathcal{V}$ is the candidate set and $\mathcal{V}'$ is the query set domain.
\end{definition}

\section{Additional Related Work}\label{related work}

\subsection{Online Submodular Maximization}

A growing body of research has advanced our understanding of online submodular maximization under diverse feedback models and constraint classes. A notable contribution is the recent work on Algorithms for Online Submodular Maximization via First-order Regret Bounds~\cite{chen2023firstorder}, which introduces a principled framework leveraging first-order regret bounds from online linear optimization to derive improved guarantees in submodular settings. At each round $t$, the algorithm selects a feasible set $S_t \in \mathcal{C} \subseteq 2^V$, where $\mathcal{C}$ encodes combinatorial constraints such as matroids or cardinality bounds, and observes an adversarially chosen submodular function $f_t$. For monotone submodular functions under matroid constraints, the method achieves a $(1 - c/e - \epsilon)$-approximate regret bound of $\mathcal{O}(kT \log(n/k))$, improving on earlier results by Streeter and Golovin~\cite{streeter2009online}, and Golovin and Krause~\cite{golovin2011adaptive}, even in the absence of curvature (i.e., $c = 1$). For non-monotone unconstrained submodular functions, a novel algorithm based on Blackwell approachability achieves a $1/2$-regret of $\mathcal{O}(n\sqrt{T})$, extending Roughgarden and Wang~\cite{roughgarden2018approximately}.

These developments complement recent advances in bandit and semi-bandit feedback models, including those by Hassani et al.~\cite{hassani2023bandit} and Chen et al.~\cite{chen2023near}, who analyze online submodular optimization in stochastic and adversarial environments, obtaining nearly minimax optimal regret bounds. Related efforts have also explored limited feedback settings with structure-aware exploration strategies (e.g., combinatorial Thompson sampling or optimism-based approaches), enhancing sample efficiency in large-scale decision spaces.

In the full-information setting, the online continuous greedy algorithm of Bian et al.~\cite{bian2017guarantees} offers near-optimal $(1 - 1/e)$-regret for monotone submodular functions under matroid constraints, while extensions by Radakovic and Mitrovic~\cite{radakovic2022delayed} tackle delayed or corrupted feedback. Meanwhile, online versions of lazy greedy~\cite{mirzasoleiman2016fast} and differentiable submodular surrogates~\cite{staib2019distribution} have enabled scalable implementations in real-world domains such as streaming recommendation and dataset summarization.

Although submodular maximization is NP-hard in general, approximation algorithms often yield near-optimal performance in practice across diverse applications~\cite{streeter2009online, golovin2011adaptive}. These include influence maximization, budget-constrained recommendation, and online resource allocation, all of which benefit from the expressive yet structured nature of submodular objectives. As such, the aforementioned theoretical advances not only deepen our algorithmic understanding but also broaden the applicability of online submodular optimization frameworks to practical domains involving limited feedback, combinatorial constraints, and dynamic inputs.

\subsection{Pruning Mechanisms}

Several recent works have explored data pruning and subset selection for efficient training, including $\DPruning$ \citep{maharana2023d2}, $\InfoMax$ \citep{tan2025data}, and $\CCS$ \citep{zheng2022coverage}. While these methods offer valuable insights into coreset selection and dataset reduction, they predominantly operate in static, full-dataset settings, in contrast to our dynamic, batch-level framework.

$\InfoMax$ formulates an objective that can be interpreted as a reformulation of the Graph Cut function, which is monotone submodular, and leverages similarity kernels such as DINO embeddings (for image tasks) or gradient-based features. This aligns conceptually with our approach, where Graph Cut is explicitly implemented as a bandit arm. However, $\InfoMax$ selects samples over the entire dataset in a static manner, whereas our framework is modular and dynamic: an InfoMax-like objective can be treated as a bandit arm and applied in batch-level pruning during training. This flexibility enables more scalable deployment in real-world training pipelines where adaptive, online selection is crucial.

$\DPruning$ frames data subset selection as a subgraph pruning task over the dataset, representing data points as nodes in a similarity graph. Selection is performed using message passing algorithms, which can be computationally intensive and challenging to scale to large datasets. Like $\InfoMax$, $\DPruning$ operates at a static, dataset-wide level, making it less suitable for dynamic batch-level pruning.

Similarly, $\CCS$ addresses static selection by optimizing for both coverage and diversity. A key contribution of CCS is its theoretical characterization of the pruning budget, identifying thresholds beyond which accuracy degradation becomes catastrophic. While our method does not primarily operate at the full-dataset level, we note that analogous effects could, in principle, occur in batch-level selection; investigating such phenomena is a potential avenue for future work.

In summary, although $\InfoMax$, $\DPruning$, and $\CCS$ provide important foundations for data pruning and coreset strategies, they are largely static and dataset-wide in nature. Our approach extends these ideas to a dynamic, scalable setting by leveraging a bandit-driven curriculum where batch-level pruning decisions are guided by validation performance. Moreover, the modularity of our framework allows for seamless integration of alternative reward signals, such as forgetting scores or other criteria discussed in prior works, enabling flexible and adaptive training in large-scale environments.

\newpage

\section{Main Theoretical Results}\label{Appendix:TheoreticalSection}

\subsection{Proof for permutation invariance of Expected Marginal Gain}

\begin{tcolorbox}
    [  colback=gray!15, colframe=gray!15,
  left=2mm, right=2mm, top=1mm, bottom=1mm,
  boxsep=0pt, width=1\linewidth,  before skip=2pt, after skip=2pt, 
    enlarge left by=0mm, enlarge right by=0mm,]
\PermInvariance*
\end{tcolorbox}

\begin{proof}
Let \( S = \mathcal{B}_t^{(<i)} \), with \( |S| = n \), and let \( \mathbf{z}_{\mathsf{val}} = \mathbf{z}_{\mathsf{val}}^t \). 
We denote \( \boldsymbol{g}_i := \boldsymbol{g}_{\boldsymbol{\theta}_t}(\mathbf{z}_i) \), \( \boldsymbol{g}_v := \boldsymbol{g}_{\boldsymbol{\theta}_t}(\mathbf{z}_{\mathsf{val}}) \), and \( \boldsymbol{\mathcal{H}}_v := \boldsymbol{\mathcal{H}}_{\mathbf{z}_{\mathsf{val}}}(\boldsymbol{\theta}_t) \).

Then the expected marginal gain as per Eq 4 is given by:
\[
\frac{1}{n} \sum_{\mathbf{z}_i \in S} \left[ \eta_t \, \boldsymbol{g}_i \cdot \boldsymbol{g}_v - \eta_t^2 \, \boldsymbol{g}_i^\top \boldsymbol{\mathcal{H}}_v \left( \frac{1}{n} \sum_{\mathbf{z} \in S} \boldsymbol{g}_{\mathbf{z}} \right) \right].
\]

Let \( \bar{\boldsymbol{g}} := \frac{1}{n} \sum_{\mathbf{z} \in S} \boldsymbol{g}_{\mathbf{z}} \). Then:
\[
= \eta_t \, \bar{\boldsymbol{g}} \cdot \boldsymbol{g}_v - \eta_t^2 \left( \frac{1}{n} \sum_{\mathbf{z}_i \in S} \boldsymbol{g}_i^\top \boldsymbol{\mathcal{H}}_v \bar{\boldsymbol{g}} \right)
= \eta_t \, \bar{\boldsymbol{g}} \cdot \boldsymbol{g}_v - \eta_t^2 \, \bar{\boldsymbol{g}}^\top \boldsymbol{\mathcal{H}}_v \bar{\boldsymbol{g}}.
\]

This expression depends only on the multiset \( S \), not the order of its elements. Therefore, for any permutation \( \pi(S) \), the same value holds:
\[
\mathds{E}_{\mathbf{z}_i \in \pi(S)} \left[ \bDelta \mathcal{U}_t(\mathbf{z}_i \mid \pi(S), \mathbf{z}_{\mathsf{val}}) \right]
= \eta_t \, \bar{\boldsymbol{g}} \cdot \boldsymbol{g}_v - \eta_t^2 \, \bar{\boldsymbol{g}}^\top \boldsymbol{\mathcal{H}}_v \bar{\boldsymbol{g}}.
\]

Hence,
\[
\mathds{E}_{\mathbf{z}_i \in S} \left[ \bDelta \mathcal{U}_t(\mathbf{z}_i \mid S, \mathbf{z}_{\mathsf{val}}) \right]
=
\mathds{E}_{\mathbf{z}_i \in \pi(S)} \left[ \bDelta \mathcal{U}_t(\mathbf{z}_i \mid \pi(S), \mathbf{z}_{\mathsf{val}}) \right],
\]
which proves the claim.
\end{proof}

\subsection{Theorem: Capacity-Controlled Risk Convergence Theorem}

\begin{restatable}[\textbf{Capacity-Controlled Risk Convergence}]{theorem}{RegretBoundCapacityControlled}\label{theorem:gyorfi}
[(\cite{gyorfi2002distribution}) \textbf{Theorem 16.3}] Let $\mathcal{M}_{\boldsymbol{\Theta}}$ be a neural network with $\boldsymbol{d}$ parameters belonging to the parameter space $\bTheta$ with an objective to minimize the empirical risk over the training data, $\boldsymbol{\mathcal{D}}=\{(\mathcal{X}_i, \mathcal{Y}_i)\}_{i=1}^n$ where $\mathcal{X}_i \in \mathbb{R}^{m}$ and $\mathcal{Y}$ are almost surely bounded and where $\mathcal{Y}_i = \boldsymbol{\vartheta}(\mathbf{z}_i) \sim \mathcal{N}(\mu_{\mathbf{z}_i}, \sigma_{\mathbf{z}_i})$ where $\boldsymbol{\vartheta} : \mathbb{R}^m \rightarrow \mathbb{R}$ and  $\mathbb{P}$ denotes the data distribution. Then for $\boldsymbol{d}$ large enough, we have the following, for any $\mathfrak{c} > 0$.

\begin{equation}
\mathbb{E}_{\boldsymbol{\mathcal{D}}}
\int_{\mathbf{z}} 
\left\lVert\mathcal{M}_{\boldsymbol{\Theta}}(\mathbf{z}) -
\mathbb{E}\big[\boldsymbol{\vartheta}(\mathbf{z})\big]\right\rVert^2
\,\partial \mathbb{P}(\mathbf{z})
\leq
\mathfrak{c}\sqrt{\frac{\ln(\boldsymbol{d})}{\boldsymbol{d}}}
\end{equation}
\end{restatable}

The above theorem follows from \citep{rawson2021convergence} and \citep{gyorfi2002distribution} and is useful to prove further bounds in our case as below.

\subsection{No-Regret Bounds under Constant $\lambda(\cdot)$}\label{Constant Lambda Proof}

We first state here the main theorem under the following assumptions as stated in our main text:

Let  $\boldsymbol{\tau}^{R}_{(a)}(t)$ denote the number of times the $a$-th submodular function $\bof^{(a)}$ is chosen in the first $t-1$ steps by the uniform branch of the algorithm.

\textbf{Assumption (a)}  (\textbf{Constant Fractional Exploration Dampening}):  
    The exploration dampening parameter \(\lambda(t)\) is time-invariant \(
    \lambda(t) = \epsilon\) where \(\quad \epsilon \in (0,1).\)

\textbf{Assumption (b)} (\textbf{Optimality Gap}):  
    There exists an optimality gap \(\boldsymbol{\varrho}\) such that for every suboptimal arm \(a_t \in \mathscr{A} \setminus \{a^*\}\) : \(
    0 \leq \boldsymbol{\varrho} \leq \bDelta_{(a_t)}(\mathcal{B}_t).
    \)

 \textbf{Assumption (c)}  (\textbf{Fractional Exploration Sharpness}):  The exploration sharpness parameter \(\pi(t)\) is a bounded quantity \(\pi(t) \in (0,1).\)

\textbf{Assumption (d)}  (\textbf{Utility Metric Approximation}):  
    The utility metric \(\mathcal{U}_t(\cdot, \cdot)\) satisfies the approximation bound as per Theorem 2 (Appendix) with constants \(\mathfrak{C}_{(a)}\) for each arm \(a \in \mathscr{A}\) and let \(n_a\) be a specific constant associated with arm $a$ such that Theorem 2 (Appendix) holds true.

\begin{tcolorbox}
    [  colback=gray!15, colframe=gray!15,
  left=2mm, right=2mm, top=1mm, bottom=1mm,
  boxsep=0pt, width=1\linewidth,  before skip=2pt, after skip=2pt, 
    enlarge left by=0mm, enlarge right by=0mm]
\RegretBound*
\end{tcolorbox}

\begin{proof}

\begin{align*}
    \begin{aligned} \mathbb{E}_{\mathcal{B}_{t}}\mathbb{E}_{a_t \in [\K]}\mathbb{E}_{\boldsymbol{\vartheta}}\Big[\boldsymbol{\vartheta}(\bof^{(a^{\ast}_t)} \vert \mathcal{B}_t)  - \boldsymbol{\vartheta}(\bof^{(a_t)} \vert \mathcal{B}_t)\Big] \\
   = \mathbb{E}_{\mathcal{B}_{t}}\Bigg[\mu_{(\boldsymbol\ast)}(\mathcal{B}_{t}) - \mathbb{E}_{a \in [\K]}\mathbb{E}_{\boldsymbol{\vartheta}}\Big[\boldsymbol{\vartheta}(\bof^{(a_t)} \vert \mathcal{B}_t)\Big]\Bigg] \quad \text{where } \mu_{(\bullet)} = \mathrm{E}_{\boldsymbol{\vartheta}}[\boldsymbol{\vartheta}(\bullet)]\\
    = \mathbb{E}_{\mathcal{B}_{t}}\Big[\mu_{(\boldsymbol\ast)}(\mathcal{B}_{t}) - \sum_{j=1}^{\K}\mu_{a}(\mathcal{B}_{t})\mathbb{P}(\bof^{(a_t)} = \bof^{(a)} \mid \mathcal{B}_{t})\Big]\\
     = \mathbb{E}_{\mathcal{B}_{t}}\sum_{a}\bDelta_{a}(\mathcal{B}_{t})\mathbb{P}(\bof^{(a_t)} = \bof^{(a)} \mid \mathcal{B}_{t})\\
     = \sum_{a}\mathbb{E}_{\mathcal{B}_{t}}\bDelta_{a}(\mathcal{B}_{t})\mathbb{P}(\bof^{(a_t)} = \bof^{(a)} \mid \mathcal{B}_{t})
    \end{aligned}
\end{align*}

\begin{align*}
    \begin{aligned}
\mathbb{E}_{\mathcal{B}_{t}}\bDelta_{a}(\mathcal{B}_{t})\mathbb{P}(\bof^{(a_t)} = \bof^{(a)} \mid \mathcal{B}_{t})\\
     \leq \mathbb{E}_{\mathcal{B}_{t}}\bDelta_{a}(\mathcal{B}_{t})\Big[ \bXi_t/\mathcal{K} + \mathbb{P}(\mathcal{M}_{\boldsymbol{\Theta},\bof^{(a_t)}} \geq \mathcal{M}_{\boldsymbol{\Theta}, \bof^{(a^{\ast}_t)}})\Big]
    \end{aligned}
\end{align*}

Let $\mathcal{M}_{\boldsymbol{\Theta}, \bof^{(a_t)}}$ indicates the trained neural network in accordance to \citep{rawson2021convergence} for action $\bof^{(a_t)}$. 

By Markov's inequality

\begin{equation}
\begin{aligned}
   \mathbb{P}(\mathcal{M}_{\boldsymbol{\Theta}, \bof^{(a_t)}} \geq \mathcal{M}_{\boldsymbol{\Theta}, \bof^{(a^{\ast}_t)}})\leq \mathbb{P}\bigg(\mathcal{M}_{\boldsymbol{\Theta}, \bof^{(a_t)}}  \geq \mu_{(a)}(\mathcal{B}_{t}) + \bDelta_{a}(\mathcal{B}_{t})/2\bigg)+  \\
\mathbb{P}\bigg(\mathcal{M}_{\boldsymbol{\Theta}, \bof^{(a^{\ast}_t)}} \leq \mu_{(\boldsymbol{\ast})}(\mathcal{B}_{t}) - \bDelta_{a}(\mathcal{B}_{t})/2\bigg)\\
= \int_{\mathds{1}\{\mathcal{M}_{\boldsymbol{\Theta}, \bof^{(a_t)}}  \geq \mu_{(a)}(\mathcal{B}_{t}) + \bDelta_{a}(\mathcal{B}_{t})/2\}}\partial \mathbb{P}_{a}
+ \int_{\mathds{1}\{\mathcal{M}_{\boldsymbol{\Theta}, \bof^{(a^{\ast}_t)}} \leq \mu_{(\boldsymbol{\ast})}(\mathcal{B}_{t}) - \bDelta_{a}(\mathcal{B}_{t})/2\}}\partial \mathbb{P}_{\boldsymbol{\ast}}\\
\leq \int_{\mathds{1}\{|\mathcal{M}_{\boldsymbol{\Theta}, \bof^{(a_t)}}-\mu_{(a)}(\mathcal{B}_{t})| \geq \bDelta_{(a)}(B_{t})/2\}}\partial\mathbb{P}_{a}
+  \int_{\mathds{1}\{|\mathcal{M}_{\boldsymbol{\Theta}, \bof^{(a^{\ast}_t)}}-\mu_{(\boldsymbol{\ast})}(\mathcal{B}_{t})| \geq \bDelta_{(a)}(B_{t})/2\}}\partial\mathbb{P}_{\boldsymbol{\ast}}\\
\leq \int 4\frac{|\mathcal{M}_{\boldsymbol{\Theta}, \bof^{(a_t)}}-\mu_{(a)}(\mathcal{B}_{t})|^2}{\bDelta_{(a)}(B_{t})^2}\partial \mathbb{P}_{a} + 
\int 4\frac{|\mathcal{M}_{\boldsymbol{\Theta}, \bof^{(a^{\ast}_t)}}-\mu_{(\boldsymbol{\ast})}(\mathcal{B}_{t})|^2}{\bDelta_{(a)}(B_{t})^2}\partial \mathbb{P}_\ast
\end{aligned}
\label{Eq:Left analysis for Theorem 4}
\end{equation}

Based on \emph{Proposition} \ref{proposition:integral-lowerboundfixed-lambda} we have $\boldsymbol{\tau}_{(a)}^R(t) \geq \frac{t -2}{2\mathcal{K}(2-\pi)}(1+(1-\pi)\epsilon)$ for all $a \in \mathrm{A}$.  Let $\mathfrak{C}_{(a)}$ indicate the constant from Theorem \ref{theorem:gyorfi} and let $n_a$ be the minimal training data size. We choose $t_{0} > 
e^{(2\mathcal{K}\max\{e, \max_a n_a\})}$. Since the $x \rightarrow \sqrt{\frac{\ln(x)}{x}}$ is monotone decreasing for $x > e$, the above expression is further bounded by

\begin{equation}
    \begin{aligned}
    \leq \mathbb{E}_{\mathcal{B}_t}\bDelta_{(a)}(\mathcal{B}_t)\frac{\epsilon_t}{\mathcal{K}} + \frac{4}{ \boldsymbol{\varrho}}\mathfrak{C}_{(a)}\sqrt{\frac{\ln(\boldsymbol{\tau}_{(a)}(t))}{\boldsymbol{\tau}_{(a)}(t)}} + \frac{4}{ \boldsymbol{\varrho}}\mathfrak{C}_\ast\sqrt{\frac{\ln(\boldsymbol{\tau}_\ast(t))}{\boldsymbol{\tau}_\ast(t)}}\\
    \leq \mathbb{E}_{\mathcal{B}_t}\bDelta_{(a)}(\mathcal{B}_t)\frac{\epsilon_t}{\mathcal{K}} + \frac{4}{ \boldsymbol{\varrho}}\mathfrak{C}_{(a)}\sqrt{\frac{\ln(\boldsymbol{\tau}^R_i(t))}{\boldsymbol{\tau}^R_i(t)}} +  \frac{4}{ \boldsymbol{\varrho}}\mathfrak{C}_\ast\sqrt{\frac{\ln(\boldsymbol{\tau}^R_\ast(t))}{\boldsymbol{\tau}^R_\ast(t)}}\\
    \leq \frac{\mathbb{E}_{\mathcal{B}_t}\bDelta_{(a)}(\mathcal{B}_t)}{t\mathcal{K}} + \frac{4}{ \boldsymbol{\varrho}}\Bigg[\mathfrak{C}_{(a)} + \mathfrak{C}_\ast \Bigg]\sqrt{\frac{\ln(\frac{t -2}{2\mathcal{K}(2-\pi)}(1+(1-\pi)\epsilon))}{\frac{t -2}{2\mathcal{K}(2-\pi)}(1+(1-\pi)\epsilon)}}\\
\end{aligned}
\end{equation}

Thus we have the following:

\begin{equation} 
\begin{aligned}
\sum_{\bof^{(a_t)}}\mathbb{E}_{\mathcal{B}_{t}}\bDelta_{a}(\mathcal{B}_{t})\mathbb{P}(\bof^{(a_t)} = \bof^{(a)} \mid \mathcal{B}_{t})\\
        \leq \frac{\max_{\bof^{(a_t)} \in \boldsymbol{\mathcal{F}}_{\text{sub}}}\mathbb{E}_{\mathcal{B}_{t}}\bDelta_{a}(\mathcal{B}_{t})}{t} +\\ \mathcal{K}^{3/2}\sqrt{2(2-\pi)}\frac{4}{ \boldsymbol{\varrho}}\Bigg[\max_{a}\mathfrak{C}_{(a)} + \mathfrak{C}_\ast \Bigg]\\
        \sqrt{\frac{\ln((t-2)(1+ \epsilon -\pi\epsilon)) - \ln(2\K(2-\pi))}{(t-2)(1+ \epsilon -\pi\epsilon)}} 
        \end{aligned}
\end{equation}

To showcase the lower bound, we have for $a$-th arm not optimal that,
\begin{equation}
    \begin{aligned}
        \mathbb{E}_{\mathcal{B}_{t}}\bDelta_{a}(\mathcal{B}_{t})\mathbb{P}(\bof^{(a_t)} = \bof^{(a)} \mid \mathcal{B}_{t}) \geq \mathbb{E}_{\mathcal{B}_{t}}\bDelta_{a}(\mathcal{B}_{t})\frac{\bXi_t}{\mathcal{K}} \geq \mathbb{E}_{\mathcal{B}_{t}}\boldsymbol{\varrho}\frac{\bXi_t}{\mathcal{K}} \geq \frac{\boldsymbol{\varrho}}{t\mathcal{K}}
    \end{aligned}
\end{equation}

\end{proof}

\vspace{2em}

\begin{restatable}[\textbf{Bound on Uniform Arm Selection Frequency}]{lemma}{UniformBranchLemma}\label{Lemma:permutation:invariance}

Since $\boldsymbol{\tau}^{R}_{(a)}(t)$ denotes the number of times the $a$-th submodular function $\bof^{(a)}$ is chosen in the first $t-1$ steps by the uniform branch of the algorithm, we have the following:

\begin{align*}
\mathbb{P}\bigg(\bigcap_{a=1}^{\mathcal{K}}\{\boldsymbol{\tau}_{(a)}^R(t) \geq \frac{t -2}{2\mathcal{K}(2-\pi)}(1+(1-\pi)\epsilon)\}\bigg)\\
\geq 1 - \mathcal{K}\exp\bigg(-\frac{3(t-2)(1+(1-\pi)\epsilon)}{28\mathcal{K}(2-\pi)}\bigg)
\end{align*}

\end{restatable}

\begin{proof}
\begin{equation}
\begin{aligned}
    \mathbb{E}(\boldsymbol{\tau}^R_{(a)}(t)) = \sum_{r=1}^{t-1}\mathbb{P}(\zeta < \bXi_r \cap \bof^{(a_t)} = \bof^{(a)})\\
    = \sum_{r=1}^{t-1}\mathbb{P}(\zeta < \bXi_r)\mathbb{P}(\bof^{(a_t)} = \bof^{(a)})
    = \sum_{r=1}^{t-1}\frac{\bXi_r}{\mathcal{K}} = \frac{1}{\mathcal{K}}\sum_{r=1}^{t-1}\frac{r}{(r+\lambda(r))^\pi}\\
    \geq \frac{1}{\mathcal{K}}\int_{x = 1}^{x=t-1}\frac{x}{(x+\lambda(x))^\pi}\partial x
    \geq \frac{t -2}{\mathcal{K}(2-\pi)}(1+(1-\pi)\epsilon)
        \end{aligned}
        \end{equation}

where, the last inequality is based on $\emph{Proposition}$ \ref{proposition:integral-lowerboundfixed-lambda}.
We define the variance of  $\boldsymbol{\tau}^R_{(a)}(t)$ as $\sigma(\boldsymbol{\tau}^R_{(a)}(t))$ and the corresponding upperbound as $\mathcal{Z}(\sigma(t))$
\begin{align*}
    \begin{aligned}
\sigma(\boldsymbol{\tau}^R_{(a)}(t)) = \sum_{r=1}^{t-1}\frac{\bXi_r}{\mathcal{K}}(1-\frac{\bXi_r}{\mathcal{K}}) \leq \frac{1}{\mathcal{K}}\sum_{r=1}^{t-1}\bXi_r = \frac{1}{\mathcal{K}}\sum_{r=1}^{t-1}\frac{r}{(r+\lambda(r))^{\pi}} = \mathcal{Z}(\sigma(t))
    \end{aligned}
\end{align*}
Using Bernstein's inequality
\begin{align*}
    \begin{aligned}
\mathbb{P}\bigg(\boldsymbol{\tau}^R_{(a)}(t) \leq \frac{\mathcal{Z}(\sigma(t))}{2}\bigg) = \mathbb{P}\bigg(\boldsymbol{\tau}^R_{(a)}(t) - \mathcal{Z}(\sigma(t)) \leq -\frac{\mathcal{Z}(\sigma(t))}{2})\bigg)\\
        \leq \exp\bigg({\frac{\frac{-\mathcal{Z}(\sigma(t))^2}{8}}{\sigma(\boldsymbol{\tau}^R_{(a)}(t)) + \frac{1}{3}\frac{\mathcal{Z}(\sigma(t))}{2}}}\bigg)
               \leq \exp\bigg({\frac{\frac{-\mathcal{Z}(\sigma(t))^2}{8}}{\mathcal{Z}(\sigma(t)) + \frac{1}{3}\frac{\mathcal{Z}(\sigma(t))}{2}}}\bigg)\\
                \le \exp\bigg(-\frac{3\mathcal{Z}(\sigma(t))}{28}\bigg)
                \leq \exp\bigg(-\frac{3(t-2)(1+(1-\pi)\epsilon)}{28\mathcal{K}(2-\pi)}\bigg)
    \end{aligned}
\end{align*}
By union bound method
\begin{align*}
    \begin{aligned}
\mathbb{P}\bigg(\bigcup_{a=1}^{\mathcal{K}}\{\boldsymbol{\tau}_{(a)}^R(t) \leq \frac{t -2}{2\mathcal{K}(2-\pi)}(1+(1-\pi)\epsilon)\}\bigg) \\
\leq \mathcal{K}\mathbb{P}\bigg(\boldsymbol{\tau}_{(1)}^R(t)
\leq \frac{t -2}{2\mathcal{K}(2-\pi)}(1+(1-\pi)\epsilon)\bigg)
\leq \mathcal{K}\exp\bigg(-\frac{3(t-2)(1+(1-\pi)\epsilon)}{28\mathcal{K}(2-\pi)}\bigg)
    \end{aligned}
\end{align*}
Therefore
\begin{align*}
\begin{aligned}
\mathbb{P}\bigg(\bigcap_{a=1}^{\mathcal{K}}\{\boldsymbol{\tau}_{(a)}^R(t) \geq \frac{t -2}{2\mathcal{K}(2-\pi)}(1+(1-\pi)\epsilon)\}\bigg)
\geq 1 - \mathcal{K}\exp\bigg(-\frac{3(t-2)(1+(1-\pi)\epsilon)}{28\mathcal{K}(2-\pi)}\bigg)
\end{aligned}
\end{align*}
\end{proof}

\newpage

\begin{restatable}[\textbf{Integral Lower Bound (Constant $\lambda$)}]{proposition}{LowerboundFixedLambda}\label{proposition:integral-lowerboundfixed-lambda}
Let $\lambda(t) = \epsilon$ with $0 < \epsilon < 1$, and $0 < \pi < 1$. Then, for 
\[
I_t = \int_{x=1}^{t-1} \frac{x}{(x + \lambda(x))^\pi} \, \mathrm{d}x,
\]
we have
\[
I_t \;\ge\; \int_{x=1}^{t-1} \frac{x}{(x + \epsilon)^\pi} \, \mathrm{d}x 
\;\ge\; \frac{t-2}{2-\pi} \,(1 + (1-\pi)\epsilon).
\]
\end{restatable}

\begin{proof}
\begin{equation}
\begin{aligned}
\int_{x=1}^{x^{t-1}} \frac{x}{(x + \lambda(x))^\pi} \, \mathrm{d}x 
&= \int_{x=1}^{x=t-1} \frac{x}{(x + \epsilon)^\pi} \, \mathrm{d}x 
&& \hspace{-5em} \text{(Substitute } \lambda(x) = \epsilon\text{)} \\
&= \frac{[t - (1 - \epsilon)]^{2 - \pi}}{2 - \pi}
  - \frac{(1 + \epsilon)^{2 - \pi}}{2 - \pi} \\
&\quad - \epsilon \cdot \frac{[t - (1 - \epsilon)]^{1 - \pi}}{1 - \pi}
  + \epsilon \cdot \frac{(1 + \epsilon)^{1 - \pi}}{1 - \pi} \\
&= [t - (1 - \epsilon)]^{1 - \pi}
    \left( \frac{(1 - \pi)(t - 1) - \epsilon}{(1 - \pi)(2 - \pi)} \right) \\
&\quad - [1 + \epsilon]^{1 - \pi}
    \left( \frac{1 - \pi - \epsilon}{(1 - \pi)(2 - \pi)} \right) \\
&= [m + \epsilon]^{1 - \pi}
    \left( \frac{(1 - \pi)m - \epsilon}{(1 - \pi)(2 - \pi)} \right) \\
&\quad - [1 + \epsilon]^{1 - \pi}
    \left( \frac{(1 - \pi) - \epsilon}{(1 - \pi)(2 - \pi)} \right) 
&& \hspace{-5em} \text{(Let } m = t - 1\text{)} \\
&= [m + \epsilon]^k \left( \frac{km - \epsilon}{k(k+1)} \right)
  - [1 + \epsilon]^k \left( \frac{k - \epsilon}{k(k+1)} \right)
&&  \text{(Let } k = 1 - \pi\text{)} \\
&= \frac{1}{k(k+1)} \left( [m + \epsilon]^k(km - \epsilon) 
  + (\epsilon - k)[1 + \epsilon]^k \right) \\
&\geq \frac{1}{k(k+1)} \left( [1 + \epsilon]^k(km - \epsilon) 
  + (\epsilon - k)[1 + \epsilon]^k \right) \\
&\quad \text{(Since } m + \epsilon \geq 1 + \epsilon\text{)} \\
&= \frac{1}{k(k+1)} [1 + \epsilon]^k (km - \epsilon + \epsilon - k) \\
&= \frac{1}{k(k+1)} [1 + \epsilon]^k (km - k) \\
&= \frac{1}{k+1} [1 + \epsilon]^k (m - 1) \\
&\geq \frac{1}{k+1} [1 + k\epsilon](m - 1) 
&& \hspace{-10em} \text{(Using } (1 + \epsilon)^k \geq 1 + k\epsilon \text{ for small } \epsilon\text{)} \\
&= \frac{1}{2 - \pi} [1 + (1 - \pi)\epsilon](t - 2) 
&& \hspace{-8em} \text{(Substitute } k = 1 - \pi,\, m = t - 1\text{)} \\
&= \frac{t - 2}{2 - \pi}(1 + (1 - \pi)\epsilon)
\end{aligned}
\end{equation}
\end{proof}

The above integral computation is used in the main paper for our proofs.

\subsection{Regret bounds in the case of growing with time exploration dampening  function}

\begin{restatable}[\textbf{Integral Lower Bound (Exponential Growing $\lambda$)}]{proposition}{LowerboundExpLambda}\label{prop:integral-lower-bound-exp}
Let $\lambda(t) = 1 - e^{-\mathfrak{i} t}$ be a time-growing function with rate $\mathfrak{i} > 0$, and let $0 < \pi < 1$. Then, for
\[
I_t = \int_{x=1}^{t-1} \frac{x}{(x + \lambda(x))^\pi} \, \mathrm{d}x,
\]
the following lower bound holds:
\[
I_t \;\ge\; \int_{x=1}^{t-1} \frac{x}{\bigl(x + 1 - e^{-\mathfrak{i} x}\bigr)^\pi} \, \mathrm{d}x
\;\ge\; \left( \frac{1}{2 \mathfrak{i}} \Bigl[ \ln\bigl(2 e^{\mathfrak{i}(t-1)} - 1\bigr) - \ln\bigl(2 e^{\mathfrak{i}} - 1\bigr) \Bigr] \right)^\pi.
\]
\end{restatable}

\begin{proof}
\begin{equation*}
\begin{aligned}
\int_{x=1}^{t-1} \frac{x}{(x + \lambda(x))^\pi} \, \mathrm{d}x
&= \int_{x=1}^{t-1} \frac{x}{(x + 1 - e^{-\mathfrak{i}x})^\pi} \, \mathrm{d}x \\
&\geq \int_{x=1}^{t-1} \frac{x^\pi}{(x + 1 - e^{-\mathfrak{i}x})^\pi} \, \mathrm{d}x \\
&\geq \int_{x=1}^{t-1} \left( \frac{x e^{\mathfrak{i}x}}{x e^{\mathfrak{i}x} + e^{\mathfrak{i}x} - 1} \right)^\pi \, \mathrm{d}x \\
&\geq \int_{x=1}^{t-1} \left( \frac{x e^{\mathfrak{i}x}}{x e^{\mathfrak{i}x} + x e^{\mathfrak{i}x} - x} \right)^\pi \, \mathrm{d}x \\
&= \int_{x=1}^{t-1} \left( \frac{e^{\mathfrak{i}x}}{2 e^{\mathfrak{i}x} - 1} \right)^\pi \, \mathrm{d}x 
&& \text{Using Jensen's Inequality}
\\
&\geq \left[ \int_{x=1}^{t-1} \frac{e^{\mathfrak{i}x}}{2 e^{\mathfrak{i}x} - 1} \, \mathrm{d}x \right]^\pi \\
&= \left( \frac{1}{2\mathfrak{i}} \left[ \ln(2e^{\mathfrak{i}(t - 1)} - 1) - \ln(2e^{\mathfrak{i}} - 1) \right] \right)^\pi
\end{aligned}
\end{equation*}
\end{proof}

\begin{restatable}[Exploration Dampening: Annealing]{lemma}{ExplorationDampeningAnnealing}
\label{lem:exploration-dampening-annealing}
Since $\boldsymbol{\tau}^{R}_{(a)}(t)$ denotes the number of times the $a$-th submodular action is chosen in the first $t-1$ steps by the uniform branch of the algorithm, $a \in [\mathcal{K}]$, then in the case of $\lambda(t) =1 - e^{-\mathfrak{i}t}$ , for $\mathfrak{i} \geq 0$(i.e. growing  exploration dampening probability), we have the following:

\begin{align*}
\mathbb{P}\bigg(\bigcap_{j=1}^{\mathcal{K}}\{\boldsymbol{\tau}_{(a)}^R(t) \geq \frac{1}{2\mathcal{K}}\Bigg(\frac{1}{2\mathfrak{i}}\Big[\ln(2e^{\mathfrak{i}(t-1)} - 1) - \ln(2e^{(a)} - 1)\Big]\Bigg)^{\pi}\}\bigg)\\
\geq 1 - \mathcal{K}\exp\bigg(-\frac{3}{28\mathcal{K}}\Bigg(\frac{1}{2\mathfrak{i}}\Big[\ln(2e^{\mathfrak{i}(t-1)} - 1) - \ln(2e^{(a)} - 1)\Big]\Bigg)^{\pi}\bigg)
\end{align*}
\end{restatable}

\begin{proof}
\begin{align*}
\begin{aligned}
    \mathbb{E}(\boldsymbol{\tau}^R_{a}(t)) = \sum_{r=1}^{t-1}\mathbb{P}(\zeta < \bXi_r \cap f_t = f^j)\\
    = \sum_{r=1}^{t-1}\mathbb{P}(\zeta < \bXi_r)\mathbb{P}(\bof^{(a_t)} = \bof^{(a)})
    = \sum_{r=1}^{t-1}\frac{\bXi_r}{\mathcal{K}} = \frac{1}{\mathcal{K}}\sum_{r=1}^{t-1}\frac{r}{(r+\lambda(r))^\pi}\\
    \geq \frac{1}{\mathcal{K}}\int_{x = 1}^{x=t-1}\frac{x}{(x+\lambda(x))^\pi}\partial x 
    =\frac{1}{\mathcal{K}}\int_{x = 1}^{x=t-1}\frac{x}{(x+1-e^{-\mathfrak{i}x})^\pi}\partial x \\
    \geq \frac{1}{\mathcal{K}}\Bigg(\frac{1}{2\mathfrak{i}}\Big[\ln(2e^{\mathfrak{i}(t-1)} - 1) - \ln(2e^a - 1)\Big]\Bigg)^{\pi}
     \end{aligned}
    \end{align*}

The last inequality comes from \emph{Proposition} \ref{prop:integral-lower-bound-exp}

    We define the variance of  $\boldsymbol{\tau}^R_{(a)}(t)$ as $\sigma(\boldsymbol{\tau}^R_{(a)}(t))$ and the corresponding upperbound as $\mathcal{Z}(\sigma(t))$

\begin{align*}
    \begin{aligned}
\sigma(\boldsymbol{\tau}^R_{a}(t)) = \sum_{r=1}^{t-1}\frac{\bXi}{\mathcal{K}}(1-\frac{\bXi}{\mathcal{K}}) \leq \frac{1}{\mathcal{K}}\sum_{r=1}^{t-1}\bXi_r = \frac{1}{\mathcal{K}}\sum_{r=1}^{t-1}\frac{r}{(r+1-e^{-\mathfrak{i}r})^{\pi}} = \mathcal{Z}(\sigma(t))
    \end{aligned}
\end{align*}

Using Bernstein's inequality

\begin{align*}
    \begin{aligned}
\mathbb{P}\bigg(\boldsymbol{\tau}^R_{a}(t) \leq \frac{\mathcal{Z}(\sigma(t))}{2}\bigg) = \mathbb{P}\bigg(\boldsymbol{\tau}^R_{a}(t) - \mathcal{Z}(\sigma(t)) \leq -\frac{\mathcal{Z}(\sigma(t))}{2})\bigg)\\
        \leq \exp\bigg({\frac{\frac{-\mathcal{Z}(\sigma(t))^2}{8}}{\sigma(\boldsymbol{\tau}^R_{a}(t)) + \frac{1}{3}\frac{\mathcal{Z}(\sigma(t))}{2}}}\bigg)
               \leq \exp\bigg({\frac{\frac{-\mathcal{Z}(\sigma(t))^2}{8}}{\mathcal{Z}(\sigma(t)) + \frac{1}{3}\frac{\mathcal{Z}(\sigma(t))}{2}}}\bigg)\\
                \le \exp\bigg(-\frac{3\mathcal{Z}(\sigma(t))}{28}\bigg)
                \leq \exp\bigg(-\frac{3}{28\mathcal{K}}\Bigg(\frac{1}{2\mathfrak{i}}\Big[\ln(2e^{\mathfrak{i}(t-1)} - 1) - \ln(2e^{(a)} - 1)\Big]\Bigg)^{\pi}\bigg)
    \end{aligned}
\end{align*}

By union bound method
\begin{equation*}
    \begin{aligned}
\mathbb{P}\bigg(\bigcup_{j=1}^{\mathcal{K}}\{\boldsymbol{\tau}_{(a)}^R(t) \leq \frac{1}{2\mathcal{K}}\Bigg(\frac{1}{2\mathfrak{i}}\Big[\ln(2e^{\mathfrak{i}(t-1)} - 1) - \ln(2e^{(a)} - 1)\Big]\Bigg)^{\pi} \}\bigg)\\
        \leq \mathcal{K}\mathbb{P}\bigg(\boldsymbol{\tau}_{1}^R(t) \leq 
         \frac{1}{2\mathcal{K}}\Bigg(\frac{1}{2\mathfrak{i}}\Big[\ln(2e^{\mathfrak{i}(t-1)} - 1) - \ln(2e^{(a)} - 1)\Big]\Bigg)^{\pi}
         \bigg)\\
        \leq \mathcal{K}\exp\bigg(-\frac{3}{28\mathcal{K}}\Bigg(\frac{1}{2\mathfrak{i}}\Big[\ln(2e^{\mathfrak{i}(t-1)} - 1) - \ln(2e^{(a)} - 1)\Big]\Bigg)^{\pi}\bigg)
    \end{aligned}
\end{equation*}

Therefore
\begin{equation*}
\begin{aligned}
\mathbb{P}\bigg(\bigcap_{j=1}^{\mathcal{K}}\{\boldsymbol{\tau}_{(a)}^R(t) \geq  \frac{1}{2\mathcal{K}}\Bigg(\frac{1}{2\mathfrak{i}}\Big[\ln(2e^{\mathfrak{i}(t-1)} - 1) - \ln(2e^{(a)} - 1)\Big]\Bigg)  ^{\pi}\}\bigg)\\
\geq 1 - \mathcal{K}\exp\bigg(-\frac{3}{28\mathcal{K}}\Bigg(\frac{1}{2\mathfrak{i}}\Big[\ln(2e^{\mathfrak{i}(t-1)} - 1) - \ln(2e^{(a)} - 1)\Big]\Bigg)^{\pi}\bigg)
\end{aligned}
\end{equation*}
\end{proof}

\newpage

\begin{restatable}[\textbf{Regret Guarantees}]{theorem}{RegretBoundExpGrow}\label{theorem:regretbound_exp_grow}
Under Assumptions \textbf{b - d}, for all \(t > t_0\) and with $\lambda(t)= 1 - e^{-\mathfrak{i}t}$, with probability at least
\[
1 - \mathcal{K}\exp\bigg(-\frac{3}{28\mathcal{K}}\Bigg(\frac{1}{2\mathfrak{i}}\Big[\ln(2e^{\mathfrak{i}(t-1)} - 1) - \ln(2e^{(a)} - 1)\Big]\Bigg)^{\pi}\bigg)
\]
the expected instantaneous regret incurred by the arm selection policy satisfies
\begin{equation}
\begin{split}
\mathds{E}[\mathrm{Regret}_t] &:= \mathds{E}_{\mathcal{B}_t} \mathds{E}_{\hat{a}_t \in \mathscr{A}} \mathds{E}_{\boldsymbol{\vartheta}} \left[ \boldsymbol{\vartheta}(a_t^* \mid \mathcal{B}_t) - \boldsymbol{\vartheta}(\hat{a}_t \mid \mathcal{B}_t) \right] \\
&= O\left(\frac{1}{t}\right) + O\Bigg(\frac{4(\mathfrak{C}_{(a)} + \mathfrak{C}_\ast)}{\varrho}\, t^{-\pi/2} \sqrt{\ln t}\Bigg)
,
\end{split}
\label{eq:regretbound}
\end{equation}
where \(\mathfrak{C}_*\) is the approximation constant corresponding to the optimal arm \(a^*\).
\end{restatable}

\begin{proof}
    
Continuing from the same step in Section \ref{Constant Lambda Proof} Theorem \ref{theorem:regretbound}, we have the following: 

For the case of $\boldsymbol{\tau}_{(a)}^R(t) \geq \frac{1}{2\mathcal{K}}\Bigg(\frac{1}{2\mathfrak{i}}\Big[\ln(2e^{\mathfrak{i}(t-1)} - 1) - \ln(2e^{(a)} - 1)\Big]\Bigg)^{\pi}$ for all $a$ via \emph{Proposition} \ref{prop:integral-lower-bound-exp}. Let $\mathfrak{C}_{(a)}$ indicate the constant from Theorem \ref{theorem:gyorfi} and let $n_a$ be the minimal training data size. We choose $t_{0} > 
e^{(2\mathcal{K}\max\{e, \max_a n_a\})}$. Since the $x \rightarrow \sqrt{\frac{\ln(x)}{x}}$ is monotone decreasing for $x > e$, the above expression is further bounded by

\begin{equation}
    \begin{aligned}
    \leq \mathbb{E}_{\mathcal{B}_t}\bDelta_{(a)}(\mathcal{B}_t)\frac{\epsilon_t}{\mathcal{K}} + \frac{4}{ \boldsymbol{\varrho}}\mathfrak{C}_{(a)}\sqrt{\frac{\ln(\boldsymbol{\tau}_{(a)}(t))}{\boldsymbol{\tau}_{(a)}(t)}} + \frac{4}{ \boldsymbol{\varrho}}\mathfrak{C}_\ast\sqrt{\frac{\ln(\boldsymbol{\tau}_\ast(t))}{\boldsymbol{\tau}_\ast(t)}}\\
    \leq \mathbb{E}_{\mathcal{B}_t}\bDelta_{(a)}(\mathcal{B}_t)\frac{\epsilon_t}{\mathcal{K}} + \frac{4}{ \boldsymbol{\varrho}}\mathfrak{C}_{(a)}\sqrt{\frac{\ln(\boldsymbol{\tau}^R_i(t))}{\boldsymbol{\tau}^R_i(t)}} +  \frac{4}{ \boldsymbol{\varrho}}\mathfrak{C}_\ast\sqrt{\frac{\ln(\boldsymbol{\tau}^R_\ast(t))}{\boldsymbol{\tau}^R_\ast(t)}}\\
    \leq \frac{\mathbb{E}_{\mathcal{B}_t}\bDelta_{(a)}(\mathcal{B}_t)}{t\mathcal{K}} + \frac{4}{ \boldsymbol{\varrho}}\Bigg[\mathfrak{C}_{(a)} + \mathfrak{C}_\ast \Bigg]\sqrt{\frac{\ln(\frac{1}{2\mathcal{K}}\Bigg(\frac{1}{2\mathfrak{i}}\Big[\ln(2e^{\mathfrak{i}(t-1)} - 1) - \ln(2e^{(a)} - 1)\Big]\Bigg)^{\pi})}{\frac{1}{2\mathcal{K}}\Bigg(\frac{1}{2\mathfrak{i}}\Big[\ln(2e^{\mathfrak{i}(t-1)} - 1) - \ln(2e^{(a)} - 1)\Big]\Bigg)^{\pi}}}\\
\end{aligned}
\end{equation}

\end{proof}

\section{Broader Impact}\label{Supplementary:Impact}

\vspace{-0.8em}

The primary aim of our work is to improve the data efficiency of machine learning training pipelines via submodular subset selection. By leveraging principled selection algorithms—such as monotone submodular functions, we can reduce the number of training examples needed without sacrificing model performance. This contributes directly to more sustainable and accessible machine learning, especially in scenarios where training data or compute is limited.

\textbf{Societal and Environmental Benefits}: Reducing the amount of data required for training has multiple practical benefits. For large-scale models, this can translate into lower energy consumption and a reduced carbon footprint. For smaller research labs or applications in low-resource settings, our approach can make training state-of-the-art models more feasible.

\textbf{Equity and Fairness}: By allowing for careful and task-informed selection of training data, our methods could help surface underrepresented or domain-critical samples early in training. However, care must be taken to ensure that the subset selection process does not reinforce existing dataset biases. We encourage practitioners to combine our framework with fairness-aware selection techniques and to audit the resulting models for any performance disparities across groups.

\textbf{Scientific Impact}: More broadly, this work highlights the growing role of data-centric approaches in machine learning research, particularly for compute efficient machine learning research.

%\newpage
\end{document}